\documentclass[9pt,twoside]{ilcss}

\usepackage{subfigure}
\usepackage{float,lipsum,tcolorbox}
\usepackage{tablefootnote}
\usepackage[flushleft]{threeparttable} 
\usepackage{alphalph}

\usepackage{array}
\newfloat{info@box}{tbp}{loi}[section]
\makeatother
\floatname{info@box}{Infobox}

\templatetype{ilcssworkingpaper} 

\title{IGNITE: Individualized GeNeration of Imputations in Time-series Electronic health records}	

\author[$\dagger$,*]{Ghadeer O. Ghosheh}
\author[$\S$]{Jin Li}
\author[$\dagger$]{Tingting Zhu}

\affil[$\dagger$]{Department of Engineering Sciences, University of Oxford}
\affil[$\S$]{Nanjing University of Information Science \& Technology (NUIST)}
\affil[*]{Correspondence: ghadeer.ghosheh@eng.ox.ac.uk}

\leadauthor{Ghosheh} 

\correspondingauthor{\textsuperscript{*}Correspondence: ghadeer.ghosheh@eng.ox.ac.uk}

\begin{abstract}
\dropcap{E}lectronic Health Records present a valuable modality for driving personalized medicine, where treatment is tailored to fit individual-level differences. For this purpose, many data-driven machine learning and statistical models rely on the wealth of longitudinal EHRs to study patients' physiological and treatment effects. However, longitudinal EHRs tend to be sparse and highly missing, where missingness could also be informative and reflect the underlying patient's health status. Therefore, the success of data-driven models for personalized medicine highly depends on how the EHR data is represented from physiological data, treatments, and the missing values in the data. To this end, we propose a novel deep-learning model that learns the underlying patient dynamics over time across multivariate data to generate personalized realistic values conditioning on an individual's demographic characteristics and treatments. Our proposed model, IGNITE (Individualized GeNeration of Imputations in Time-series Electronic health records), utilises a conditional dual-variational autoencoder augmented with dual-stage attention to generate missing values for an individual. In IGNITE, we further propose a novel individualized missingness mask (IMM), which helps our model generate values based on the individual's observed data and missingness patterns. We further extend the use of IGNITE from imputing missingness to a personalized data synthesizer, where it generates missing EHRs that were never observed prior or even generates new patients for various applications. We validate our model on three large publicly available datasets and show that IGNITE outperforms state-of-the-art approaches in missing data reconstruction and task prediction.

\end{abstract}


\begin{document}
\maketitle
\thispagestyle{firststyle}
\ifthenelse{\boolean{shortarticle}}{\ifthenelse{\boolean{singlecolumn}}{\abscontentformatted}{\abscontent}}{}

\section{Introduction}
Personalized medicine is emerging as a new direction of research, in which patients are holistically examined as ``individuals'' rather than using symptoms or diagnoses based on the general population to determine the optimal treatment~\citep{branco2021bioinformatics,lourida2021constraints}. Patients differ in physiology and response to treatment, where each individual has unique characteristics and patterns that reflect their genetic, molecular, and cellular makeup, as well as other environmental and behavioral factors~\citep{lourida2021constraints}. 
Electronic health records (EHRs) have opened the door to the use of longitudinal data from real-world patients for a wide range of research, such as creating clinical decision support systems for different medical applications ~\citep{cowie2017electronic,kim2019evolving}. The adoption of EHR has increased worldwide, including in low-middle-income countries (LMICs)~\citep{syzdykova2017open}. Furthermore, the wealth of data collected in EHRs presents a great source for developing approaches in medicine that tailor interventions to best suit each individual patient~\citep{abul2019personalized}. In addition, recent work investigated the use of EHRs to generate digital twins of patients, where real-time monitoring, diagnosis, prognosis, and treatment optimization are applied to personalize the medicine~\citep{kamel2021digital,venkatesh2022health}.

Data-driven approaches, specifically machine learning models~\citep{xiao2018opportunities,shamout2020machine},  have shown promising results in various clinical applications, especially when using longitudinal EHR data. However, EHRs tend to be multivariate, highly missing, and irregularly sampled~\citep{madden2016missing}. Therefore, the success of machine-learning models for personalized medicine is highly dependent on how the EHR data are represented conditioning on the time-varying physiology, treatments, and the missing values in the data. The existing body of work on personalized medicine mostly emphasizes the utilization of observable data~\cite{chen2019deep,wang2016learning}. However, this approach tends to overlook the crucial aspect of missingness, which is essential to achieve true individualization and to obtain meaningful insights. Missingness in the healthcare domain can occur due to various reasons, such as recording errors and device or system failure, irregular sampling, and inconsistent medical visits~\citep{kreindler2006effects}. Even high-cost and dangerous data acquisition, such as invasive or radiological procedures, may be missing due to the risk or lack of feasibility of collection for all patients~\citep{bulas2013benefits,kim2017dangers}. Furthermore, the measurement frequency of physiological data could also be related to factors such as patient severity and deterioration~\citep{agniel2018biases,ghosheh2022review}, lack of medical need~\citep{duff2019frequency},  bias and low quality of care~\citep{weber2021gender,wells2013strategies}. The aforementioned characteristics exhibit variability between patients, underscoring the importance of representing the individualized missingness in personalized models.

The handling of missingness in longitudinal EHRs is a critical step before performing statistical or machine learning analysis, as missingness can reach as high as 80\% or more in many EHR datasets~\citep{silva2012predicting,johnson2016mimic,piri2020missing}. Processing missingness usually involves data deletion or imputation. Simply deleting recordings or variables with missing values often leads to compromised results and is not recommended practice, especially when the data is highly missing~\citep{scheffer2002dealing,king1998list,newman2014missing}. Missing value imputation and replacing unobserved values with substitute values methods gained more attention with the advancements in statistical and machine-learning-based models. Current missing-value imputation methods can be classified into three main categories: The first category of methods, known as simple imputations, includes replacing missing values with population mean, median, or personalized Last (observed) Value Carried Forward (LOCF)~\citep{acuna2004treatment,shao2003last}. Although useful and commonly used in various clinical applications such as clinical trials, imputing by the LOCF presents an issue in observational EHRs, particularly when the feature is never observed for the patient, hence no value to carry forward. The second type of imputation method is a simulation-based statistical class of methods, often called multiple imputation methods. For multiple imputation methods,  $N$ complete datasets are simulated from the incomplete dataset by imputing the missing values $N$ times. These completed $N$ datasets are analyzed and then pooled into a single dataset~\citep{jakobsen2017and}. A common multiple imputation method is multivariate imputation by chained equations (MICE), which is often used by many epidemiological and clinical research~\citep{sterne2009multiple}.\textcolor{black}{ Despite the usefulness of many approaches, they often fail to handle missingness in longitudinal EHRs as they cannot account for time-series sequential data. Although MICE is useful for certain imputation tasks, they may not be as effective in modeling the sequential nature of time-series data as they read the data in a flattened form. This approach often treats each time point independently, which can lead to suboptimal imputations when dealing with the continuous and interdependent nature of time-series data.} The third and most advanced category of imputation methods is based on machine learning algorithms. Matrix Factorization (MF)~\citep{ranjbar2015imputation}, Expectation Maximization (EM)~\citep{gold2000treatments}, K-Nearest Neighbor (KNN) -based imputation~\citep{malarvizhi2012k}, along with Recurrent Neural Network (RNN)~\citep{cao2018brits}.

Recently, deep generative neural network-based models such as Generative Adversarial Networks (GAN)~\citep{goodfellow2014generative} and Variational Autoencoder (VAEs)~\citep{kingma2013auto} were proposed to generate synthetic data by learning the underlying distribution and dynamics of real datasets. The applications of deep generative models in healthcare range from generating synthetic patient records~\citep{choi2017generating,li2023generating}, to estimating treatment effects~\citep{yoon2018ganite}, detecting un-diagnosed patients in large-scale ~\citep{li2018semi}, and generating missing value imputations~\citep{yoon2018gain,nazabal2020handling,fortuin2020gp,luo2019e2gan}. \textcolor{black}{The generative capabilities of deep generative models make them naturally suitable to generate not only new synthetic records but also missing data imputations, where they show superior performance to most of the commonly used statistical methods in tabular and longitudinal data~\citep{kim2020survey,ghosheh2024survey,wang2024deep}}.
Despite their high reported performance, most deep generative methods assume \textit{missingness is missing at random}, which limits their potential in real healthcare applications, where missingness can be informative and reflect individual underlying patient state~\citep{scheffer2002dealing}. Furthermore, many of these models can only generate \textit{population-level} rather than \textit{personalized-level} information conditioned on the already observed patient data, characteristics and treatments. 

In our evaluation of IGNITE, we have focused on multivariate time-series imputation within the context of electronic health records (EHRs), where continuous monitoring and recording of patient information over time is prevalent. This necessitates the inclusion of methods that are directly comparable in this domain. Although influential methods such as MD-MTS (Xu 2020) and SMILES (Zhang 2020) have been cited in the literature, these methods are primarily focused on static data imputation or specific to non-multivariate time-series data. Our work specifically targets the unique challenges of multivariate time-series imputation in EHRs. A comparison of previous works is shown in Table \ref{tab:comparison}.

\begin{table*}[!ht]
    \centering
         \caption{Comparison of different missing data imputation methods in terms of the architecture, compatibility with time-series data, data type and ability to account for individualized missingness.}
    \resizebox{0.95\linewidth}{!}{

\begin{threeparttable}
\begin{tabular}{lcccc}

    \toprule
    \textbf{Model} & \textbf{Architecture} & \textbf{Generative Model}& \textbf{Time-series Data Type} &\textbf{Individualized Missingness Indicators} \\
    \midrule
        MICE \citep{van2011mice} & Statistical regression & \xmark &N.A. & \xmark \\ 

        GAIN \citep{yoon2018gain} & GAN & \bluecheck  &N.A. & \xmark \\ 
        BRITS \citep{cao2018brits} & RNN & \xmark  &continuous& \xmark \\
        HI-VAE \citep{nazabal2020handling} & VAE & \bluecheck  &N.A. & \xmark\\ 
        GP-VAE \citep{fortuin2020gp} & VAE & \bluecheck &continuous & \xmark\\
        TimesNet \citep{wu2022timesnet} & Temporal 2D-Variation Modeling &\xmark & continuous &  \xmark\\
        SAITS \citep{du2023saits} &  Self-attention-based model & \xmark &continuous & \xmark\\ 

        IGNITE & VAE-GAN & \bluecheck  & continuous w/ conditioning on discrete & \bluecheck  \\

        \bottomrule
    \end{tabular}
    \begin{tablenotes}
  \item[*] N.A: Note Applicable, as the methods are proposed for static data.
  \item[a] Acrynoms in Full Form: GAN: Generative Adversarial Networks, RNN:  Recurrent Neural Network, and VAE: Variational Autoencoder : 
  \end{tablenotes}
  \end{threeparttable}}
    \label{tab:comparison}
\end{table*}

The utilization of binary missingness masks that indicate the presence or absence of observations as an additional input feature to prediction models has shown improved performance compared to models built on observed data only~\citep{sharafoddini2019new,che2018recurrent}. Despite the utility and predictiveness of binary missingness masks, they are simplistic and lack indications of the patterns and frequencies of individual-level missingness. We believe that to generate missing values at a personalized level, robust models that leverage heterogeneous and multiple data types (such as continuous physiological data, discrete treatment data, and individualized missingness masks) are required. Utilizing these mixed data types would capture the non-linear dependencies across high-dimensional longitudinal patient records and generate EHR information that is truly personalized.  

\begin{figure*}[!htbp]
\begin{center}
\includegraphics[width= 0.9\textwidth]{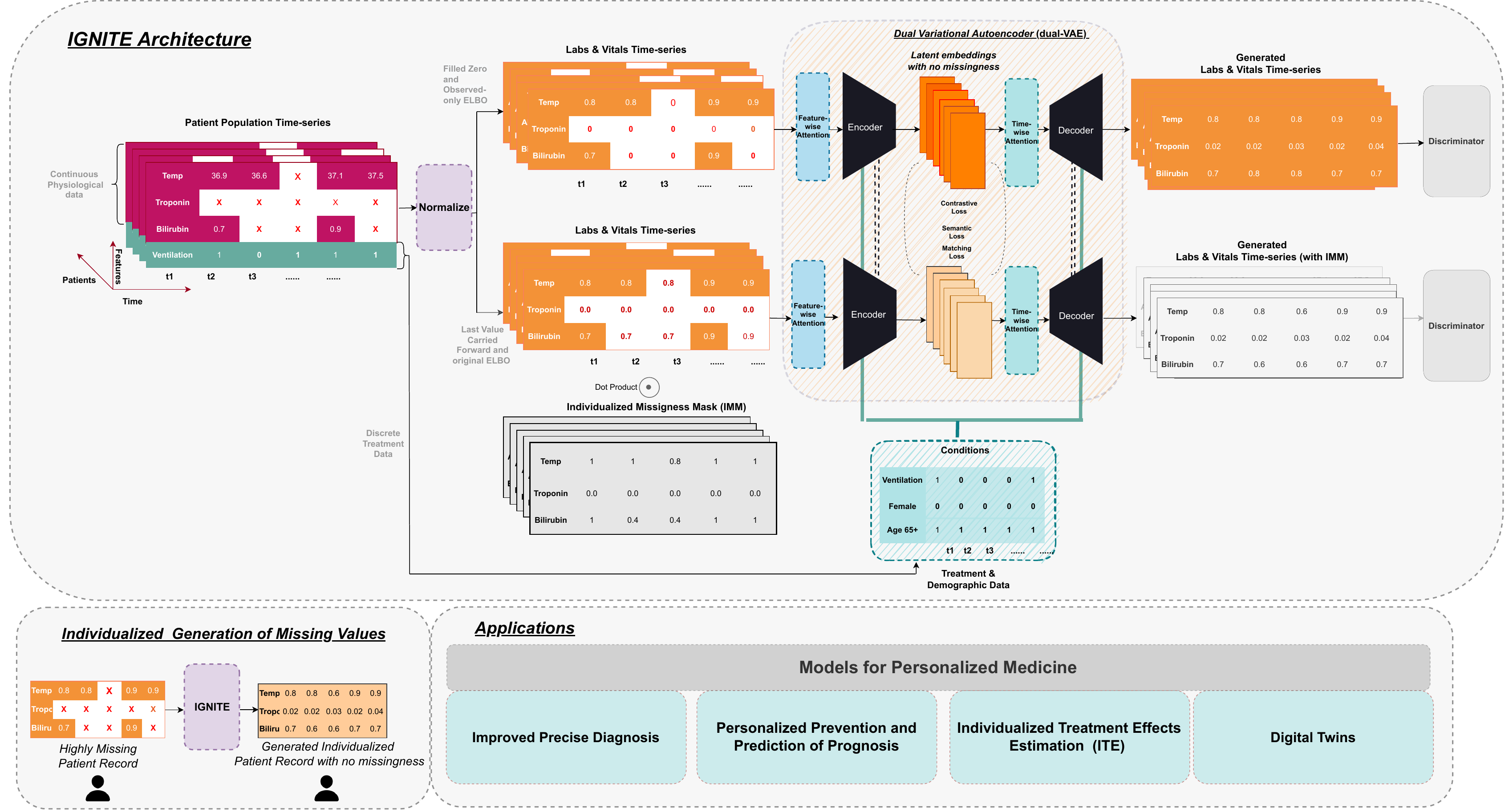}
\end{center}
\caption{An overview of the architecture and applications of our proposed model, IGNITE,  for generating individualized time-series EHRs. The evidence lower bound (ELBO) is calculated for the observed value only for the upper VAE and for the augmented data from the full individualized missingness mask (IMM) in the lower VAE. By utilizing treatment data and individualized missingness patterns, IGNITE is capable of generating EHRs that facilitate various applications of personalized medicine.}
  \label{fig:overview}
\end{figure*}

To this end, we propose a novel end-to-end generative model for (\textbf{I}ndividualized \textbf{G}e\textbf{N}eration of \textbf{I}mputations in \textbf{T}ime-series \textbf{E}lectronic health records (denoted as IGNITE). \textcolor{black}{IGNITE generates a complete patient record taking into account differences in physiological data, treatment strategies as well as individualized missingness patterns. Adopting a personalized approach to missingness generation, IGNITE could pave the way for creating digital twins for precision medicine, allowing real-time health monitoring, risk phenotyping and prediction, and treatment strategies at an individual level. We view that deep generative models present a great opportunity not only for imputing values missing at random but also for generating new features that were never measured for the patient. For this purpose, we bring new insights into missingness as a generative task for generating samples and even complete features for patients without any observations of the respective features, which we refer to as feature-wise missingness. By combining these elements, IGNITE not only addresses the persistent challenges of incompleteness and variability of data in EHRs, but also provides more accurate and personalized risk assessment and treatment strategies, which contributes significantly to the improvement of patient outcomes and the optimization of clinical workflows}. An overview of our model is shown in Figure~\ref{fig:overview}. Our main contributions are as follows:
\begin{enumerate}
    \item \textbf{Individualized Time-series Imputation Model}: We propose a generative model comprising dual-variational autoencoders that can impute any personalized missing values in time-series EHR conditioned on discrete treatment (intervention) data and patient characteristics.  Our model also includes dual-stage attention networks to improve representation learning across both the feature dimension and the temporal dimension.
    
    \item \textbf{Personalized Missingness Mask}: We introduce a novel individualized missingness mask (IMM) that accounts for individual-level differences in missingness frequencies and patterns across time and feature dimensions. This mask is used to augment the input data to one of the dual-VAEs to better generate personalized imputations for missing EHR values.

    \item {\textbf{New Missingness Evaluation Framework.}} We propose a framework for investigating the performance of the imputation models across sub-populations with various frequencies of sample and feature missingness. Our evaluation framework aims to demonstrate the robustness of these models to various missingness patterns, especially when complete features were never observed. 
    \item  \textbf{Benchmarking on real-world data.} We perform rigorous evaluations of models on three widely used publicly available Intensive Care Unit (ICU) datasets. 
\end{enumerate}

\section*{Results}

\subsection{Downstream Task}

\textcolor{black}{To evaluate the performance of different imputation models, we used the imputed data in a downstream task --- mortality prediction, on each of the three datasets: PhysioNet 2012, HiRID and eICU. In all models, we first imputed the data and then used the imputed data for evaluation in the respective downstream task. }The higher the predictive performance of the model, the better the imputation of the data. In the mortality prediction task for the PhysioNet 2012 dataset, our IGNITE performed the best with an area under the receiving operating curve (AUROC) of 0.834, followed by TimesNet and transformer AUROC of 0.831 and 0.827, respectively. Models trained on the LOCF-imputed method were ranked the worst with an AUROC of 0.777. The various baseline methods were also evaluated in terms of area under the precision-recall curve (AUPRC), which represents discrimination with respect to class imbalance. 
The results were also consistent on the eICU and HiRID datasets, where IGNITE consistently outperformed all other benchmarks. For instance, IGNITE achieved an AUROC of 0.822 in eICU and 0.968 in HiRID, respectively. A similar trend was observed in the AUPRC results for both datasets. It is worth noting that the eICU imputations generated by IGNITE significantly improved the performance with an increase of more than 10\% and 20\% when compared to the next best or worst models, respectively. \textcolor{black}{We also implemented Wilcoxon Signed-Ranked test to assess for the statistical significance and show the results }.




\begin{table*}[!htb]
    \centering
      \caption{Performance for a mortality prediction task using an LSTM model reported in terms of AUROC and AURPC. $\ast$ indicates that IGNITE is statistically significant ($p<0.05$) when compared to the corresponding benchmark.} 
        \resizebox{0.9\linewidth}{!}{
    \begin{tabular}{lcccccc}
    \toprule
         \textbf{Imputation method} & \multicolumn{3}{c}{\textbf{AUROC}}& \multicolumn{3}{c}{\textbf{AUPRC}}\\

          \cmidrule(lr){2-4}\cmidrule(lr){5-7}
       & PhysioNet 2012 & eICU & HiRID  & PhysioNet 2012 & eICU & HiRID\\

     \midrule
    LOCF & 0.777 (0.014)* & 0.736 (0.001)*  & 0.935 (0.007)* &0.389 (0.025)* & 0.316 (0.015)* & 0.593 (0.031)*\\
    MICE & 0.823 (0.009)* & 0.765 (0.014)*  & 0.939 (0.007)* &0.469 (0.025)*& 0.289 (0.034)* & 0.563 (0.023)*\\
    GP-VAE &0.780 (0.013)*  & 0.798 (0.009)*  & 0.948 (0.006)* &0.371 (0.024)*  & 0.397 (0.036)* & 0.583 (0.025)* \\ 
    Transformer &0.831 (0.010)* & 0.729 (0.008)*  & 0.961 (0.004)* & \textbf{0.477 (0.028)}*& 0.249 (0.017)* & 0.658 (0.022)*\\
    BRITS & 0.818 (0.010)*  & 0.735 (0.001)*  & 0.959 (0.005)* & 0.452 (0.025)*  & 0.318 (0.015)* & 0.641 (0.027)* \\
    TimesNet &0.827 (0.001)*  &0.750 (0.010)*  &0.956 (0.004)*&0.352 (0.014)*&	0.305 (0.014)* &0.640 (0.022)*\\
    SAITS &0.821 (0.010)*  & 0.608 (0.007)*  & 0.961 (0.005)* & 0.467 (0.035)* & 0.115 (0.008)* & 0.655 (0.030)* \\
    IGNITE & \textbf{0.834 (0.009)} & \textbf{0.822 (0.008)} & \textbf{0.968 (0.004)} &0.458 (0.032)& \textbf{0.345 (0.014)} & \textbf{0.723 (0.030)}\\
    \bottomrule
    \end{tabular}
  }
    \label{tab:performance}
\end{table*}

The baseline imputation methods were also evaluated in terms of performance with varied percentages and types of missingness for feature-wise and sample-wise missingness, respectively. The three datasets were stratified based on the percentage of overall feature-wise missingness, where $z \%$ of features were never observed. As shown in Table~\ref{tab:percent_missing3}, we note that for the PhysioNet 2012 dataset, only $\leq$ 25\% and 25\% - 75\% missingness were evaluated, as there were not enough training samples with $\geq$ 75\% missingness. IGNITE achieved the highest performance with an AUROC of 0.790, an AUPRC of 0.561 for $\leq$ 25\%, and 0.791 and 0.346 in terms of AUROC and AUPRC in the 25\% - 75\% task, respectively.  Similarly, IGNITE consistently outperformed the other baselines in assessing AUPRC for eICU in the population subset with $\leq$ 25\% of feature-wise missingness. For the population with feature-wise missingness of  25\%- 75\% for eICU, SAITS achieved the highest performance, followed by IGNITE. For HiRID, on the other hand, the highest performance of the population with feature-wise missingness of  25\%- 75\% was achieved by the model trained on the data imputed by the Transformer model in terms of AUROC, in terms of AURPC, IGNITE achieved the highest performance. As most of the population has < 75\% feature-wise missingness, HiRID and PhysioNet 2012 were not used for the experiments with > 75\% missingness. For eICU, IGNITE achieved the highest performance, followed by SAITS in terms of AUROC for the population with > 75\% missingness. We note that the performance of the GP-VAE, MICE, and LOCF imputation ranked last across multiple percentages of missingness, indicating the lack of robustness. 
\begin{table*}[!ht]
    \centering
      \caption{Performance for a mortality prediction task for patients with at least n\% of features never measured (completely missing)}
        \resizebox{\linewidth}{!}{

       
          \begin{tabular}{lccccccccc}
    \toprule
     \textbf{\% Feature-wise Missingness} &  \multicolumn{3}{c}{\textbf{$\leq$ 25\%}}& \multicolumn{3}{c}{\textbf{25-75\%}} & \multicolumn{3}{c}{\textbf{$\geq$ 75\%}}\\
        \cmidrule(lr){2-4}\cmidrule(lr){5-7} \cmidrule(lr){8-10}
       & PhysioNet 2012 & eICU & HiRID  & PhysioNet 2012 & eICU & HiRID & PhysioNet 2012 & eICU & HiRID\\
       
    \midrule
    \textbf{Population Description}  &&&\\
    \midrule
     Population Size (n) & 4434 & 6213 & 4513 
     & 7604 & 45,845 & 2913
     & NA & 2329 & NA\\
     Positive Outcome (n, \%) & 982 (22.7\%) & 1342 (21.6\%) & 574 (12.7\%)  
     & 723 (9.51\%) & 1614 (3.52\%) &157 (5.39\%)
     & NA & 131 (5.62\%) & NA\\
     \midrule

      \textbf{AUROC} &&&\\
     \midrule
    LOCF &0.710& 0.832 & 0.913  
         &0.788& 0.658 & 0.919
         &NA & 0.587 &  NA \\
    MICE &0.769& 0.754 & 0.834 
         &0.703& 0.650 & 0.907
         & NA & 0.651 & NA\\
    GP-VAE &0.705& 0.817 &\textbf{0.957}&0.746& 0.697 &0.947&NA& 0.634 &NA\\
   Transformer &0.764& 0.887 & 0.924  
   &\textbf{0.820}& 0.699 & \textbf{0.973}   
   &NA & 0.690 & NA\\
    BRITS &0.757& 0.866 & 0.932
          &0.817& 0.682 &  0.939
          &NA & 0.697 & NA\\
    TimesNet &0.717&0.811&0.922&0.745&0.733&0.935&NA&0.719&NA\\

    SAITS &0.759& 0.900 & 0.925
          &0.798& \textbf{0.728} & 0.921
          &NA & 0.701 & NA\\
    IGNITE &\textbf{0.790}& \textbf{0.920} &  0.952
           &0.791& 0.705 & 0.961
           &NA & \textbf{0.751} & NA\\
     \midrule
    \textbf{AUPRC} &&&\\
     \midrule
    LOCF &0.386& 0.688 & 0.724
         &0.297& 0.123 & 0.565
         &NA & 0.207 & NA\\
    MICE & 0.543& 0.590 & 0.695
         &0.273 & 0.121 & 0.798
         &NA & 0.201 & NA\\
    GP-VAE &0.427& 0.643 &0.806&0.271& 0.159 &0.681&NA &  0.233 &NA\\
    Transformer &0.495& 0.734 & 0.811
                &0.326& 0.235 & 0.807
                &NA & 0.273 & NA\\
    BRITS &0.501& 0.724 & 0.847
          &0.328& 0.219 & 0.828
          &NA & 0.290 & NA\\
    TimesNet &0.454&0.680&0.841&0.336&0.295&0.834&NA&0.280&NA\\

    SAITS &0.514& 0.782  & 0.852
          &0.290& \textbf{0.282} & 0.825
          &NA & \textbf{0.367} & NA\\

    IGNITE &\textbf{0.561} & \textbf{0.807} & \textbf{0.870}
           &\textbf{0.346}& 0.261 & \textbf{0.863}
           & NA & 0.321 & NA\\
    \bottomrule
    \end{tabular}
  }
  				
    \label{tab:percent_missing3}
\end{table*}
	
To evaluate the population for varying sample-wise missingness, where values are not observed across time and features, 
we stratified the dataset into three groups $\leq$ 25\%,  25\%- 75\%, and $\geq$ 75\% sample-wise missingness. However, in all three datasets, no patients had $\leq$ 25\% sample-wise missingness, indicating the high missingness in EHRs. The high prevalence of missingness in all the included datasets resulted in two main subgroups for 25\%- 75\%, and $\geq$ 75\% sample-wise missingness for PhysioNet 2012 and HiRID. On the other hand, more than 95\% of the samples in the eICU dataset had more than 75\% sample-wise missingness, resulting in one subgroup for eICU. In terms of performance, IGNITE consistently outperformed all baselines in terms of AUROC in all three datasets. Specifically, IGNITE achieved the highest performance of 0.816, followed by BRITS for the 25\%- 75\% group in the PhysioNet 2012 dataset. For AURPC, IGNITE achieved the second-highest performance with 0.419, where the highest performance was achieved by GP-VAE. Similar trends were observed in the eICU and HiRID datasets, with an AUROC reaching 0.959 for the HiRID dataset for samples with > 75\% sample-wise missingness. Although SAITS and Transformer showed high performance across all population subsets, the performance deteriorated for the population with low sample-wise missingness. 
We also observed that mortality outcome was less prevalent in patients with higher overall sample-wise missingness, 13.4\% when compared to those with lower sample-wise missingness 25.1\% in the PhysioNet 2012 dataset. Similar trends were also observed in the HiRID dataset, where the outcome prevalence decreased from 23.9\% to 5.43\% for the samples with high sample-wise missingness.  

\begin{table*}[!htb]
    \centering
      \caption{Performance for a mortality prediction task for patients with at varying sample-wise missingness. The results are reported in terms of AUROC and AUPRC, respectively.}
        \resizebox{\linewidth}{!}{
    \begin{tabular}{lcccccccccccccccccc}
    \toprule
     \textbf{\% Sample-wise Missingness}& \multicolumn{3}{c}{\textbf{25-75\%}} & \multicolumn{3}{c}{\textbf{$\geq$ 75\%}}\\
        \cmidrule(lr){2-4}\cmidrule(lr){5-7} \cmidrule(lr){8-10}
    & PhysioNet 2012 & eICU & HiRID & PhysioNet 2012 & eICU & HiRID\\
     \midrule

    \textbf{Population Description}  &&&\\
          \midrule
     Population Size (n) &825&NA&1768&11175&54,370& 5709\\
    Positive Outcome (n, \%) &207 (25.1\%) &NA&422 (23.9\%)&1500 (13.4\%) & 3071 (5.65\%) & 310 (5.43\%)\\
    \midrule
      \textbf{AUROC} &&&\\
     \midrule
    LOCF &0.762&NA& 0.850  &0.754 & 0.622 & 0.912
\\
    MICE &0.786&NA& 0.815  &0.798 & 0.604 & 0.791\\
    GP-VAE &0.727&NA& 0.833 & 0.754 & 0.671 & 0.937\\
    Transformer &0.737&NA&  0.855  &0.808 & 0.640 & 0.935\\
    BRITS &0.804&NA&  0.831   & 0.801 & 0.680 & 0.913\\
     TimesNet &0.761&NA &0.837&0.723&0.689&0.906\\

    SAITS &0.770&NA&  0.826 & 0.815 & 0.662 & 0.929\\

    IGNITE &\textbf{0.816}&NA&  \textbf{0.861} &\textbf{0.826} & \textbf{0.747} & \textbf{0.959}\\
     \midrule
    \textbf{AUPRC} &&&\\
     \midrule
    LOCF &0.465&NA& 0.716  &0.362 & 0.199 & 0.582\\
    MICE &0.592&NA&  0.649 &0.379 & 0.218 & 0.629\\
    GP-VAE &0.477&NA& 0.610 &0.334 & 0.258 & 0.647\\
    Transformer &0.529&NA&  0.771  &\textbf{0.429} & 0.244 & 0.831\\

    BRITS &0.622&NA&  0.785  & 0.421 & 0.269 & 0.810\\
     TimesNet &0.500&NA &0.781&0.294&0.254 &0.793\\

    SAITS & 0.533&NA& 0.737  & 0.420 & 0.224 & 0.790\\

    IGNITE &\textbf{0.626}&NA& \textbf{0.796} &0.419 & \textbf{0.302} & \textbf{0.823} \\

    \bottomrule
    \end{tabular}
  }
    \label{tab:percent_missing9}
\end{table*}

\subsection{Reconstruction Task}
\begin{figure}[h!]
	\centering
	\subfigure[{Imputation for a patient with sample-wise missingness.}]{\includegraphics[width=0.4\textwidth]{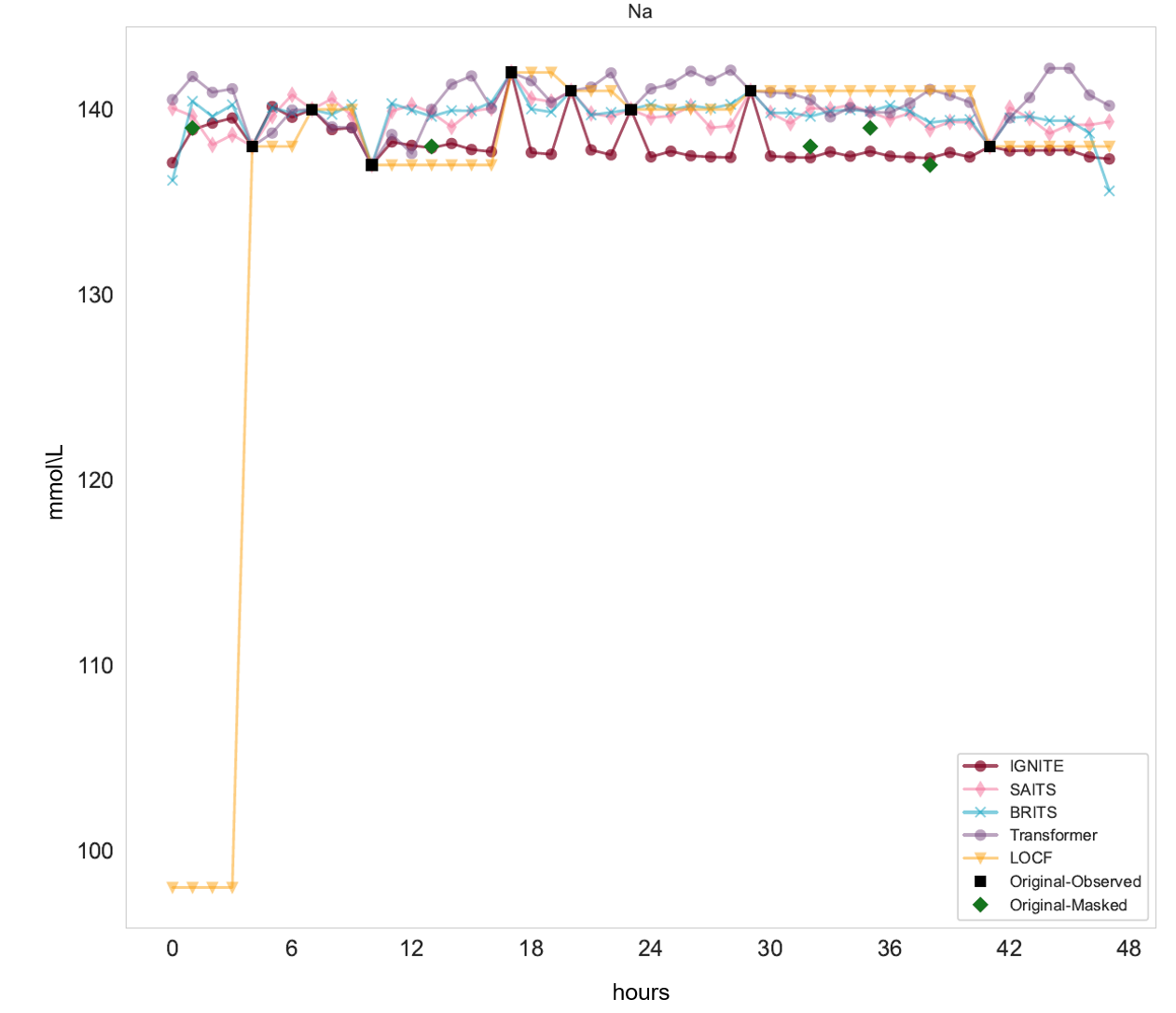}
		\label{Colab}}
  \vspace{-0.2cm}
	\subfigure[{Imputation for a patient with feature-wise missingness.}]{\includegraphics[width=0.4\textwidth]{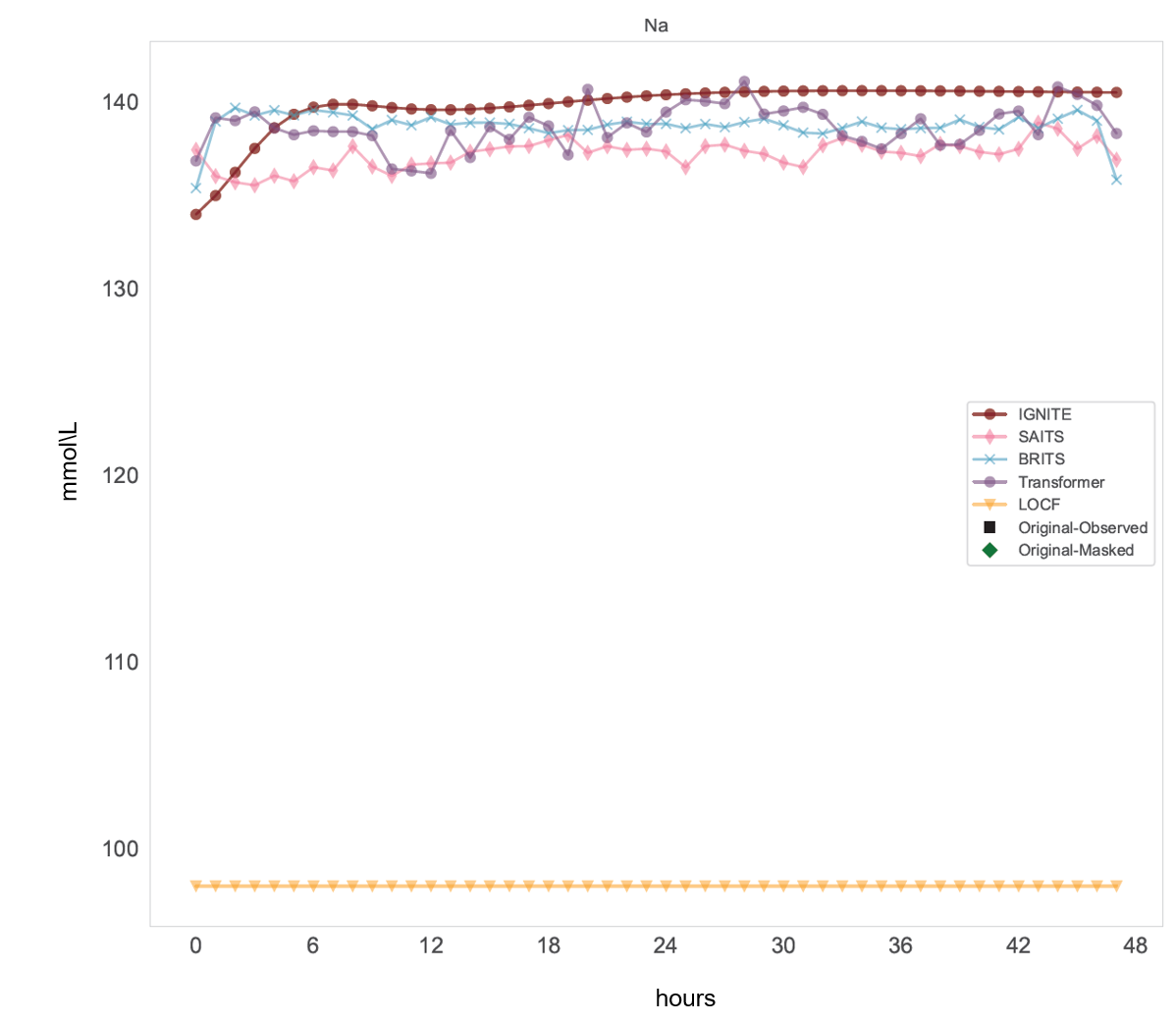}}

	\caption{Visualized imputations for patients from PhysioNet 2012 dataset. In a and b, we show examples of patients with different types of missingness.  The original observed values are shown as black. In a, we show an example of a patient with sample-wise missingness, while in b, we show a patient with feature-wise missingness indicating no observed measurements for that feature across all time-steps. We further masked 50\% of the observed values and considered them to be the ground truth as shown in green. Various imputation methods are compared with respect to ground truth. }

	\label{Figure7}
\end{figure}

\begin{figure}[h!]
\centering

	\subfigure[{Urine Output}]{\includegraphics[width=0.3\textwidth]{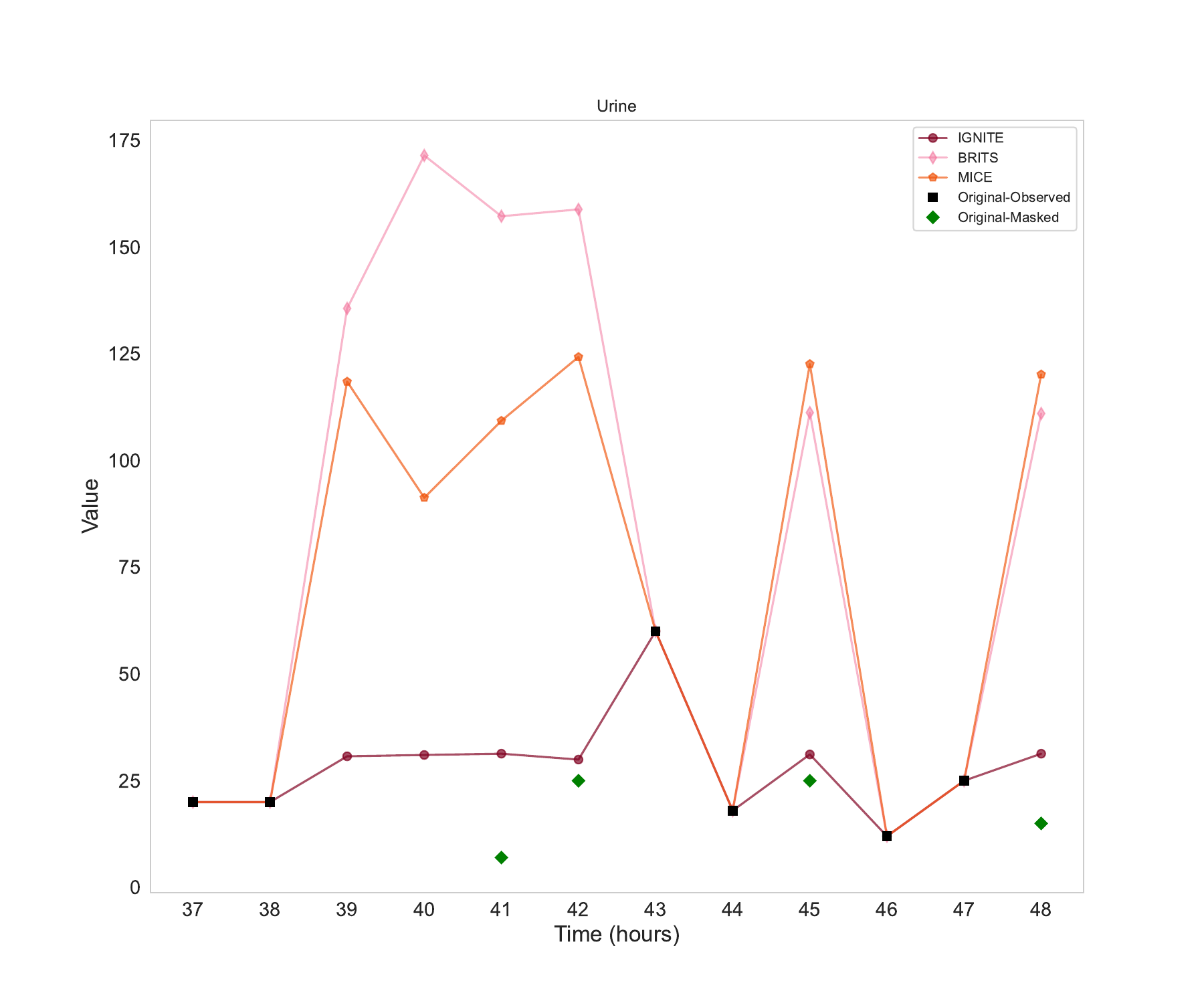}}
	\subfigure[{FiO2}]{\includegraphics[width=0.3\textwidth]{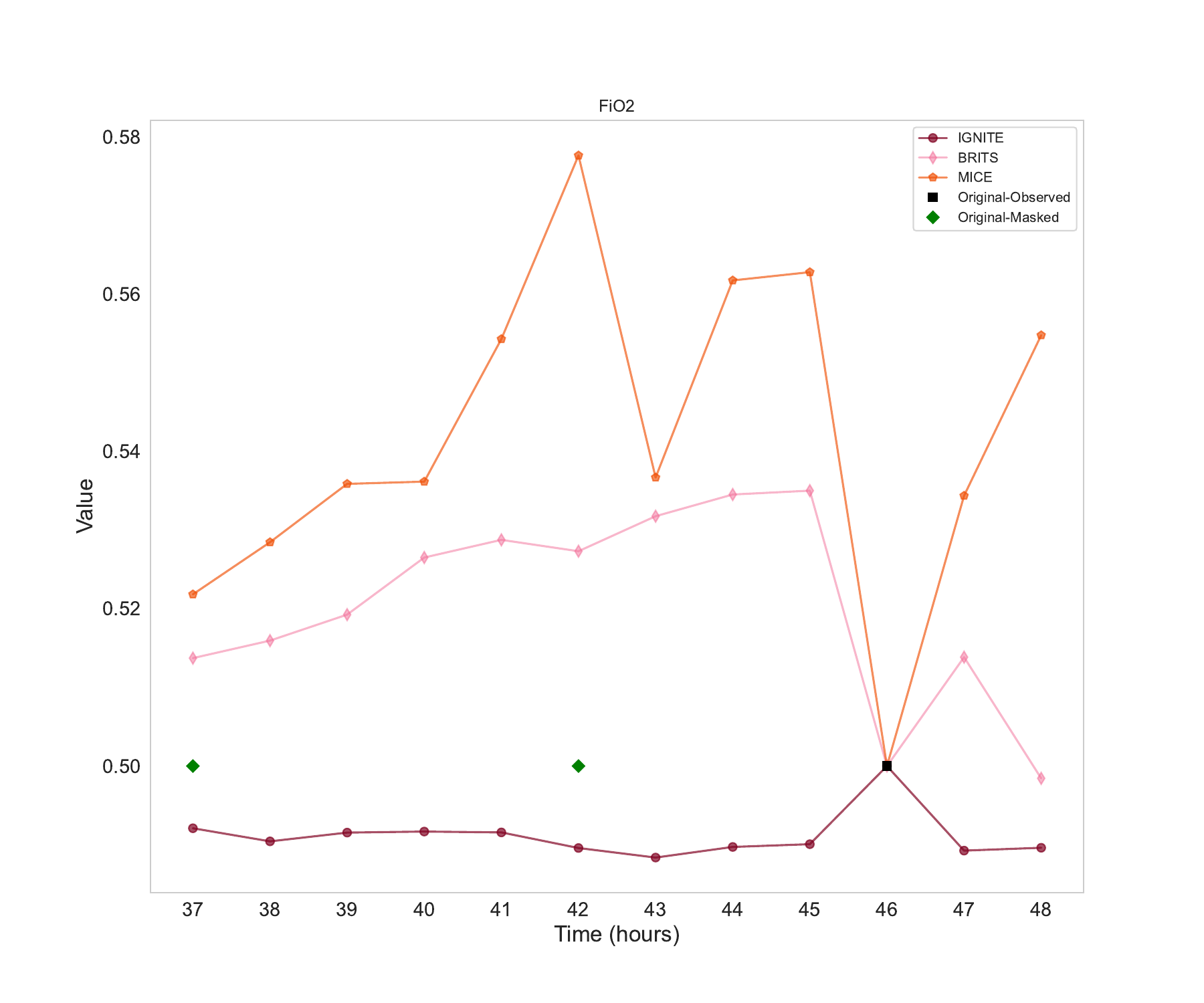}}
 	\subfigure[{HCO3}]{\includegraphics[width=0.3\textwidth]{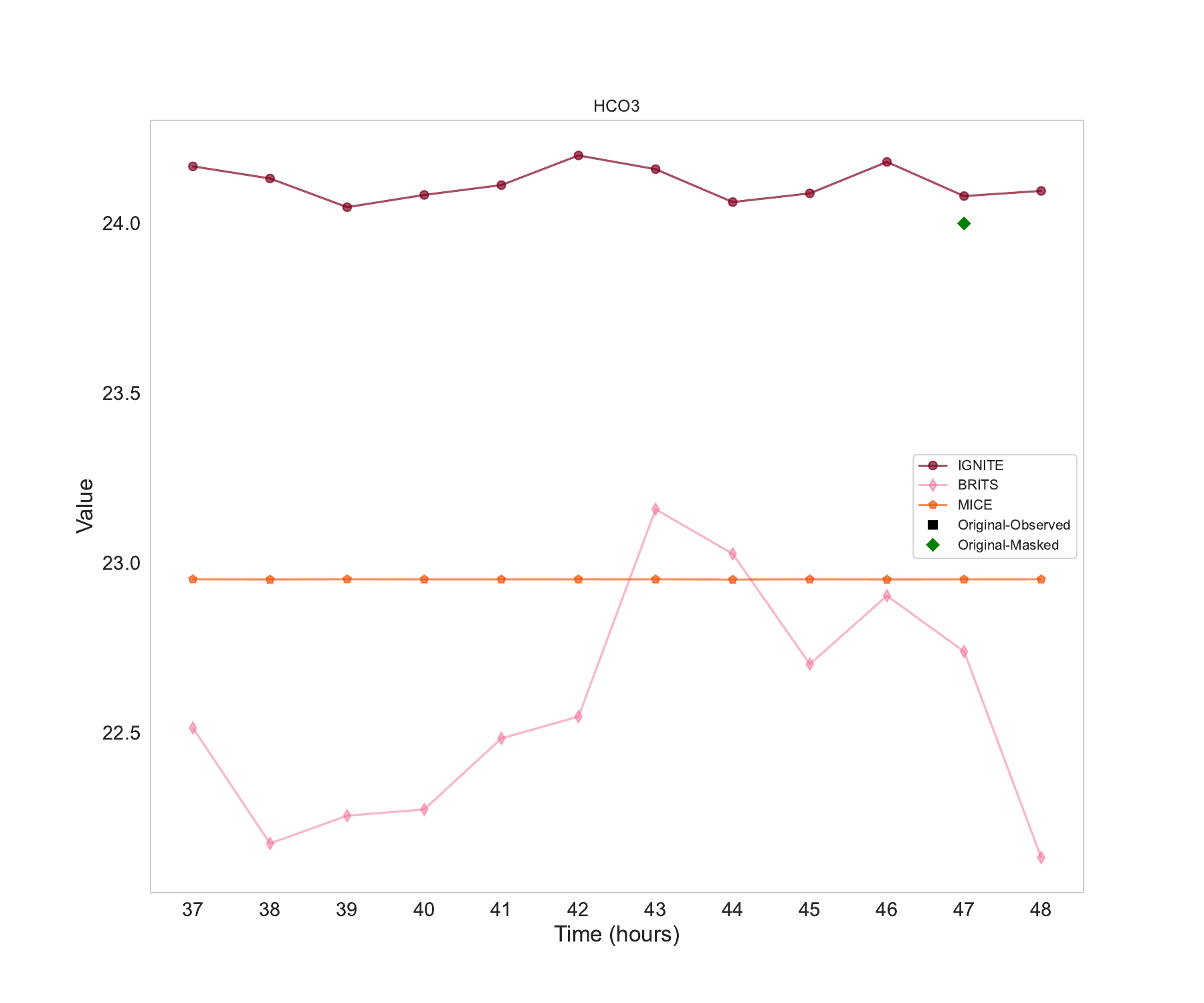}}

      \caption{\textcolor{black}{Visualized imputations for a dead patient, where we compare IGNITE to two models, namely BRITS and MICE.  The original observed values are shown as black and the masked values that are considered ground truth are shown as green. We present the rest of the imputed variables for this patient in the Supplementary Information \ref{dead}.}}
      \label{dead_f}
 \end{figure}

Other than measuring the performance of IGNITE in predictive modelling tasks, we also evaluate its ability to recover and reconstruct randomly introduced missing values (i.e., missing at completely random). In the reconstruction experiments, IGNITE consistently outperformed the other baselines across the two metrics, Root-Mean-Square Error (RMSE) and Mean Absolute Error (MAE) for each patient. The metrics shown in Table~\ref{mae}, are reported in the mean and standard error of the errors across all the patients in the test set. Specifically, IGNITE had an RMSE of 0.100 (0.038), followed by MICE with 0.109 (0.059) for the setting where 10\% missingness was introduced in the PhysioNet 2012 data. Similar trends were observed in the 20\% and 50\% experiments, where IGNITE achieved the lowest reconstruction error and LOCF achieved the highest error.  It is worth noting that the error in IGNITE generally remained stable despite increasing the percentage of missingness, indicating its robustness in reconstructing data with high missingness. We showcase the full results for the reconstruction experiments on HiRID and eICU datasets in the Supplementary materials~\ref{reconst}, where IGNITE consistently outperformed all the baselines. 
\textcolor{black}{In addition to its methodological contributions and high performance in reconstruction tasks, IGNITE  significantly boosts prediction accuracy by improving how imputed data align with actual patient statuses — a crucial factor in effective clinical decision-making. By accurately imputing missing data in a way that reflects real-time changes in patient health, IGNITE preserves essential temporal trends and patient-specific dynamics within the EHR data. We show example patient imputations with different patient states in sample visualization of the reconstructed values via top-performing models shown in Figure \ref{Figure7}. In Figure~\ref{Figure7}a and Figure~\ref{Figure7}b, we show the imputations generated by the various models for two patients with different types of missingness ie: with feature-wise and sample-wise missingenss, where IGNITE was the model with the lower error and best at recovering masked values. In Figure~\ref{Figure7}b, we see that IGNITE and the other deep learning models generated realistic values. However, LOCF imputation fails as there are no observed values. The other methods' reconstructions are shown in Supplementary materials~\ref{vis}}.
\textcolor{black}{Furthermore, In Figure\ref{dead_f}, We present a comparison of imputed values using IGNITE against BRITS and MICE both with strong performance. In this section, we showcase imputed values for three key variables, with additional results provided in the Supplementary Material \ref{dead}. Our results demonstrate that IGNITE is the only model that imputes urine output within the true range and trend, as seen in a case where the patient's urine output decreases prior to death. In contrast, BRITS and MICE show an increasing trend, far from the ground truth. A similar observation is made for HCO3, where IGNITE is the only model to predict values close to the ground truth. For FiO2, typically flat around 0.5 for patients on mechanical ventilation, IGNITE accurately captures this, providing realistic imputations. These examples highlight that IGNITE not only improves predictive performance but also generates clinically realistic imputations.}

\begin{table}[!ht]
    \centering
          \caption{Performance of the various baselines of the reconstruction task on the PhysioNet 2012 dataset. The results are calculated across all the patients in the test set and reported in terms of RMSE and MAE with mean and standard deviation. $\ast$ indicates that IGNITE is statistically significant ($p<0.05$) when compared to the corresponding benchmark.}
                  \resizebox{\linewidth}{!}{

    \begin{tabular}{lccccccc}
    \toprule
     & \multicolumn{3}{c}{\textbf{RMSE}}&\multicolumn{3}{c}{\textbf{MAE}}
   \\
   \midrule
     \textbf{Introduced Missingness}& \textbf{10\%}&  \textbf{20\%}&\textbf{50\%} & \textbf{10\%} &  \textbf{20\%}&\textbf{50\%} \\
   \midrule
       LOCF & 0.183 (0.065)* & 0.192 (0.051)* & 0.202 (0.036)* & 0.106 (0.040)* & 0.108 (0.034)* & 0.115 (0.028)* \\ 
    
        MICE & 0.109 (0.046)* & 0.112 (0.037) & 0.122 (0.034) *& 0.066 (0.023) & 0.066 (0.019) & 0.066 (0.017) \\
        GP-VAE &  0.395 (0.067)* & 0.398 (0.521)* & 0.398 (0.044)* &0.317 (0.056) *& 0.318 (0.045)* & 0.317 (0.038)* \\ 
        Transformer & 0.119 (0.059)* & 0.128 (0.05)* & 0.123 (0.039)* & 0.068 (0.028) & 0.07 (0.025)* & 0.068 (0.021)* \\ 
        BRITS & 0.115 (0.054)* & 0.116 (0.044)* & 0.118 (0.038)* & 0.066 (0.027) & 0.066 (0.022) & 0.065 (0.02) \\ 
        TimesNet & 0.266 (0.158)* &0.263 (0.045)*& 0.266 (0.144)* &0.071 (0.026)* &0.069 (0.021)* & 0.079* (0.027
        )\\
        SAITS & 0.122 (0.058)* & 0.127 (0.051)* & 0.123 (0.038)* & 0.071 (0.029)* & 0.069 (0.025)* & 0.067 (0.021) \\ 
        IGNITE & \textbf{0.100 (0.038)} & \textbf{0.103 (0.032)} & \textbf{0.104 (0.027)} &\textbf{ 0.062 (0.021)} & \textbf{0.063 (0.018)} &\textbf{ 0.063 (0.016)} \\ 
        \bottomrule
    \end{tabular}}
    \label{mae}
\end{table}

\subsection{Ablation Study}
To demonstrate the added value of each of the model's components, we conducted an ablation study on the three datasets The studied components were mainly the time-series conditional component as well as the IMM mask used in IGNITE. \textcolor{black}{As presented in Table \ref{ablate}, we observe consistent performance gains in terms of AUROC as new components are added across the three datasets. The highest performance gain was attributed to adding IMM to the model, where the final model achieved 0.835, 0.774 and 0,968 of AUROC for PhysioNet 2012, eICU and HiRID respectively.}
\begin{table}[!ht]
    \centering
      \caption{Performance when ablating various components of the proposed models, reported in terms of AUROC}
        \resizebox{0.7\linewidth}{!}{
    \begin{tabular}{lcccc}
    \toprule
     \textbf{Ablations}  & \textbf{PhysioNet 2012} & \textbf{eICU} & \textbf{HiRID}\\
     \midrule
      IGNITE (Dual VAE, No Condition) & 0.818 & 0.746 & 0.927\\
      IGNITE (Dual VAE, Condition) & 0.823 & 0.754 & 0.933 \\
      IGNITE (Dual VAE, Condition, IMM) & 0.830 & 0.762 &0.952 \\
     IGNITE (Dual VAE, Condition, IMM, MIT loss) & 0.832 & 0.765 & 0.965\\
     IGNITE (Dual VAE, Condition, IMM, MIT loss, discriminator) & \textbf{0.835} & \textbf{0.774} & \textbf{0.968} \\

    \bottomrule
    \end{tabular}
  }
    \label{ablate}
\end{table}


\section*{Discussion}
In this work, we proposed a deep end-to-end generative model that synthesizes personalized data in highly sparse and irregularly sampled or even completely missing EHR conditioned on treatments and demographic characteristics, as well as the individual level of missingness patterns. We note that our model, IGNITE, outperformed the baseline models in the full population downstream mortality prediction task in terms of AUROC and AUPRC. 
Unlike previous works, we proposed a new framework for the evaluation of imputation models trained on various types of missingness across features and samples in time-series data. In the proposed downstream task evaluation, IGNITE showed consistent robustness despite varying the percentage of feature-wise and sample-wise missingness. By augmenting our training with a novel individualized missingness mask, we believe that IGNITE captured better dynamics and underlying indicators of the patient's health status.  \textcolor{black}{Unlike traditional imputation methods that often assume a specific missingness mechanism, IGNITE is designed to be adaptable to various types of missing data scenarios commonly encountered in healthcare datasets. This adaptability improves the utility of the model in a wide range of clinical applications beyond the datasets used in this study.}

This paper has several strengths. First, this is the first work to introduce a novel missingness mask that represents measurement patterns and frequencies in an individualized way. The proposed IMM mask showed high utility in imputation tasks, we believe such masks can be potentially utilized in other diverse time-series tasks such as prediction~\cite{han2019review}, detection~\cite{blazquez2021review}, clustering~\cite{aghabozorgi2015time} and more. Second, the proposed work is the first to investigate feature-wise missingness and sample-wise missingness in observed healthcare records, bringing a new understanding of missingness patterns in real-world data. Although several works investigated theoretical missingness assumptions such as MCAR, MAR and MNAR~\cite{curran1998identifying}, such missingness mechanisms are hard to prove and remain far from patterns observed in real-world health data.  We found a correlation between the prevalence of positive outcomes and sampling frequency by investigating the percentage of outcome prevalence across populations with features that were never observed and those with different overall sample missingness. There was a significantly higher percentage of positive (mortality) outcomes in patients with $\leq$25\%  of features never observed across all three datasets, where the prevalence of outcomes increased more than double compared to those with higher feature-wise missingness. Similar trends were also observed in sample-wise missingness, suggesting that the missingness pattern is personalized and could inform the underlying patient's health status for various personalized medical applications. Imputation methods that consider individual missingness patterns, such as IGNITE, could lead to more accurate predictions and tailored treatments. Moreover, our feature-wise missingness experiments showcased IGNITE's power to generate features that were never observed in patients. In traditional statistical approaches, many patients with such never-observed features are generally omitted/removed, leading to a waste of data or a reduction in model performance.  However, with IGNITE, such features can be generated based on other readily available information about an individual. We believe that a flexible generative approach proposed by IGNITE can open doors for a new way of rethinking missingness imputation as a generative task, making various new applications utilizing generative models.

Another strength of our proposed work is that IGNITE is designed to generate imputations conditioned on individualized patient observed and missingness patterns and dynamics, making the generated imputations of high quality and more aligned with personalized medicine applications. Generating digital twins to drive personalized medicine applications requires understanding and representing patient data beyond observed values, which IGNITE incorporates into its generative process. Lastly, IGNITE was trained and tested across three large-scale EHRs, outperforming state-of-the-art imputation models across a series of experiments. Other experiments, such as the reconstruction experiments, demonstrate IGNITE's robustness and ability to recover original patient values even when random missingness is introduced to patient records. 
This robust performance across datasets and experimental settings showcases IGNITE's robustness in learning individualized representations and generating high-quality imputations. 


Overall, this study has some limitations. IGNITE and the other imputation models were tested only on retrospective ICU datasets. In future work, we plan to investigate IGNITE's ability to learn individualized patient representations with higher sparsity and missingness patterns in primary care and wearable data.  Furthermore, while our work expands on the evaluation metrics proposed in the literature, IGNITE was only evaluated in prediction tasks involving mortality as limited by the available datasets such as Physionet 2012. In future work, we aim to investigate the impact of imputations on various tasks in healthcare settings, such as treatment recommendation and phenotyping. Another area of future work involves the theoretical analysis of the various missingness patterns in the data to better understand the strengths and limitations of the plethora of time-series techniques. \textcolor{black}{In addition, the datasets selected for the evaluation of the IGNITE model were chosen based on their open-access nature to ensure that our results are transparent and reproducible. These ICU datasets are commonly used in healthcare analytics research, allowing us to benchmark our results against well-established methods in the field. Although the current study focuses on ICU datasets due to the availability and richness of their clinical data, we recognize that this may limit the perceived applicability of our method. However, it is important to note that the principles underlying the IGNITE model are broadly applicable to other types of clinical data. Future research will aim to apply IGNITE to additional datasets from diverse medical settings, demonstrating its broader applicability and exploring its potential in other areas of clinical practice. This expansion will also allow us to test the model's robustness across different healthcare environments, addressing any unique challenges they present, and supporting the reproducibility of our findings. Another potential area of future work is to include more benchmarks that might be relevant in medical statistics such as \cite{zhang2020predicting,sun2019mice}. For the sake of this work, we selected traditional statistical methods such as MICE and LOCF alongside advanced deep learning models, to provide a comprehensive perspective on imputation effectiveness. This selection strategy ensures a robust comparison across different methodological approaches, demonstrating the advances deep learning offers over traditional benchmarks, particularly when learning individualized imputations. Lastly, while deep learning models like IGNITE require more computational resources during training compared to traditional methods, they significantly expedite the inference process. In our analysis, the improvements in data quality and model performance using IGNITE are statistically significant, validating the trade-off between increased training time and enhanced operational efficiency in practical applications.}

\section*{Methods}
\subsection{\textbf{Notations}}
We denote a multivariate time-series EHRs dataset as $\mathbf{D}=\left\{\left(\mathbf{x}_{i, 1: T_{i}}\right)\right\}_{i=1}^{N}$, which includes a set of individual patient records indexed by $i\in\{1,2, \ldots N\}$.  Each patient record comprises two main components: time-series and static components. The time-series component includes continuous time-series of physiological data and the discrete time-series of treatment data, recorded over time steps $\mathbf{T}=(\mathbf{t}_{1}, \ldots, {t}_{N})$. Each record has a corresponding static outcome label $ \mathbf{y}=(\mathbf{y}_{1}, \ldots, {y}_{N})$ and static demographic information, indicating the age and sex of the patient. For each individual patient time-series matrix, we extract a binary mask indicating the missingness of each value, where 0 and 1 denote the missing value and the observed value, respectively.

\subsection{Proposed Model}
\subsubsection{\textbf{Dual Variational-AutoEncoders with Dual-Stage Attention}}
We utilize a pair of VAEs, each with an encoder and a decoder, to map the multi-variate time-series into reversible low-dimensional dense representations with no missingness. We use long short-term memory (LSTM)~\cite{hochreiter1997long} neural networks for the architecture of VAE. The first VAE is trained to reconstruct the multivariate time-series by calculating the evidence lower bound (ELBO) on the observed values only and imputing the missing values with zero, an approach adopted by \citep{nazabal2020handling, fortuin2020gp}. The second VAE, on the other hand, generates the data based on the ELBO calculated on the full data that is first imputed with the last value carried forward and masked with the \textit{Individualized Missingness Mask} (IMM) that is discussed in section~\ref{mask}. We refer to the first and second VAE as $VAE^{\mathcal{OO}}$and $ VAE^{\mathcal{IMM}}$, respectively. Each LSTM encoder is complemented with input attention highlighting feature importance. Specifically, the attention component computes the weights for each feature conditioned on the encoder's hidden state of the previous time step.  The computed weights are obtained using a deterministic fully connected network, by referring to the previous hidden state $\mathbf{h}_{t-1}$ and the cell state $\mathbf{s}_{t-1}$ in the encoder LSTM unit with:
$$
e_{t}^{k}=\mathbf{v}_{e}^{\top} \tanh \left(\mathbf{W}_{e}\left[\mathbf{h}_{t-1} ; \mathbf{s}_{t-1}\right]+\mathbf{U}_{e} \mathbf{x}^{k}\right)
$$
and
$$
\alpha_{t}^{k}=\frac{\exp \left(e_{t}^{k}\right)}{\sum_{i=1}^{n} \exp \left(e_{t}^{i}\right)},
$$
where $\alpha_{t}^{k}$ is the attention weight of feature $k$ at time $t$. The learnable parameters are denoted in the form of $\mathbf{v}_{e} \in \mathbb{R}^{T}, \mathbf{W}_{e} \in \mathbb{R}^{T \times 2 m}$ and $\mathbf{U}_{e} \in \mathbb{R}^{T \times T}$. The learned attention weights are passed to a softmax function, followed by a dense network trained with the encoder parameters. To this end, the learned weights are multiplied by the input features as shown in 
$$
\tilde{\mathbf{x}}_{t}=\left(\alpha_{t}^{1} x_{t}^{1}, \alpha_{t}^{2} x_{t}^{2}, \cdots, \alpha_{t}^{n} x_{t}^{n}\right)^{\top} .
$$
which can be used to update the hidden state at time $t$:

$$
\mathbf{h}_{t}=f_{1}\left(\mathbf{h}_{t-1}, \tilde{\mathbf{x}}_{t}\right),
$$
The LSTM encoder unit are fed with  $\tilde{\mathbf{x}}_{t}$ instead of  ${\mathbf{x}}_{t}$, giving more attention to important features to encode a better representation of the time-series input.
After encoding the data with feature-wise attention, a temporal attention mechanism is used in the decoder to select relevant encoder hidden states across all time steps. To do so, an attention weight is calculated for each encoder hidden state at time $t$. The calculation of the attention weight is based on the previous hidden state at time $t-1$. Specifically, the attention weight of each encoder hidden state at time $t$ is calculated based on the previous hidden state of the decoder $\mathbf{d}_{t-1} \in \mathbb{R}^{p}$ and the cell state of the LSTM unit $\mathbf{s}_{t-1}^{\prime} \in \mathbb{R}^{p}$ with

$$
l_{t}^{i}=\mathbf{v}_{d}^{\top} \tanh \left(\mathbf{W}_{d}\left[\mathbf{d}_{t-1} ; \mathbf{s}_{t-1}^{\prime}\right]+\mathbf{U}_{d} \mathbf{h}_{i}\right), \quad 1 \leq i \leq T
$$

and

$$
\beta_{t}^{i}=\frac{\exp \left(l_{t}^{i}\right)}{\sum_{j=1}^{T} \exp \left(l_{t}^{j}\right)},
$$

where $\left[\mathbf{d}_{t-1}; \mathbf{s}_{t-1}^{\prime}\right] \in \mathbb{R}^{2 p}$ is a concatenation of the previous hidden state and the cell state of the LSTM unit. $\mathbf{v}_{d} \in \mathbb{R}^{m}$, $\mathbf{W}_{d} \in \mathbb{R}^{m \times 2 p}$ and $\mathbf{U}_{d} \in \mathbb{R}^{m \times m}$ are parameters to learn.  The attention weight $\beta_{t}^{i}$ represents the importance of the $i$-th encoder hidden state at time $t$. Since each encoder hidden state $\mathbf{h}_{i}$ is mapped to a temporal component of the input, the attention mechanism computes the context vector $\mathbf{c}_{t}$ as a weighted sum of all the encoder hidden states $\left\{\mathbf{h}_{1}, \mathbf{h}_{2}, \cdots, \mathbf{h}_{T}\right\}$,

$$
\mathbf{c}_{t}=\sum_{i=1}^{T} \beta_{t}^{i} \mathbf{h}_{i} .
$$
We note that the context vector $\mathbf{c}_{t}$ is distinct at each time step. Once we get the weighted summed context vectors. The decoded data are based on the stacked weighted content vectors, allowing for better representation over time. 
\subsubsection{\textbf{Individualized Missingness Mask (IMM)}}
\label{mask}
 Although multiple works investigated the importance of missingness indicators and their potential impact on prediction tasks \citep{sharafoddini2019new,che2018recurrent,twala2008good}, these works did not account for individual patient-record observation frequencies. Specifically, a time-varying physiological feature that is never measured for a patient can inform the model about the patient's state and medical needs. 
 Moreover, the frequency of measurements can be related to the patient's physiological state and possible deterioration. For example, multiple measurements for cardiac troponins, sensitive biomarkers of myocardial injury, can indicate that doctors suspect a cardiac-related diagnosis or complication ~\citep{daubert2010utility}. In the same intuition, a missing measurement for the same cardiac troponins can indicate that the patient is not suspected of developing such complications, as the clinical guidelines only recommend this measurement for diagnosing myocardial injury~\citep{collinson2022troponin}.  However, this kind of feature-level missingness, can not be represented in the traditional binary missingness masks. 
 For this purpose, we design a mask inspired by natural language processing document frequency metrics, such as the term frequency-inverse document frequency (TFIDF) \citep{ramos2003using} where a metric is calculated based on the relative frequency of appearance of words in a record.  We refer to our mask as the Individualized Missingness Mask (IMM). Specifically, for each time-series feature in an individual patient record, we assign a value of 1 if the sample was observed. We formally define the mask for each time-series feature in a patient record: 
\begin{equation}
\mathbf{IMM} =
\begin{cases}
  1 \text{  if } 
       \begin{aligned}[t]
       m^f_t & =1\\
       \end{aligned}
\\
  \dfrac{\sum_{t}^{T} m^f_i= 1}{T}\text{if } 
       \begin{aligned}[t]
       m^f_t =0\\\
       \end{aligned}
\end{cases}
\end{equation}
where f is a single time-series feature of length T, and  $m^f_t$, corresponds to the binary mask indicating the presence/missingness of a single sample in feature \textit{f} at time t for an individual patient record. 
 \begin{figure*}
     \centering
     \includegraphics[width= 0.7\textwidth]{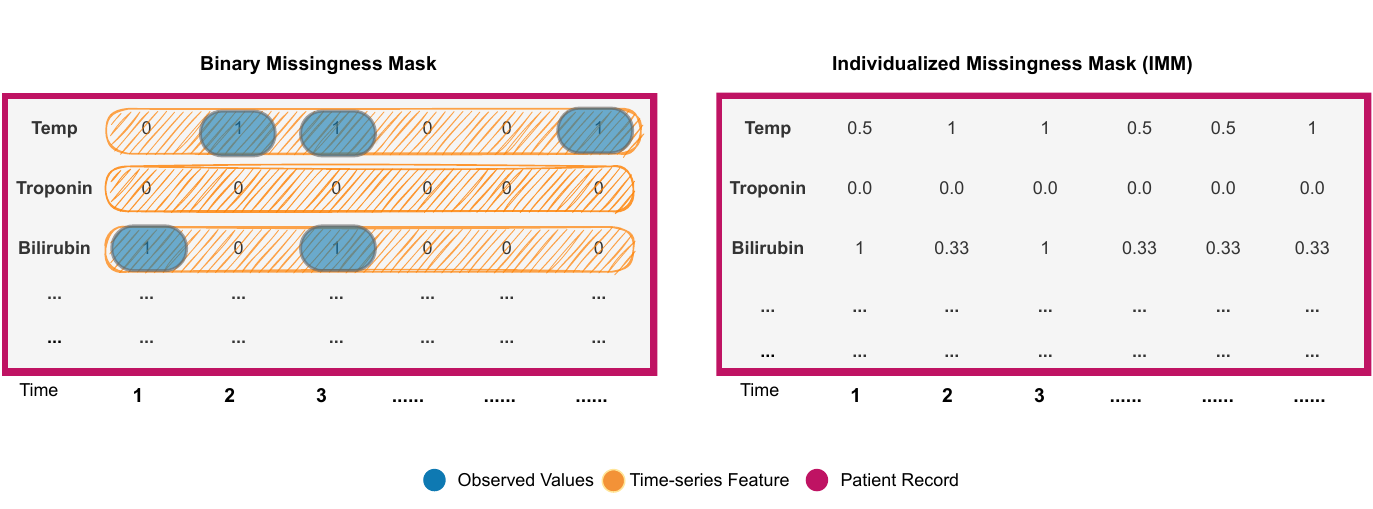}
     \caption{An example showcasing the difference between a binary missingness mask and an individualized missingness mask (IMM)}
     \label{fig:IMM}
 \end{figure*}
 
\subsubsection{\textbf{Conditional Generation}}
To incorporate the discrete treatment data and patient demographic information, we utilize a conditional VAE architecture that allows for targeted data generation \citep{sohn2015learning}, making the imputation more specific to the treatments and patient's age group and sex.  Specifically, we utilize the 3D matrix of time-series treatments and concatenate it with the one-hot encoding of two demographic features (age and sex), which are used as conditions for our generative model. We note that the treatment and demographic features are fully observed, which makes them ideal for conditioning the VAEs.

\subsubsection{\textbf{Latent-Space Losses}}
While the vanilla VAE losses are based on ELBO losses, the sum of the  Kullback-Leibler divergence and reconstruction loss in the observed space, we define multiple loss constraints for the optimization of the generative imputation model in latent space, namely: matching, semantic and contrastive losses.

\textbf{Matching Loss.}  By mapping the time-series into a shared low-dimensional space, a shared weight constraint, and assuming that both data instances are representations of the same patient, we expect that the data generated by $VAE^{\mathcal{OO}}$ and $ VAE^{\mathcal{IMM}}$ should be close in the latent space. Therefore, we optimize the Euclidean distance between the representations as follows:
$$
\mathcal{L}^{\text {Match}}=\mathbb{E}_{\mathbf{z} \sim p_{\mathbf{z}}}\left[\left\|\mathbf{z}^{\mathcal{OO}}-\mathbf{z}^{\mathcal{IMM}}\right\|^{2}\right]
$$
where $\mathbf{z}^{\mathcal{OO}}$ represent and $\mathbf{z}^{\mathcal{IMM}}$ represent the latent space mappings produced by each of the VAEs.

\textbf{\textbf{Semantic loss}}. The semantic loss is used to ensure the learned representations are useful for performing downstream tasks. For this purpose, we concatenate the learned representations from both VAEs and implement a classifier to predict an outcome of interest. The loss is calculated based on the cross-entropy: 
\begin{equation*}
    {{{{\mathcal{L}}}}}^{{{{\rm{Semnatic}}}}}={{\mathbb{E}}}_{{{{{\bf{z}}}}}={{{\rm{CE}}}}\left({f}_{{{{\rm{classifier}}}}}({{{{\bf{z}}}}}),{{{\bf{y}}}}\right)}
\end{equation*}
where z represents the representations, and y is the predicted outcome. CE stands for the standard cross entropy loss used for classification models~\citep{mannor2005cross}.

\textbf{\textbf{Contrastive loss}}. We also measure how close each pair of representations learned by both VAEs for the same patient, compared to those of other patients using contrastive loss \citep{chopra2005learning}. Given a dataset of $N$ patient records, the objective of contrastive loss is to have latent vectors derived from $\mathbf{VAE}^{\mathcal{OO}}$ and  $\mathbf{VAE}^{\mathcal{IMM}}$ for the same patient to be similar to each other, yet different from those derived from different patients. Therefore the equation is based on minimizing the distance between the latent space vectors $\mathbf{z}^{\mathcal{OO}}$ and $\mathbf{z}^{\mathcal{IMM}}$ and when they correspond to the same patient and maximize the distance when the latent vectors are for other patients. For this, we use the same formulation derived by \citep{li2023generating}:
$$
\mathcal{L}_{i, j}^{\text{Contra }}=-\log \frac{\exp \left(\operatorname{sim}\left(\mathbf{z}^{\mathcal{OO}}_i, \mathbf{z}^{\mathcal{IMM}}_j\right) / \tau\right)}{\sum_{i^{k = 1}}^{2 N} {\left[ {k} \neq i \right]} \exp \left(\operatorname{sim}\left(\mathbf{z}^{\mathcal{OO}}_i, \mathbf{z}^{\mathcal{IMM}}_k\right) / \tau\right)}
$$
where the similarity function in the numerator is the cosine distance between the latent-space vectors, while the denominator is the cosine similarity between a vector belonging to one patient, and those belonging to all others. The final contrastive loss is calculated across all patients included in the dataset and summed for both. 

\subsubsection{\textbf{Masked Imputation Task (MIT) Loss}}
In addition to the reconstruction loss used in VAEs, we introduce a new loss that forces the model to minimize the loss of artificially introduced missing values, which we refer to as  Masked Imputation Task (MIT) loss as suggested in the work of~\citep{du2023saits}. The reconstruction loss used in calculating the ELBO in the VAE component is calculated only for the observed values, since the ground truth of those values cannot be accessed. The MIT loss masks a percentage of the observed values and introduces them to the model as missing values, simulating a missingess setup to force the model to generate accurate imputations. The MIT loss is calculated based on the mean squared error of the ground truth and the predicted values for the masked values.

\subsubsection{\textbf{Discriminator}}
The dual-VAE architecture is complemented with an LSTM-based classifier that outputs the probability that each sample belongs to either the real or imputed (generated) classes. Inspired by the vanilla GAN's discriminator~\citep{goodfellow2014generative}, we refer to this classifier as the discriminator. As IGNITE generates high-quality data samples, the aim is to have the discriminator not be able to tell which ones are real and which ones are the generated imputations, implying that the imputations are as real as possible.  The error produced by the discriminator is optimized along with the overall model's losses.

In summary, IGNITE is optimized jointly using the total loss function
$$
\mathcal{L}_{OO}=\gamma\mathcal{L}^{\mathrm{Reconst.}}+\delta\mathcal{L}^{\text {KL}} + \epsilon\mathcal{L}^{\text {Match.}}+\zeta\mathcal{L}^{\text {Semantic}} +\eta\mathcal{L}^{\text {Contra.}} + \theta \mathcal{L}^{\mathrm{Discriminator}} + \iota \mathcal{L}^{\mathrm{Discriminator}}$$
where $\gamma, \delta, \epsilon, \zeta, \theta $, and $\iota$ are scalars used to determine the ratio of each loss in the overall model's loss, which are all finetuned on each dataset. 

\textcolor{black}{The optimization of IGNITE relies critically on the precise selection of hyperparameters, which was achieved by using a Bayesian hyperparameter search while minimizing the reconstruction loss.  For each dataset, a customized Bayesian search was performed to accommodate its unique characteristics, such as variability in missing data patterns and feature scale.  The final selection of hyperparameters was determined based on their ability to maximize imputation effectiveness and prediction accuracy, as evidenced by extensive validation experiments. This targeted approach ensured that the IGNITE framework was optimally tuned to each dataset's specific challenges, enhancing both imputation quality and clinical utility. The searched ranges of IGNITE scalar weights are shown in the Supplementary Information Section~\ref{hyper_ignite}}. While both VAEs generate reconstructions, we are interested in the one produced by the $ VAE^{\mathcal{OO}}$, as it maps the data to the observed space and incorporates information from the IMM via the joint training with $ VAE^{\mathcal{IMM}}$.

\subsection{\textbf{Datasets}}
To evaluate the proposed model we use three large-scale EHR datasets for intensive care unit patients from the United States and the Netherlands. The datasets include patients admitted to the ICU for medical, surgical and trauma care, all of whom have not experienced a mortality outcome during the first 48 hours. The included datasets are as follows:

\begin{enumerate}

\item {\textbf{PhysioNet Challenge 2012}, an open-access dataset for ICU patients used by most time-series imputations, where the overall missingness is more than 80\% missingness~\citep{silva2012predicting}. The full dataset includes 12,000 encounters and 35 time-series physiological data features. The dataset had one time-series feature reporting administered treatments: mechanical ventilation and a binary outcome indicating mortality with 14.25\%. 

\item{\textbf{eICU}, a multi-center database for ICU data, from more than 208 hospitals throughout the United States between 2014-2015, making it a good choice for validating models across multi-centers~\citep{pollard2018eicu}. This database includes more than 200,000 admissions and 827 million observations. We follow the same preprocessing applied by Li et al.~\cite{li2023generating}. The included encounters for this study are those who had at least 48 hours of ICU stay, with at least 24 hours without mortality outcome.  The final dataset included 54,423 patient encounters with 55 time-series physiological data features, as shown in the Supplementary Information section~\ref{A}. For the treatment features, we aggregated treatments in terms of treatment concepts and kept those with at least 10\% administration to patients. This resulted in three included time-series treatment features: oxygen therapy, vasopressors, and antibiotics, respectively. }

\item {\textbf{HiRID}}, a single centre critical care dataset collected from Intensive Care Medicine of the Bern University Hospital, Switzerland. The dataset contains admissions for more than 33,000 patients. Like eICU, for this study, we only included encounters for patients with a minimum of 48 hours of ICU stay and at least 24 hours without mortality outcome. The included physiological time-series features are 50 features, shown in the Supplementary Information section~\ref{B}. We also applied the treatment concept grouping and chose those with at least 10\% prevalence. The included treatments included seven features: oxygen therapy, crystalloids, vasopressors, vasodilators, insulin, painkillers and anticoagulants, respectively.} 
\end{enumerate}

The datasets are preprocessed so that the measurements taken in each hour are aggregated, resulting in 48-time steps for each feature. If the feature was measured more than once, the arithmetic mean of the measurement was recorded each hour. The value is considered missing if the feature is not measured in that respective hour.  The three datasets are described in terms of missingness and included features in Table~\ref{miss}, Supplementary sections~\ref{A} and~\ref{B} for PhysioNet 2012, eICU and HiRID datasets, respectively. The first column refers to the overall missingness across all patient records, features and time steps, which we refer to as sample-wise missingness. The second column refers to the percentage of patients who did not have any observations of the corresponding feature, which indicates a type of missingness we refer to as feature-wise missingness.  We formally define the two types of missingness below: 
\begin{itemize}
\item \textbf{Feature-wise missingness}: \textcolor{black}{This refers to the percentage of patients who have no observations for a specific feature. It captures the extent to which a particular feature is entirely missing for a subset of patients. In this case, missingness is evaluated at the level of the feature for each patient, highlighting if some features are completely absent in certain patient records.}

\item \textbf{Sample-wise missingness}:  \textcolor{black}{This refers to the overall percentage of missing values across all patient records, features, and time steps. It is calculated as the ratio of the total number of missing entries to the total number of possible entries. This metric captures how much data is missing at the level of individual data points, giving a sense of the overall missingness across the dataset.}
\end{itemize}

\begin{table}[!ht]
\centering
       \caption{Description of Included Features of PhysioNet Challenge 2012 dataset}
    \resizebox{0.4\linewidth}{!}{
    \begin{tabular}{lcc}
    \toprule
    &\multicolumn{2}{c}{\textbf{Time-series Missingness}}\\
    \cmidrule(lr){2-3}
    \textbf{Feature} & \textbf{\% Sample-level} & \textbf{\% Feature-level } \\
    \midrule
         ALP & 98.36 & 57.53 \\
        HR & 9.89 & 1.53 \\
        DiasABP & 45.87 & 29.78 \\
        Na & 92.90 & 1.77 \\
        Lactate & 95.89 & 45.20 \\
        NIDiasABP & 57.97 & 12.62 \\
        PaO2 & 88.49 & 24.55 \\
        WBC & 93.27 & 1.78 \\
        pH & 87.94 & 24.09 \\
        Albumin & 98.74 & 59.42 \\
        ALT & 98.32 & 56.56 \\
        Glucose & 93.18 & 2.45 \\
        SaO2 & 95.99 & 55.28 \\
        Temp & 62.91 & 1.54 \\
        AST & 98.32 & 56.55 \\
        Bilirubin & 98.30 & 56.55 \\
        BUN & 92.77 & 1.53 \\
        RespRate & 75.96 & 72.27 \\
        Mg & 92.91 & 2.43 \\
        HCT & 90.51 & 1.58 \\
        SysABP & 45.86 & 29.78 \\
        FiO2 & 84.32 & 32.37 \\
        K & 92.42 & 2.08 \\
        GCS & 67.98 & 1.54 \\
        Cholesterol & 99.83 & 92.10 \\
        NISysABP & 57.94 & 12.40 \\
        TroponinT & 98.92 & 78.03 \\
        MAP & 46.18 & 29.93 \\
        TroponinI & 99.80 & 95.29 \\
        PaCO2 & 88.47 & 24.54 \\
        Platelets & 92.64 & 1.65 \\
        Urine & 30.80 & 2.58 \\
        NIMAP & 58.55 & 12.80 \\
        Creatinine & 92.73 & 1.53 \\
        HCO3 & 92.92 & 1.73 \\
        \bottomrule
        
    \end{tabular}}
    \label{miss}
\end{table}
\subsection{Experimental Setting}
\subsubsection{\textbf{Downstream Task}}
Due to the absence of ground-truth labels for highly missing EHRs data, 
we evaluate IGNITE's imputations and compare them with those produced by the baseline methods using an extensive downstream task framework based on the original data missingness. To do so, we train an LSTM~\cite{hochreiter1997long} classifier to predict mortality in any of the components of the framework explained below. \textcolor{black}{During the development of the IGNITE model, we strictly adhered to protocols that ensure that there is no overlap or information exchange between training and test sets. The imputation model was trained exclusively on the designated training data. This data set includes comprehensive features but is entirely separate from the test dataset, which was set aside at the outset of the study.} All imputation models were trained in the training set with a ratio of 80\% of the full datasets and evaluated at inference by producing imputations for an unseen test set. The Supplementary Material section~\ref{hyper_downstream} shows the hyperparameter search ranges for each predictive model. The final performance is reported in terms of both the area under the receiving operating curve (AUROC)~\citep{fan2006understanding} and the area under the precision recall curve (AUPRC)~\citep{sofaer2019area}. AUPRC measures the area under the precision and recall curve, making it more informative for class imbalance cases in all three datasets.  In addition to the downstream task usually evaluated in imputation works, we propose an evaluation framework that tests IGNITE's ability to impute if trained and tested on data with various types and percentages of naturally occurring missingness. 
The proposed downstream task framework includes: 
\begin{enumerate}
\item {\textbf{Performance across the full population}}
In this analysis, we evaluate the machine learning model's performance in a mortality classification task using the full dataset, irrespective of the missingness rate. This is the common evaluation used by related works in the literature \citep{fortuin2020gp,cao2018brits,luo2019e2gan,luo2018multivariate}.

\item {\textbf{Performance across patients with overall sample-wise missingness}}
In this analysis, we investigate the performance of the mortality prediction classifier when trained in population subgroups where the sample-wise missingness across all features and time-steps is $\leq$ 25\%, 25-75\% and $\geq$ 75\%, respectively. We did not report results for a subset that has fewer than 800 samples, as they are unreliable.
\item {\textbf{Performance across patients with high feature-wise missingness}}
While modelling overall missingness is important, for this analysis, we focus on patients with a high percentage of features that were completely missing, i.e., features that were not observed for a particular patient. This analysis aims to evaluate the impact of various imputation methods when some features are never measured for population subgroups. Like the sample-wise missingness experiments, the investigated subgroups were also $\leq$ 25\%, 25-75\% and $\geq$ 75\%, respectively. Similarly to the sample-wise missingness experiments, the results of subsets with low-sample sizes were excluded. 

\end{enumerate}
\subsubsection{Reconstruction Task}
Despite not having the ground-truth labels for the missing values in the EHR dataset, there are observed values which we can use to evaluate the accuracy of reconstruction using the proposed model. Therefore, we randomly mask a percentage of the observed values for each patient in the test set as hidden and then observe the model’s ability to recover the original values. A similar approach was previously used by \citep{luo2018multivariate,fortuin2020gp,nazabal2020handling}. The main purpose of this validation method is to measure the models’ ability to reconstruct the original data and quantitatively compare it to other baseline methods. \textcolor{black}{In our evaluation, the quality of the imputed data is assessed by comparing the reconstructions to the observed values using Root-Mean-Square Error (RMSE) and Mean Absolute Error (MAE). To account for variability in the amount of data missing from each patient's record, we introduce missingness randomly across the test set population. The RMSE for each patient is then calculated and weighted according to the proportion of missingness introduced in their records, ensuring that our metrics accurately reflect the severity and distribution of missing data. For this analysis, missingness was introduced at varying levels of 10\%, 20\%, and 50\% to test the robustness of our imputation methods under different scenarios. The final results are presented as the mean and standard deviation of RMSE and MAE across all patients for each baseline method, providing a comprehensive view of performance across different degrees of data incompleteness.}
\subsubsection{\textbf{Baselines}}
We benchmark our model with several commonly used and state-of-the-art imputation methods listed below:  
\begin{itemize}
       \item \textbf{LOCF}: Last Observation Carried Forward, an imputation method often used in clinical studies, where the last observed value for the patient is used to impute all subsequent missing observation points~\citep{shao2003last}.
    \item \textbf{MICE}: Multivariate Imputation by Chained Equations is a simulation-based statistical model where N complete datasets are produced from the incomplete dataset by imputing the missing N times. These completed N datasets are analyzed and then pooled into a single dataset~\citep{jakobsen2017and}. This method is often used in epidemiological studies and is designed for tabular datasets.
 
    \item\textbf{GP-VAE}: a deep probabilistic model that utilizes a Gaussian-Process with a Cauchy Kernel and a VAE to impute the missing multi-variate time-series data \citep{fortuin2020gp}.
    \item{\textbf{Transformer}}: An implementation of attention is all you need paper~\citep {vaswani2017attention}, adapted for time-series imputations.
    \item{\textbf{BRITS}:} Bidirectional Recurrent Imputation for time-series, an RNN-based model where bidirectional long short-term memory network (BiLSTM) is used to predict the intermediate states of partially observed data, given its past and future states \citep{cao2018brits}.
    \item{\textbf{SAITS}}: Self-Attention-based Imputation for time-series, a model based on using the self-attention mechanism where it learns to impute missing values from a weighted combination of diagonally-masked self-attention (DMSA) blocks~\citep{du2023saits}. 

\end{itemize}

\section*{Data availability}
 The data used in this work can be downloaded directly from the PhysioNet website \url{(https://physionet.org/content/challenge-2012/1.0.0/)}. The other two datasets could be downloaded from the PhysioNet website after completing the necessary training at \url{https://physionet.org/content/HiRID/1.1.1/} and \url{https://physionet.org/content/eicu-crd/2.0/}, for HiRID and eICU respectively. 

 \section*{Code availability}
To allow the reproducibility of our work, we make our code publicly available at \url{https://github.com/Ghadeer-Ghosheh/IGNITE}.

\acknow{TZ was supported by the Royal Academy of Engineering under the Research Fellowship scheme and the National Institute for Health Research Oxford Biomedical Research Centre. JL was funded by the National Natural Science Foundation of China (NSFC) 82302352.}
\showacknow{} 

\bibliography{main}

\newpage
\appendix
\onecolumn
\renewcommand*{\thesubfigure}{(\arabic{subfigure})}

\section*{\Large{Supplementary Information}}
\section{Hyper-parameters for IGNITE}
\label{hyper_ignite}
The hyperparameters used for IGNITE optimization. The final parameters were chosen based on a hyper-parameter Bayesian search tuned for each dataset.

\begin{table}[!ht]
    \centering
    \caption{Hyper-parameter Search Ranges}
        \resizebox{0.4\linewidth}{!}{
    \begin{tabular}{lc}
           \toprule
         \textbf{Parameter} & \textbf{Search Range} \\
         \midrule
           $\gamma$ Reconstruction & [2.e-02 - 2.e+02] \\
            $\delta$ KL Divergence &  [2.e-04 - 2.e+04] \\
            $\epsilon$ Matching & [1.e-04 - 1.e+04] \\
             $\zeta$ Semantic & [1.e-04 - 1.e+04] \\
              $\eta$ Contrastive & [1.e-04 - 1.e+04] \\
          $\theta$ MIT & [1.e-01 - 3.e+01] \\
           $\iota$ Discriminator & [1.e-04 - 1.e-01] \\
%
           \bottomrule
    \end{tabular}}

    \label{tab:my_label}
\end{table}
\clearpage
\section{Data set Description Table: eICU}
The included features for the eICU dataset along with the sample-wise and feature-wise missingness statistics, are shown in Table~\ref{miss_eicu}.
\label{A}
\begin{table}[!ht]
\centering
       \caption{Description of Included Features of eICU dataset}
    \resizebox{0.4\linewidth}{!}{
    \begin{tabular}{lcc}
    \toprule
    &\multicolumn{2}{c}{\textbf{Time-series Missingness}}\\
    \cmidrule(lr){2-3}
    \textbf{Feature} & \textbf{\% Sample-wise} & \textbf{\% Feature-wise } \\
    \midrule
  
         glucose & 83.53 & 3.87 \\ 
        SpO2 & 45.19 & 15.19 \\ 
        Noninvasive BP Diastolic & 19.40 & 1.12 \\ 
        Noninvasive BP Systolic & 19.40 & 1.12 \\ 
        Noninvasive BP Mean & 19.80 & 1.42 \\ 
        RR & 36.36 & 8.26 \\ 
        -basos & 98.20 & 50.07 \\ 
        -eos & 98.10 & 47.34 \\ 
        -lymphs & 97.96 & 44.91 \\ 
        -monos & 97.97 & 45.14 \\ 
        -polys & 98.18 & 50.36 \\ 
        ALT (SGPT) & 98.73 & 62.18 \\ 
        AST (SGOT) & 98.71 & 61.69 \\ 
        BUN & 95.62 & 6.05 \\ 
        FiO2 & 99.06 & 78.71 \\ 
        HCO3 & 98.95 & 76.76 \\ 
        Hct & 95.64 & 7.88 \\ 
        Hgb & 95.59 & 7.94 \\ 
        MCH & 96.39 & 15.38 \\ 
        MCHC & 96.21 & 10.94 \\ 
        MCV & 96.20 & 10.93 \\ 
        PT & 98.93 & 71.43 \\ 
        PT - INR & 98.89 & 70.55 \\ 
        PTT & 99.15 & 82.92 \\ 
        RBC & 96.13 & 8.74 \\ 
        RDW & 96.38 & 14.63 \\ 
        WBC x 1000 & 96.12 & 8.47 \\ 
        albumin & 98.49 & 56.90 \\ 
        alkaline phos. & 98.72 & 62.11 \\ 
        anion gap & 96.49 & 25.14 \\ 
        bicarbonate & 95.84 & 10.83 \\ 
        calcium & 95.76 & 8.73 \\ 
        chloride & 95.58 & 5.83 \\ 
        creatinine & 95.60 & 5.94 \\ 
        lactate & 99.53 & 88.61 \\ 
        magnesium & 97.46 & 39.11 \\ 
        pH & 98.90 & 75.86 \\ 
        paCO2 & 98.90 & 75.79 \\ 
        paO2 & 98.88 & 75.50 \\ 
        phosphate & 98.35 & 57.94 \\ 
        platelets x 1000 & 96.10 & 8.89 \\ 
        potassium & 94.81 & 5.34 \\ 
        sodium & 95.32 & 5.70 \\ 
        total bilirubin & 98.75 & 62.82 \\ 
        total protein & 98.72 & 61.92 \\ 
        troponin - I & 99.66 & 91.34 \\ 
        Base Excess & 99.13 & 80.24 \\ 
        urinary specific gravity & 99.84 & 93.12 \\ 
        Heart Rate & 34.00 & 8.29 \\ 
        Temperature & 75.99 & 6.12 \\ 
        Tidal Volume (set) & 97.24 & 84.81 \\ 
        MPV & 97.42 & 39.02 \\ 
        Exhaled MV & 98.31 & 89.33 \\ 
        Exhaled TV (patient) & 98.70 & 89.80 \\ 
        SaO2 & 97.74 & 81.36 \\ 
        \bottomrule
        
    \end{tabular}}
    \label{miss_eicu}
\end{table}
\clearpage
\section{Data set Description Table: HiRID}
The included features for HiRID dataset along with the sample-wise and feature-wise missingness statistics, are shown in Table~\ref{miss3}.
\label{B}
\begin{table}[!ht]
\centering
       \caption{Description of Included Features of HiRID dataset}
    \resizebox{0.45\linewidth}{!}{
    \begin{tabular}{lcc}
    \toprule
    &\multicolumn{2}{c}{\textbf{Time-series Missingness}}\\
    \cmidrule(lr){2-3}
    \textbf{Feature} & \textbf{\% Sample-wise} & \textbf{\% Feature-wise } \\
    \midrule
  	        Heart rate & 1.44 & 0.00 \\ 
         Body Temperature Core & 78.08 & 38.45 \\ 
        Arterial Blood Pressure systolic & 11.23 & 4.12 \\ 
        Arterial Blood Pressure diastolic & 11.25 & 4.12 \\ 
        Arterial Blood Pressure mean & 11.23 & 4.09 \\ 
        Non-invasive Blood Pressure systolic & 89.82 & 37.85 \\ 
        Non-invasive Blood Pressure diastolic & 89.82 & 37.86 \\ 
        Non-invasive Blood Pressure mean & 89.82 & 37.89 \\ 
        Pulmonary Blood Pressure mean & 92.18 & 84.95 \\ 
        Pulmonary Blood Pressure systolic & 92.17 & 84.94 \\ 
        Pulmonary Blood Pressure diastolic & 92.32 & 84.97 \\ 
        Cardiac Output & 92.23 & 85.18 \\ 
        mixed venous oxygen saturation & 92.34 & 85.37 \\ 
        central venous pressure & 37.26 & 29.32 \\ 
        ECG ST1 & 38.97 & 32.14 \\ 
        ECG ST2 & 56.94 & 36.75 \\ 
        ECG ST3 & 64.99 & 43.55 \\ 
        Saturation Oxygen peripheral/capillary & 3.95 & 0.03 \\ 
        End tidal CO2 concentration & 68.95 & 43.68 \\ 
      Respiratory rate & 32.66 & 0.07 \\ 
   Hourly urinary output & 73.52 & 1.48 \\ 
     ICP & 94.15 & 90.97 \\ 
      Measurement of output from drain & 96.65 & 92.36 \\ 
    Base excess in Arterial blood by calculation & 90.60 & 14.77 \\ 
      Carboxyhemoglobin in Arterial blood
 & 90.65 & 15.29 \\ 
      Hemoglobin in Arterial blood
 & 91.52 & 24.18 \\ 
        Bicarbonate in Arterial blood~ & 90.60 & 14.77 \\ 
       Lactate in Arterial blood
 & 90.67 & 14.04 \\ 
       Methemoglobin in Arterial blood
 & 90.65 & 15.30 \\ 
       a\_pH & 91.43 & 24.73 \\ 
      a\_pCO2 & 91.63 & 24.93 \\ 
        a\_PO2 & 91.52 & 24.82 \\ 
       a\_SO2 & 90.65 & 15.27 \\ 
        Potassium & 89.48 & 2.19 \\ 
Sodium & 89.40 & 2.10 \\ 
     Cl- & 90.84 & 15.89 \\ 
     Ca2+i & 90.61 & 15.26 \\ 
        Phosphate & 97.12 & 22.66 \\ 
        Magnesium in Blood
 & 97.27 & 22.95 \\ 
      Urea & 96.98 & 21.28 \\ 
       Crea & 96.36 & 8.49 \\ 
      INR & 97.89 & 45.10 \\ 
       Glucose & 79.55 & 0.94 \\ 
       C-reactive protein & 96.53 & 8.96 \\ 
 Hemoglobin in Blood
 & 95.49 & 3.83 \\ 
     Total white blood cell count	 & 95.55 & 3.73 \\ 
      Platelet count	 & 95.71 & 4.57 \\ 
     MCH & 95.57 & 3.81 \\ 
       MCHC & 95.57 & 3.87 \\ 
       MCV  & 95.56 & 3.80 \\ 
        \bottomrule
      \end{tabular}}
    \label{miss3}
\end{table}
\clearpage

\section{Hyper-parameters for Downstream Tasks}
The hyper-parameters range that were used for training the downstream LSTM network. 
\label{hyper_downstream}
\begin{table}[!ht]
    \centering
    \caption{Hyper-parameter Search Ranges}
        \resizebox{0.4\linewidth}{!}{
    \begin{tabular}{lc}
    \toprule
         \textbf{Parameter} & \textbf{Search Parameters} \\
         \midrule
           Dropout & [1.e-01 - 9.e-01]\\
           Learning Rate &  [1.e-04 - 1.e-01]\\
           Batch Size & [64-1024]\\
           \bottomrule
    \end{tabular}}

    \label{tab:my_label}
\end{table}

\clearpage

\section{Statistical Analysis of Imputation Methods}
\label{stat}
\textcolor{black}{
To assess the statistical significance of the performance differences between IGNITE and other imputation methods, we applied the Wilcoxon signed-rank test. This non-parametric test is suitable for comparing paired samples and does not assume a normal distribution of the differences, making it ideal for our scenario where performance metrics may vary significantly across datasets.}

\textcolor{black}{We compared the AUROC and AUPRC metrics of the IGNITE model with those of six other imputation methods: LOCF, MICE, GP-VAE, Transformer, BRITS, and SAITS. The comparison was performed separately for each dataset: PhysioNet 2012, eICU, and HiRID.The following tables summarize the p-values of the Wilcoxon signed-rank test for each comparison. A p-value below 0.05 indicates a statistically significant difference favoring the IGNITE model.}

\begin{table*}[htb]
\centering
\caption{Wilcoxon Signed-Rank Test Results (P-values)}
\begin{minipage}{0.32\linewidth}
\centering
\caption*{PhysioNet 2012 Dataset}
\begin{tabular}{ll}
\toprule
\textbf{Metric \& Method} & \textbf{P-value} \\
\midrule
AUROC: IGNITE vs LOCF & 0.03125 \\
AUROC: IGNITE vs MICE & 0.4375 \\
AUROC: IGNITE vs GP-VAE & 0.03125 \\
AUROC: IGNITE vs Transformer & 0.625 \\
AUROC: IGNITE vs BRITS & 0.125 \\
AUROC: IGNITE vs SAITS & 0.125 \\
\midrule
AUPRC: IGNITE vs LOCF & 0.03125 \\
AUPRC: IGNITE vs MICE & 0.0625 \\
AUPRC: IGNITE vs GP-VAE & 0.03125 \\
AUPRC: IGNITE vs Transformer & 0.125 \\
AUPRC: IGNITE vs BRITS & 0.125 \\
AUPRC: IGNITE vs SAITS & 0.4375 \\
\bottomrule
\end{tabular}
\end{minipage}
\hfill
\begin{minipage}{0.32\linewidth}
\centering
\caption*{eICU Dataset}
\begin{tabular}{ll}
\toprule
\textbf{Metric \& Method} & \textbf{P-value} \\
\midrule
AUROC: IGNITE vs LOCF & 0.03125 \\
AUROC: IGNITE vs MICE & 0.03125 \\
AUROC: IGNITE vs GP-VAE & 0.03125 \\
AUROC: IGNITE vs Transformer & 0.0625 \\
AUROC: IGNITE vs BRITS & 0.03125 \\
AUROC: IGNITE vs SAITS & 0.03125 \\
\midrule
AUPRC: IGNITE vs LOCF & 0.03125 \\
AUPRC: IGNITE vs MICE & 0.03125 \\
AUPRC: IGNITE vs GP-VAE & 0.03125 \\
AUPRC: IGNITE vs Transformer & 0.03125 \\
AUPRC: IGNITE vs BRITS & 0.03125 \\
AUPRC: IGNITE vs SAITS & 0.03125 \\
\bottomrule
\end{tabular}
\end{minipage}
\hfill
\begin{minipage}{0.32\linewidth}
\centering
\caption*{HiRID Dataset}

\begin{tabular}{ll}
\toprule
\textbf{Metric \& Method} & \textbf{P-value} \\
\midrule
AUROC: IGNITE vs LOCF & 0.03125 \\
AUROC: IGNITE vs MICE & 0.03125 \\
AUROC: IGNITE vs GP-VAE & 0.4375 \\
AUROC: IGNITE vs Transformer & 0.0625 \\
AUROC: IGNITE vs BRITS & 0.03125 \\
AUROC: IGNITE vs SAITS & 0.0625 \\
\midrule
AUPRC: IGNITE vs LOCF & 0.03125 \\
AUPRC: IGNITE vs MICE & 0.03125 \\
AUPRC: IGNITE vs GP-VAE & 0.03125 \\
AUPRC: IGNITE vs Transformer & 0.0625 \\
AUPRC: IGNITE vs BRITS & 0.4375 \\
AUPRC: IGNITE vs SAITS & 0.0625 \\
\bottomrule
\end{tabular}
\end{minipage}
\end{table*}

\textcolor{black}{The Wilcoxon signed-rank test results show that the IGNITE model consistently outperforms most other imputation methods, particularly in the eICU dataset, where the differences were statistically significant across all comparisons. For the PhysioNet 2012 and HiRID datasets, significant differences were observed in several comparisons, though not all. These results highlight the robustness of the IGNITE model in handling missing data and achieving superior performance across different clinical datasets.}
\clearpage

\section{Reconstruction Results for HiriD\& eICU}
The performance of the baseline methods compared to IGNITE in terms of the reconstruction task. We present the results for HiRID and eICU in Tables~\ref{mae} and ~\ref{mae2}, respectively. In both experiments and in all percentages of masking, IGNITE consistently outperformed all other baselines.
\label{reconst}

\begin{table}[!ht]
    \centering
    \caption{Performance of the various baselines of the reconstruction task on the HiRID dataset}
    \resizebox{\linewidth}{!}{
    \begin{tabular}{lccccccc}
    \toprule
     & \multicolumn{3}{c}{\textbf{RMSE}}&\multicolumn{3}{c}{\textbf{MAE}}
   \\
   \midrule
     \textbf{Introduced Missingness}& \textbf{10\%}&  \textbf{20\%}&\textbf{50\%} & \textbf{10\%} &  \textbf{20\%}&\textbf{50\%} \\
   \midrule
       LOCF & 0.181 (0.058) & 0.184 (0.052)  & 0.217 (0.045) &  0.117 (0.042) & 0.119 (0.039) & 0.136 (0.040) \\ 
        MICE & 0.082 (0.023) & 0.082 (0.021) & 0.081 (0.019) &  0.060 (0.017) &  0.059 (0.015) &  0.058 (0.014)  \\
        GP-VAE & 0.147 (0.028)  & 0.172 (0.023)  & 0.204 (0.019)   &  0.104 (0.023) &  0.122 (0.021) &  0.153 (0.019)  \\ 
        Transformer &  0.085 (0.025)  & 0.085 (0.023 )  & 0.082 (0.021) &  0.061 (0.018) &  0.061 (0.016)& 0.058 (0.015) \\ 
        BRITS &  0.084 (0.026)  & 0.084 (0.023) & 0.085 (0.023) &  0.061 (0.018) & 0.060 (0.017) &  0.061 (0.018)\\ 
        SAITS & 0.086 (0.025) & 0.085 (0.023) & 0.083 (0.021)  &  0.062 (0.018) &  0.060 (0.016) & 0.059 (0.016)\\ 
        IGNITE & \textbf{0.079 (0.023)}   & \textbf{0.079 (0.021)} & \textbf{0.078 (0.019)}  &  \textbf{0.057 (0.017)}  & \textbf{0.057 (0.015)} &  \textbf{0.056 (0.015)}  \\
        \bottomrule
    \end{tabular}}
    \label{mae}
\end{table}

\begin{table}[!ht]
    \centering
    \caption{Performance of the various baselines of the reconstruction task on the eICU dataset}
    \resizebox{\linewidth}{!}{
    \begin{tabular}{lccccccc}
    \toprule
     & \multicolumn{3}{c}{\textbf{RMSE}}&\multicolumn{3}{c}{\textbf{MAE}}
   \\
   \midrule
     \textbf{Introduced Missingness}& \textbf{10\%}&  \textbf{20\%}&\textbf{50\%} & \textbf{10\%} &  \textbf{20\%}&\textbf{50\%} \\
   \midrule
       LOCF & 0.092 (0.048) & 0.096 (0.040) & 0.109 (0.032) & 0.044 (0.023) & 0.044 (0.018) & 0.049 (0.016) \\ 
        MICE & 0.028 (0.015) &  0.028 (0.013) & 0.029 (0.012) & 0.014 (0.007) & 0.014 (0.006) &  0.014 (0.005)  \\
        GP-VAE & 0.028 (0.015) & 0.028 (0.013) & 0.028 (0.012) & 0.014 (0.006) & 0.014 (0.006) & \textbf{0.013 (0.005)} \\ 
        Transformer & 0.028 (0.016) & 0.029 (0.014)  & 0.030 (0.012) & 0.015 (0.007) & 0.014 (0.006) & 0.015 (0.005)   \\ 
        BRITS & 0.029 (0.017) & 0.030 (0.014) & 0.031 (0.012) &  0.014 (0.008) & 0.014 (0.006) & 0.014 (0.005) \\ 
        SAITS & 0.032 (0.019) & 0.034 (0.017) & 0.034 (0.014) &  0.016 (0.009) & 0.016 (0.008) &  0.016 (0.006)  \\ 
        IGNITE & \textbf{0.025 (0.016)} & \textbf{0.026 (0.016)} & \textbf{0.026 (0.016)} & \textbf{0.013 (0.007)} & \textbf{0.013 (0.007)} & \textbf{0.013 (0.007)} \\
        \bottomrule
    \end{tabular}}
    \label{mae2}
\end{table}

\clearpage
\section{Visualizations for Reconstructions}
\label{vis}
\begin{figure}[h!]
	\centering
	\subfigure[{Imputation for a patient with sample-wise missingness.}]{\includegraphics[width=0.4\textwidth]{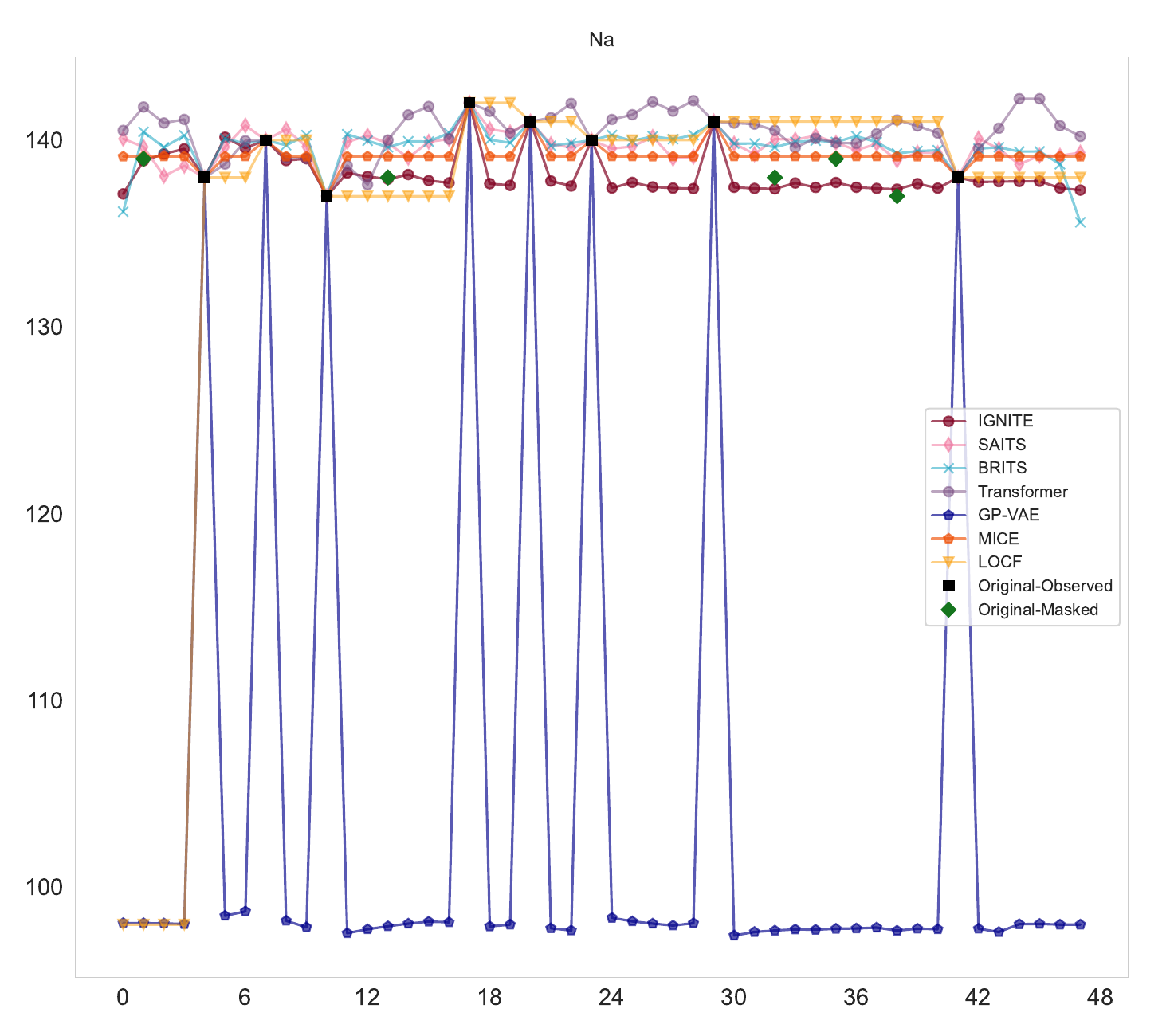}}
		\label{Colab}
   \vspace{-0.2cm}
	\subfigure[{Imputation for a patient with feature-wise missingness.}]{\includegraphics[width=0.4\textwidth]{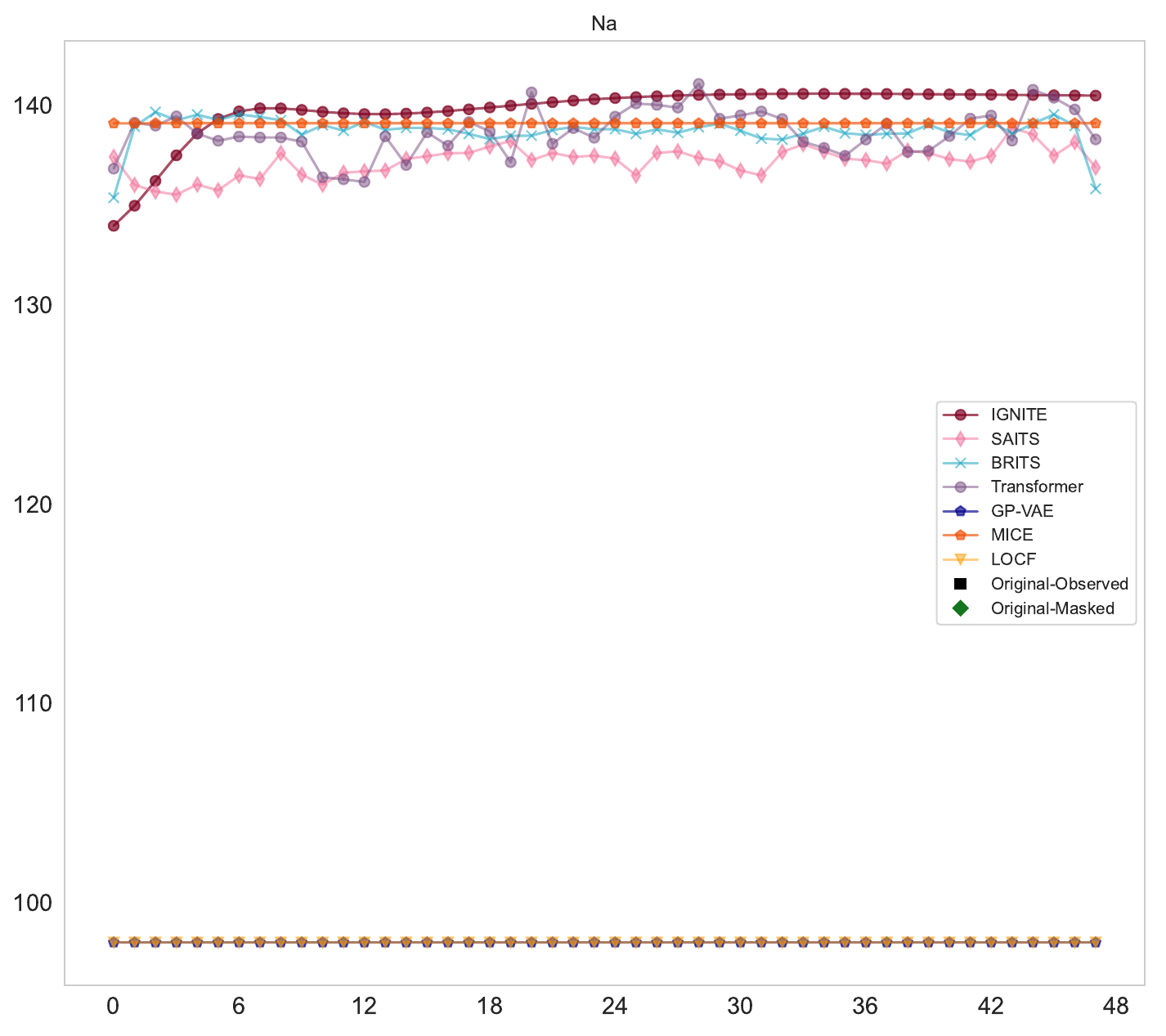}}
	
	\caption{Visualized imputations for two patients from the PhysioNet 2012 dataset. In a and b, we show examples of patients with different types of missingness. The examples shown are from the test set used for reconstruction experiments where 50\% of the observed values in the overall patient record are masked, and various imputations are compared with the ground truth. The original masked values are shown in green.}

	\label{Figure79}
\end{figure}

Below are the visualized reconstructions for two patients from the physioNet dataset. In this experiment, trained models were used to impute randomly masked samples for each patient. The approach used here was to introduce 50\% missingness in the observed values for each patient record. We can see that GP-VAE seems to overfit the training set, while LOCF is imputed by zero when there is no observed value at the start of the patient observation recording.  The visualizations for all evaluated models are shown in~\ref{Figure79}.

\clearpage
\section{Visualizations of Imputations for a dead Patient}
\label{dead}
\textcolor{black}{
This section provides a detailed overview of imputed values, for example, variables from dead patients, using IGNITE, BRITS and MICE models. In Figure \ref{dead_f}, we present a focused comparison of imputed values for urine output, HCO3, and FiO2, showcasing the performance of each model. These three variables serve as key examples, with additional imputed results for the full set of patient variables included in this section. These results underscore the superior ability of IGNITE to generate clinically meaningful imputations for all patient variables.}
\begin{figure}[h!]
\centering
	\subfigure[{ALP}]{\includegraphics[width=0.3\textwidth]{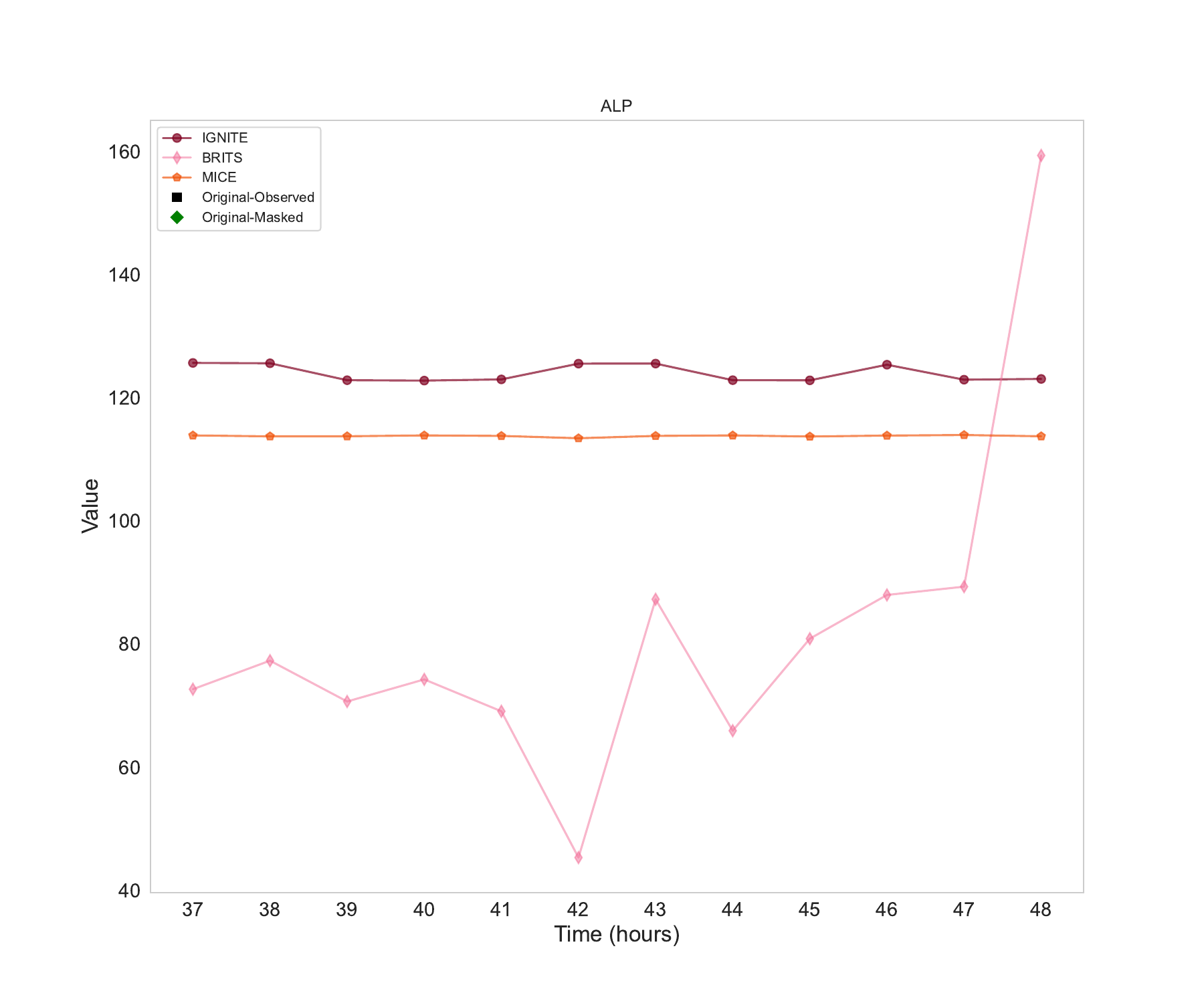}}
		\label{Colab}
	\subfigure[{Temperature}]{\includegraphics[width=0.3\textwidth]{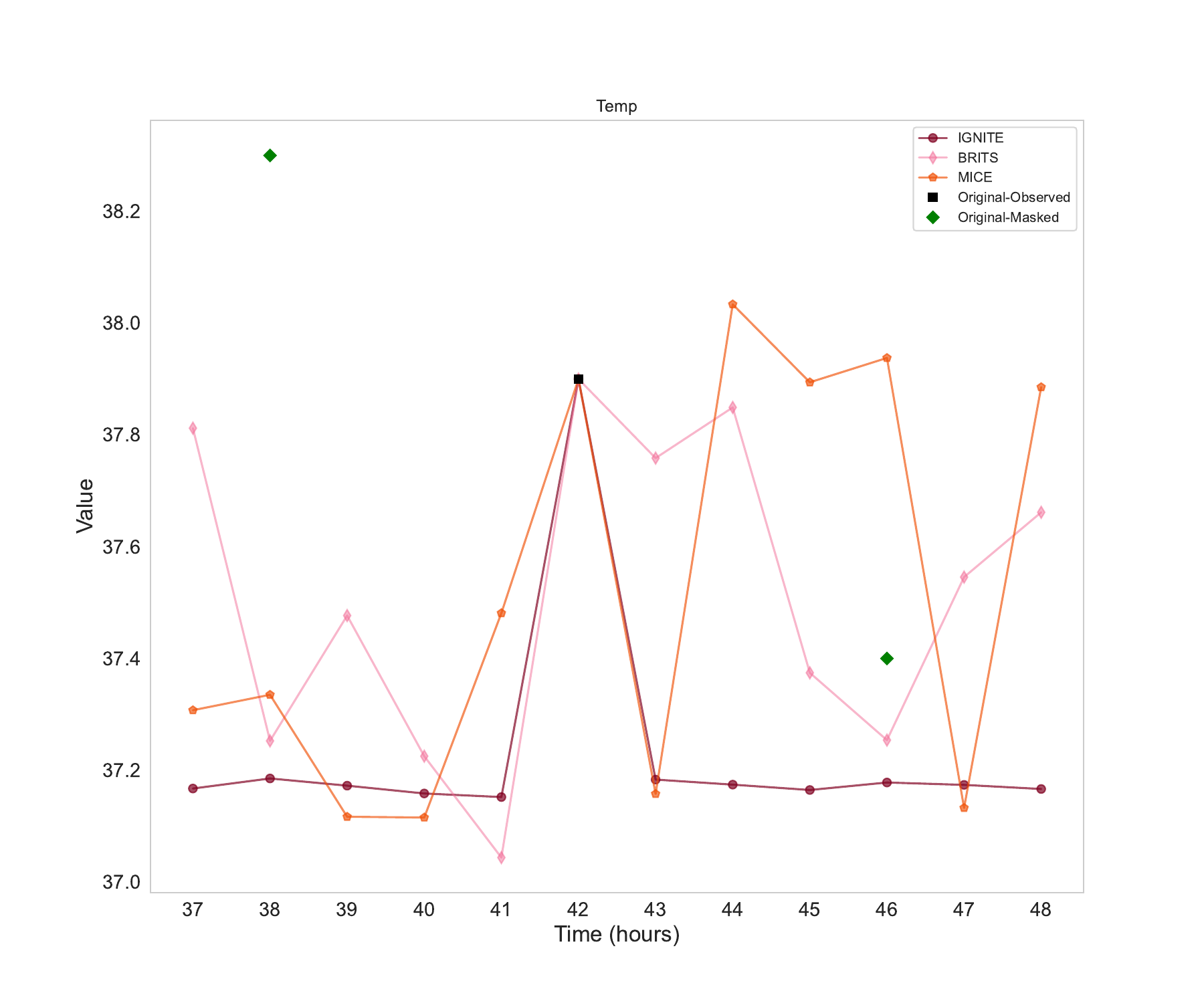}}
 	\subfigure[{Lactate}]{\includegraphics[width=0.3\textwidth]{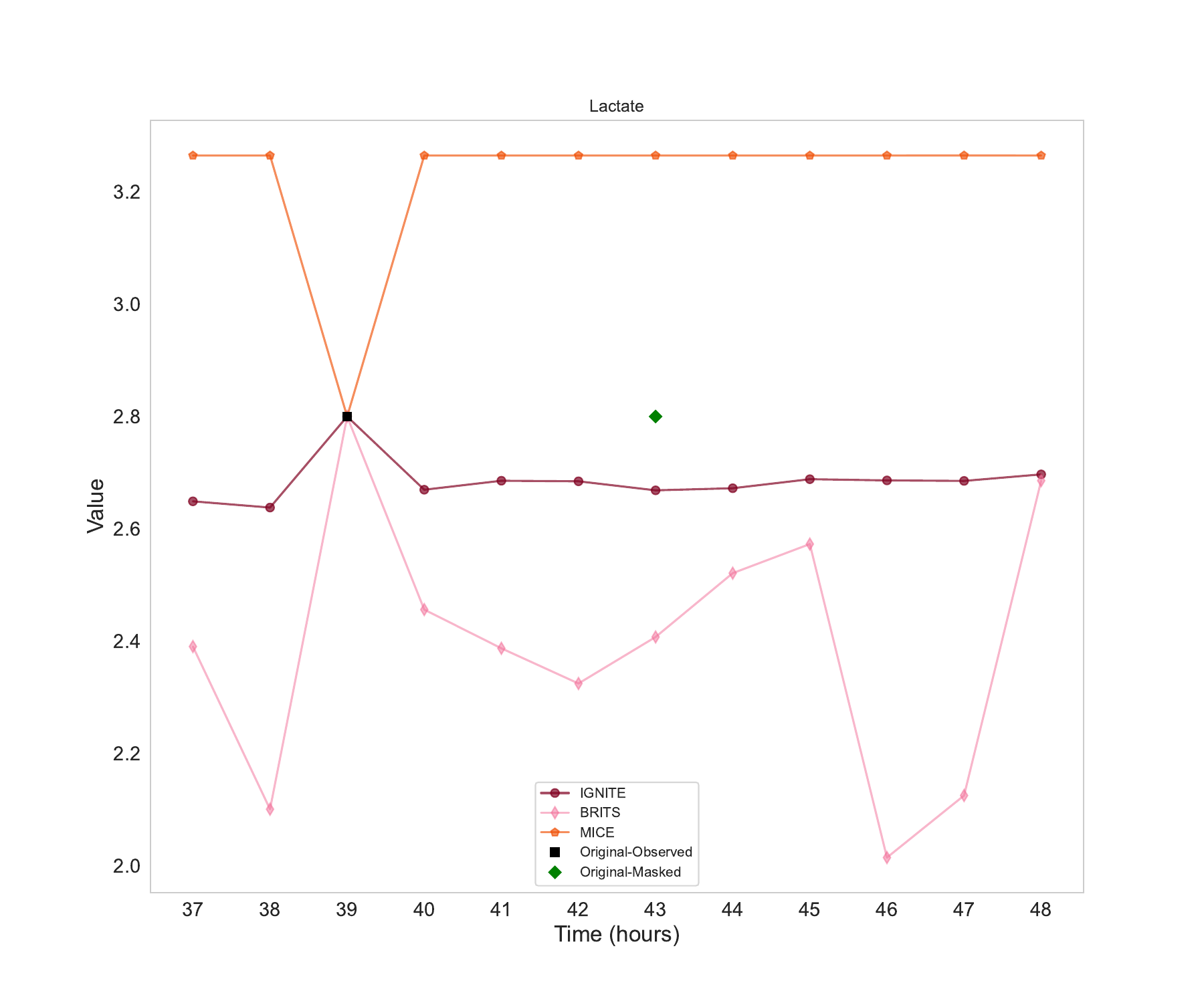}}
  	\subfigure[{WBC}]{\includegraphics[width=0.3\textwidth]{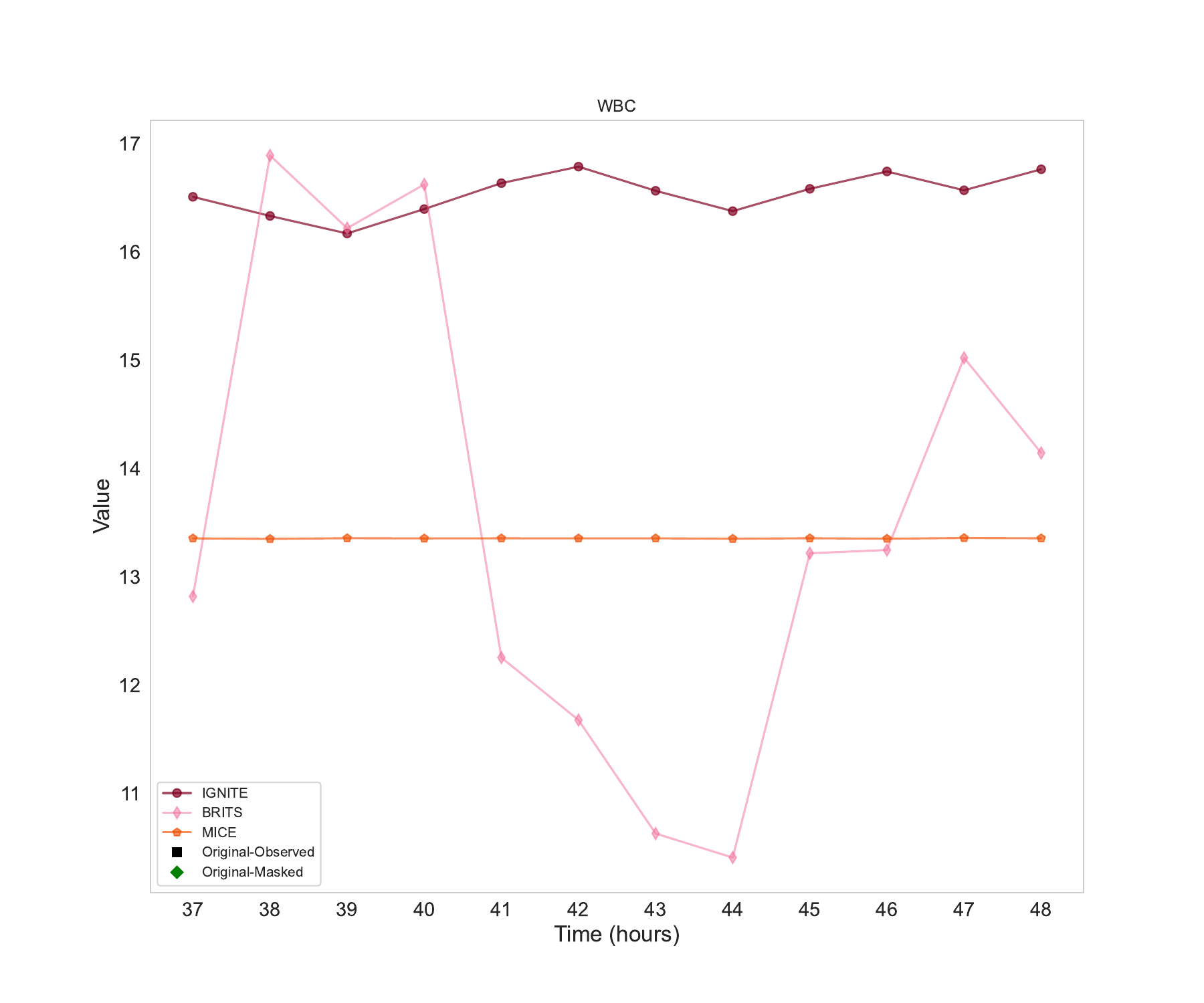}}	\subfigure[{Na}]{\includegraphics[width=0.3\textwidth]{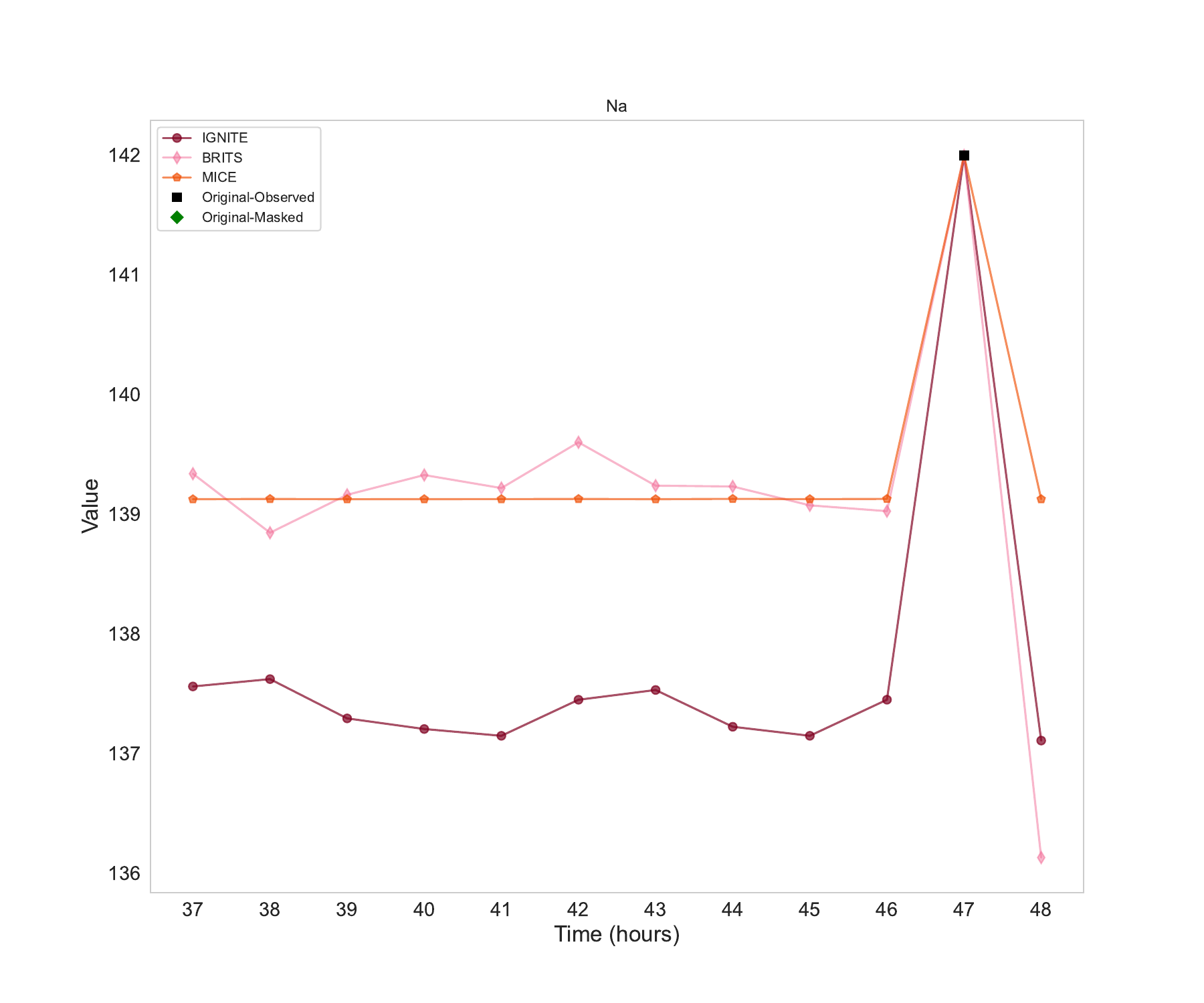}}	\subfigure[{Non Invasive Diastolic Blood pressure}]{\includegraphics[width=0.3\textwidth]{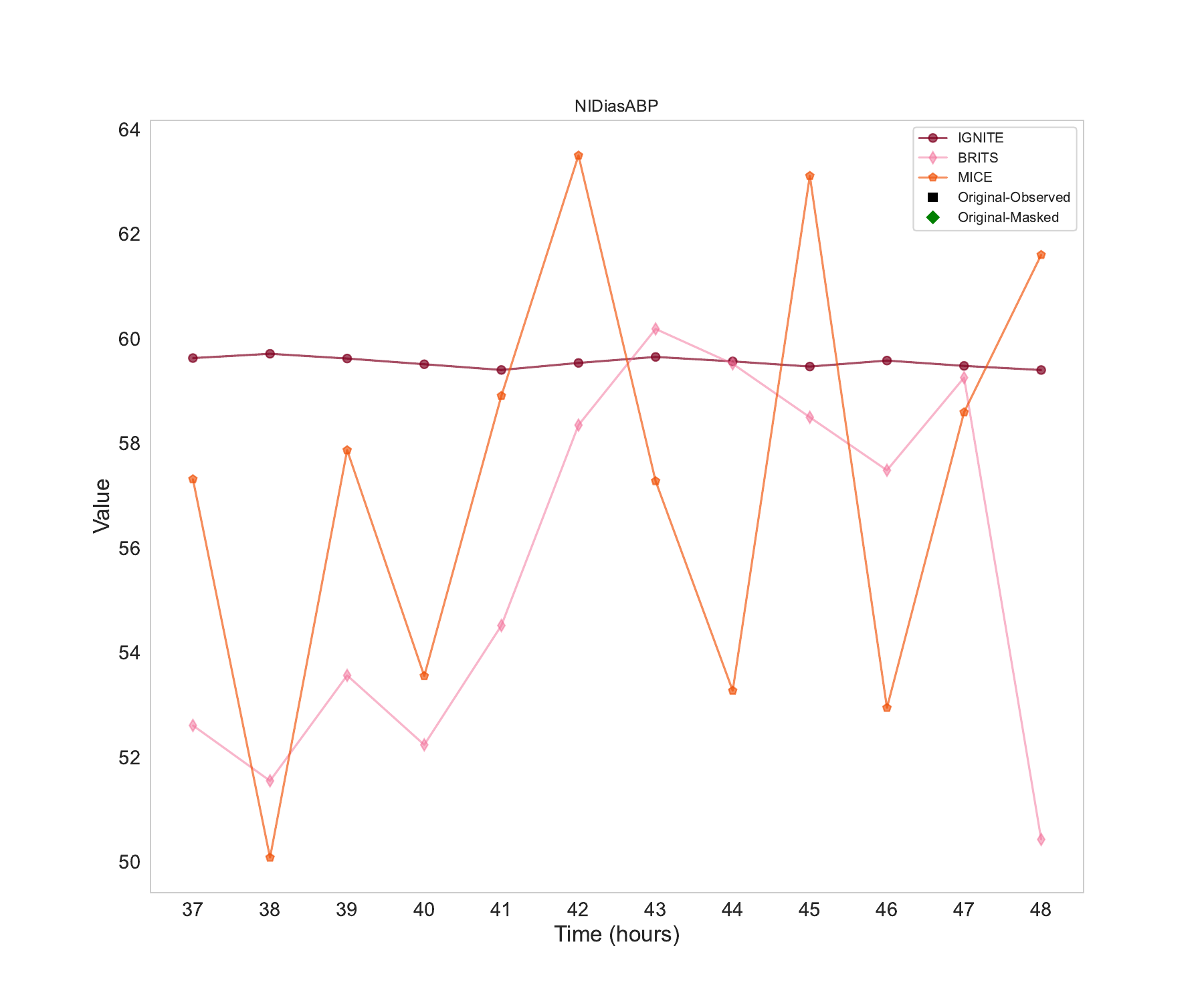}}	\subfigure[{Heart Rate}]{\includegraphics[width=0.3\textwidth]{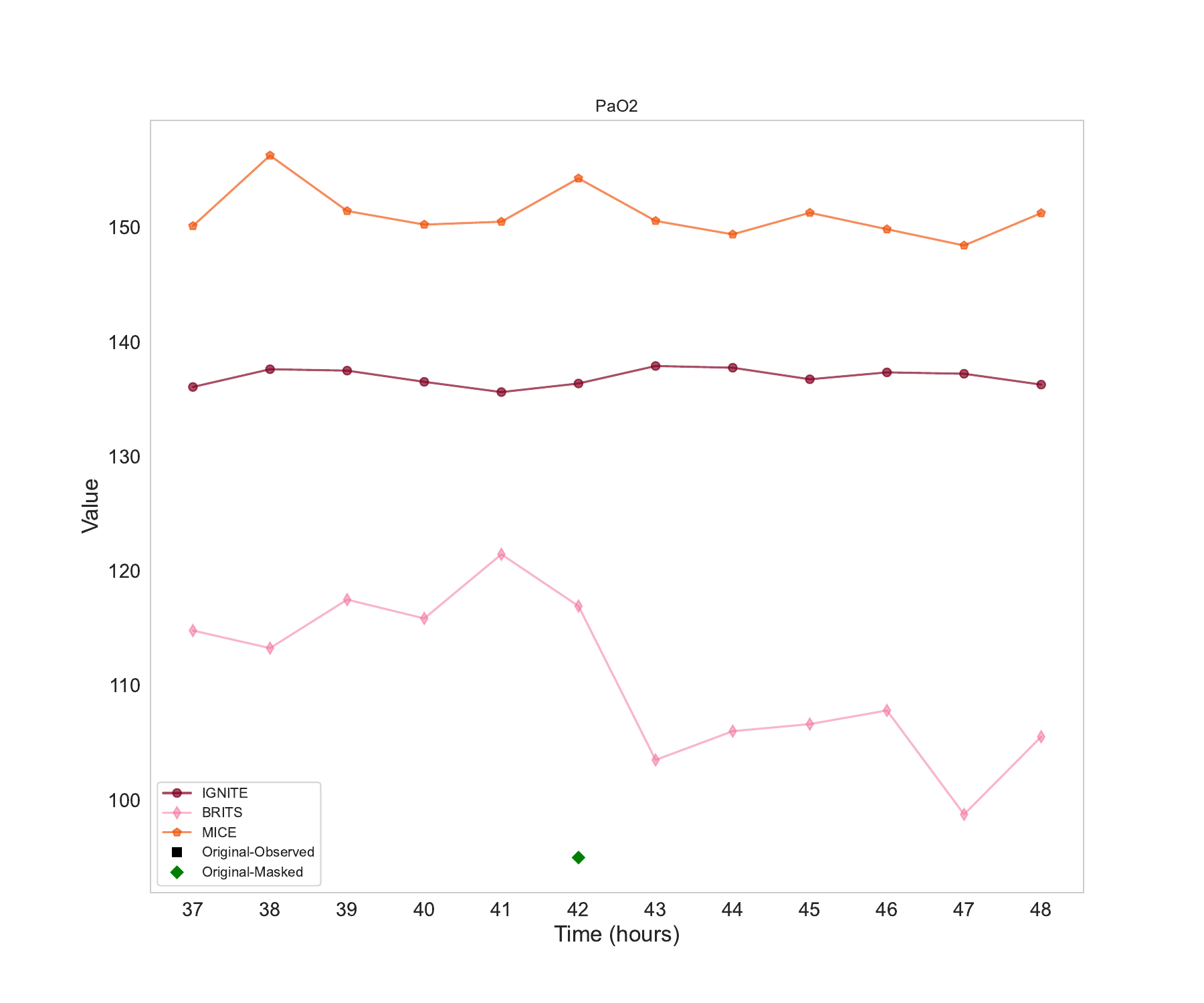}}	\subfigure[{Diastolic Blood Pressure}]{\includegraphics[width=0.3\textwidth]{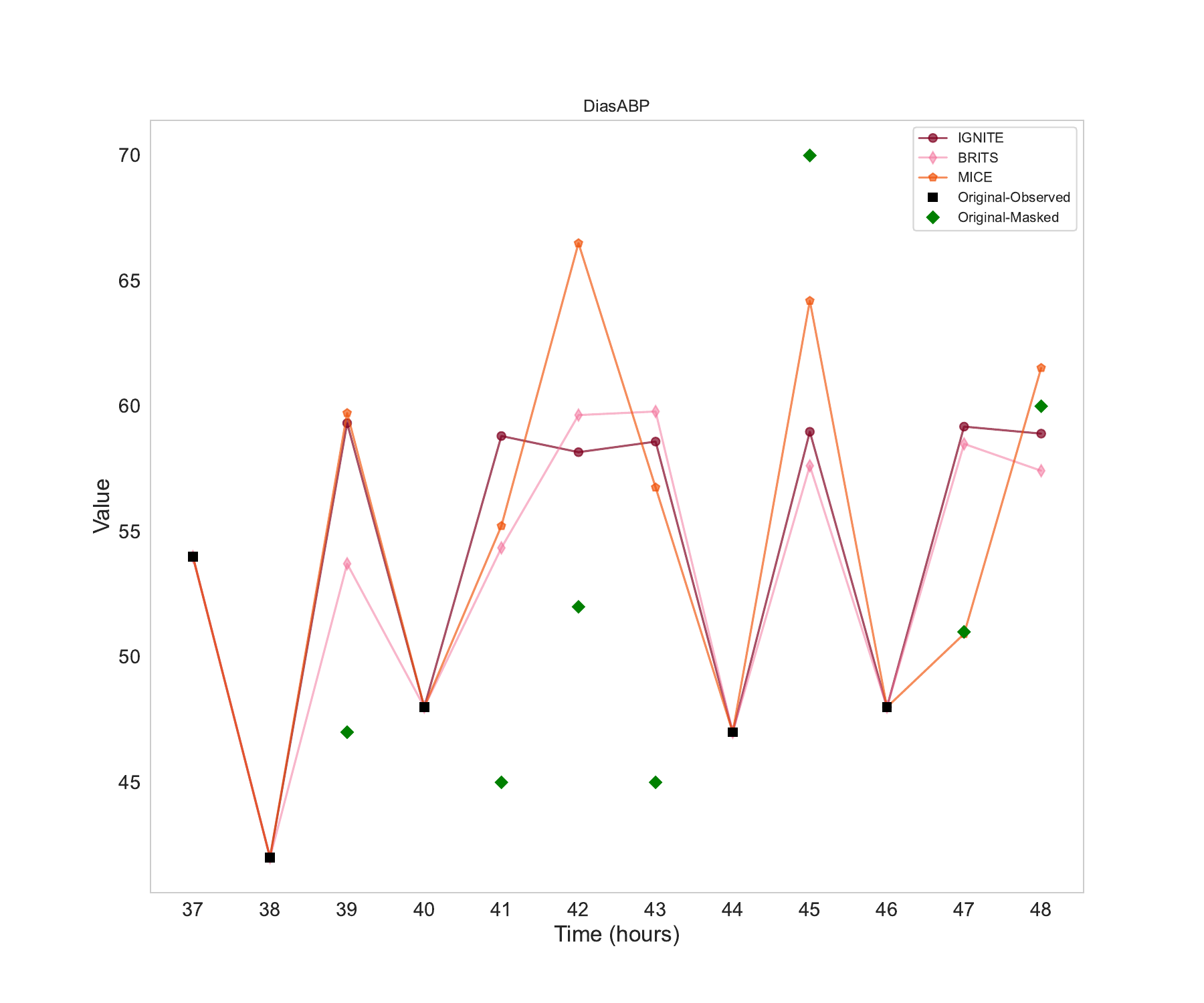}}
\subfigure[{pH}]
   {\includegraphics[width=0.3\textwidth]{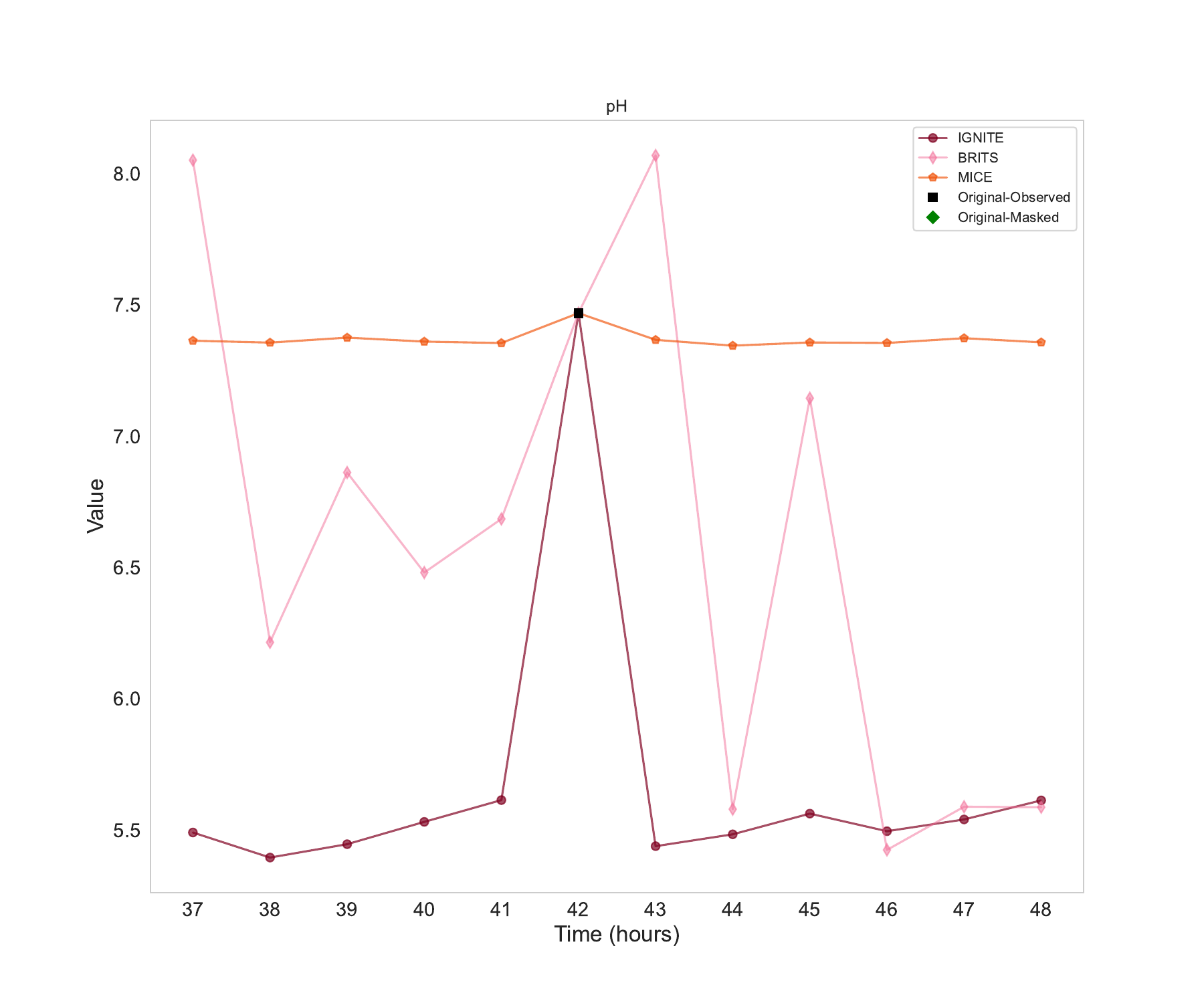}}		

 \end{figure}

 \begin{figure}[h!]
\centering
	\subfigure[{Mg}]{\includegraphics[width=0.3\textwidth]{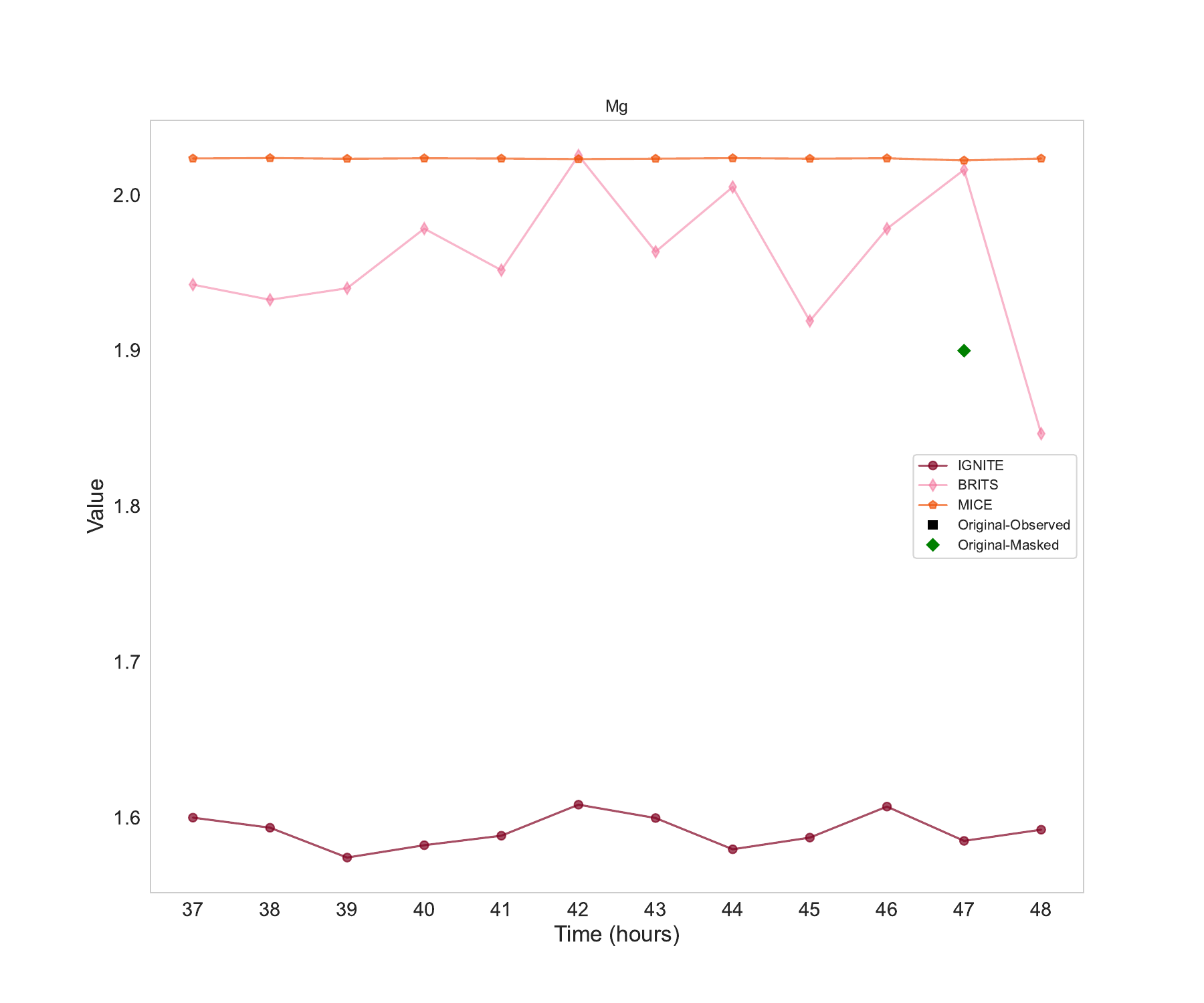}}
		\label{Colab}
	\subfigure[{TroponinI}]{\includegraphics[width=0.3\textwidth]{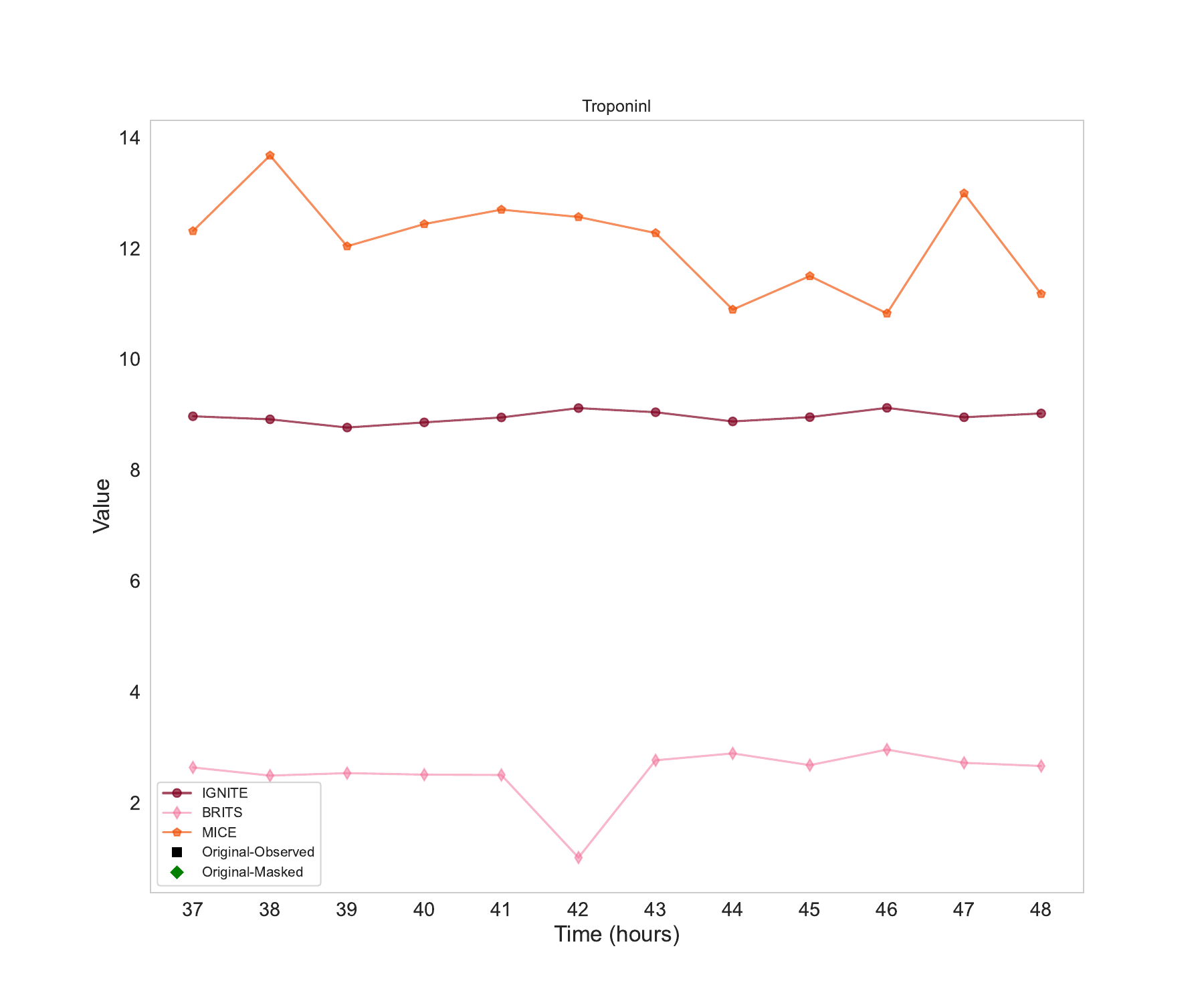}}
	\subfigure[{TroponinT}]{\includegraphics[width=0.3\textwidth]{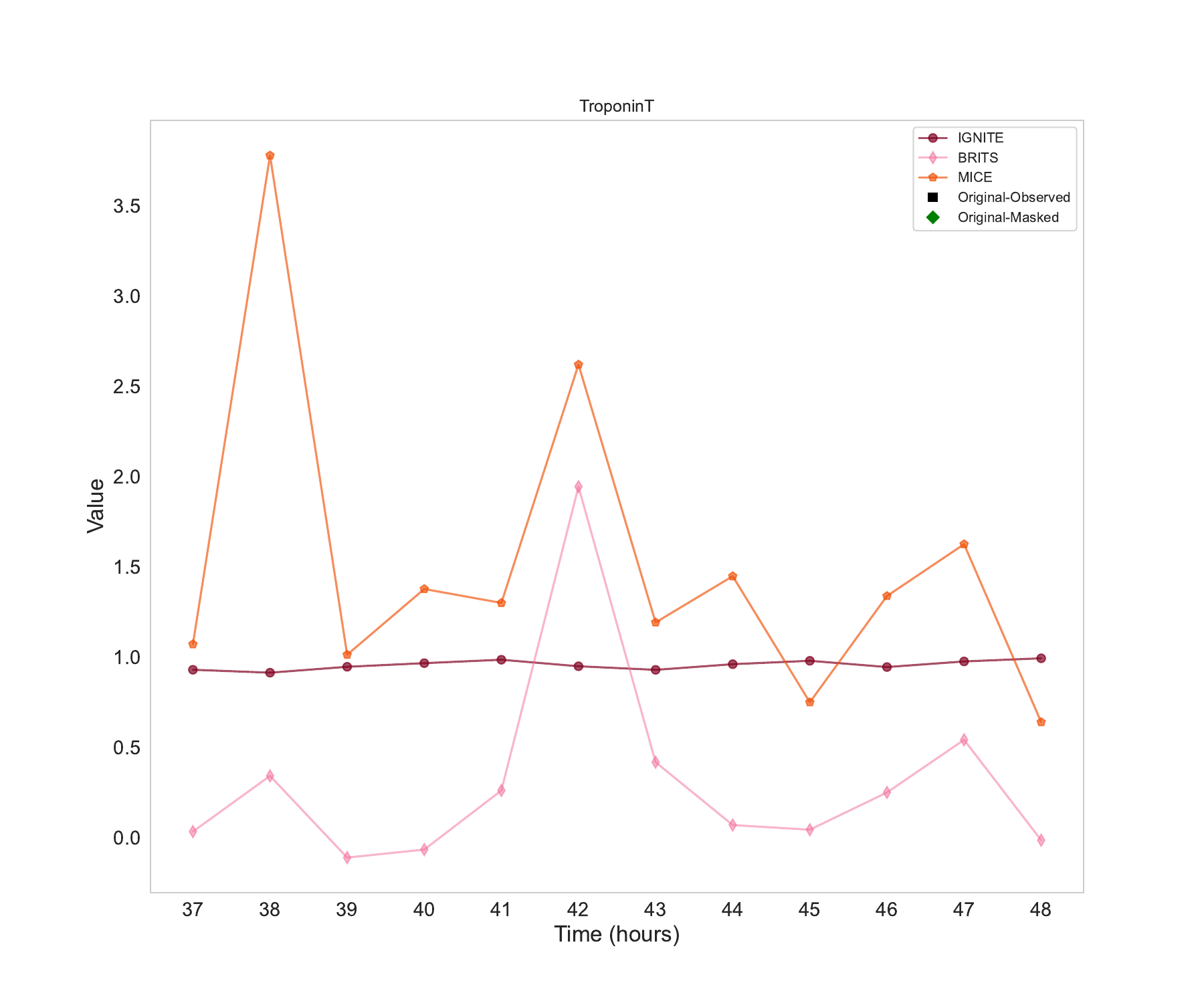}}
 	\subfigure[{Bilirubin}]{\includegraphics[width=0.3\textwidth]{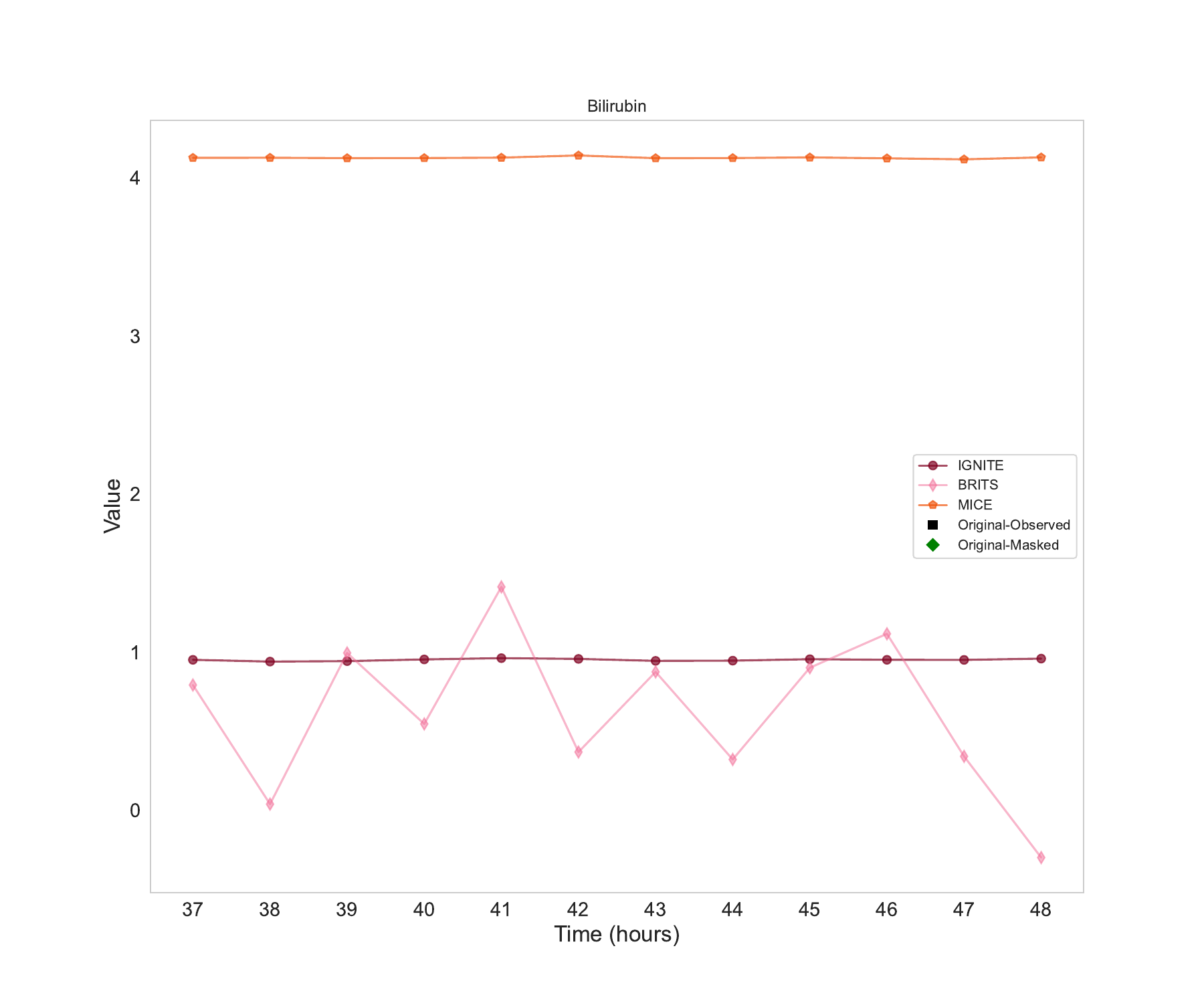}}
  	\subfigure[{BUN}]{\includegraphics[width=0.3\textwidth]{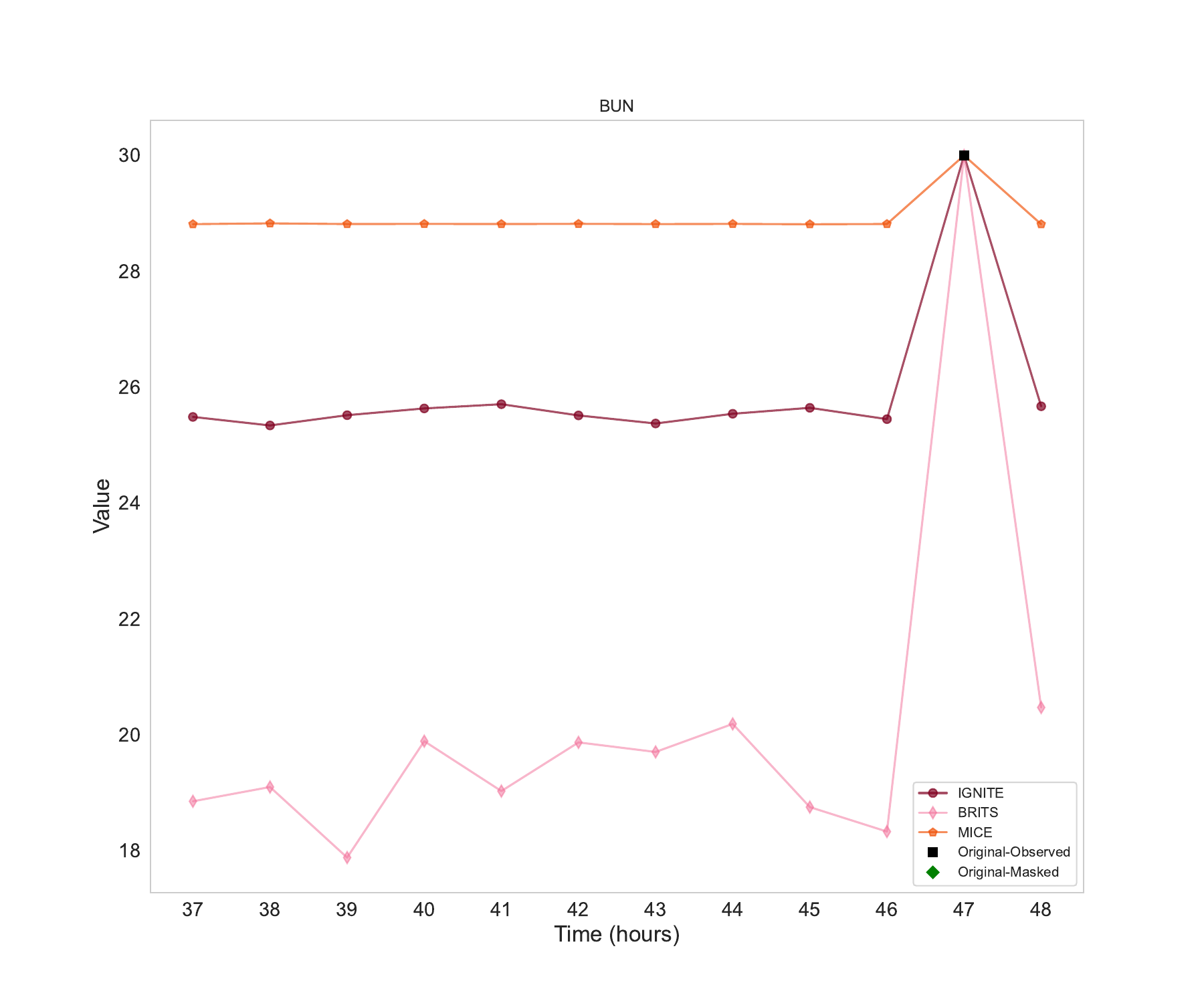}}	\subfigure[{Albumin}]{\includegraphics[width=0.3\textwidth]{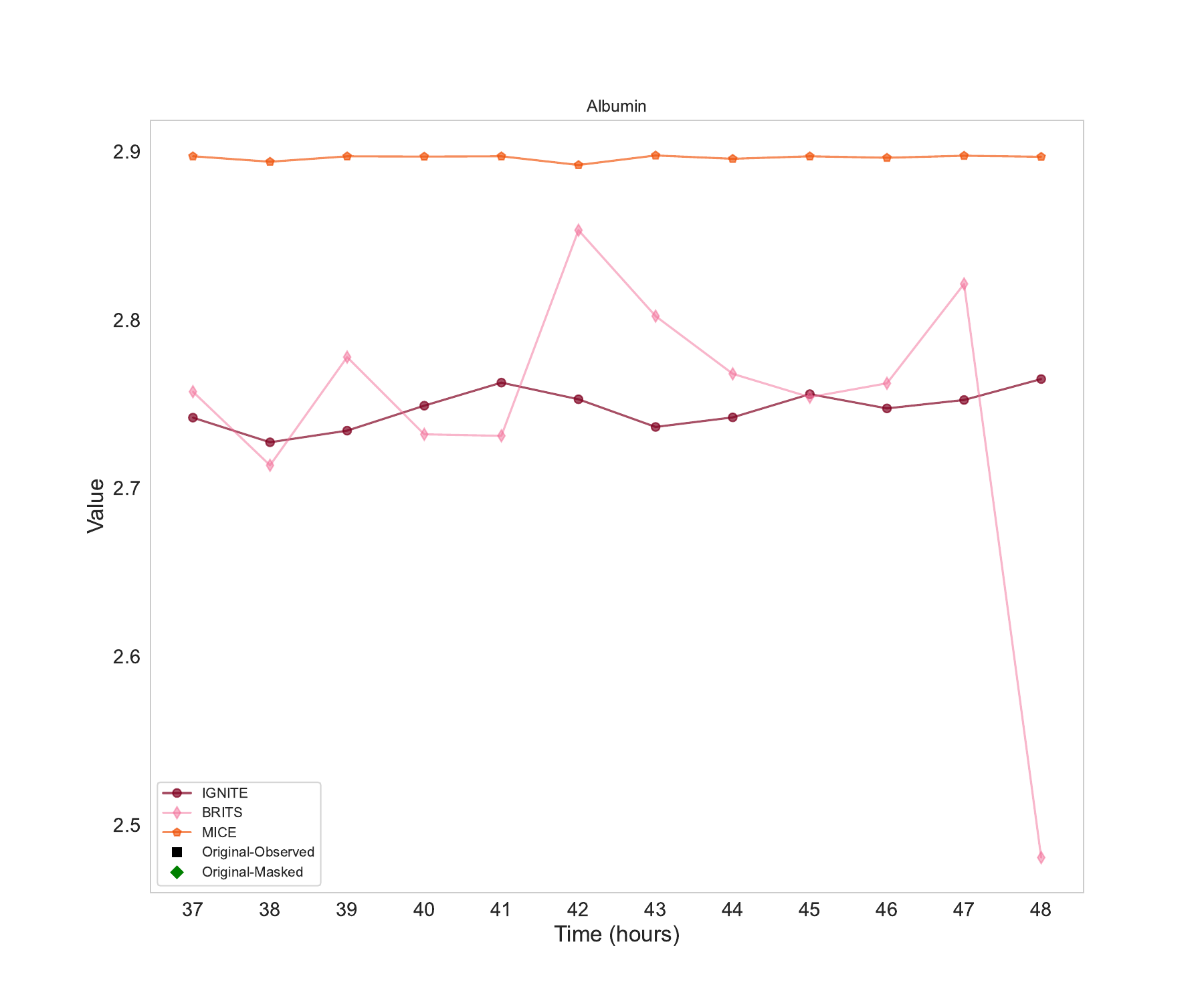}}	\subfigure[{AST}]{\includegraphics[width=0.3\textwidth]{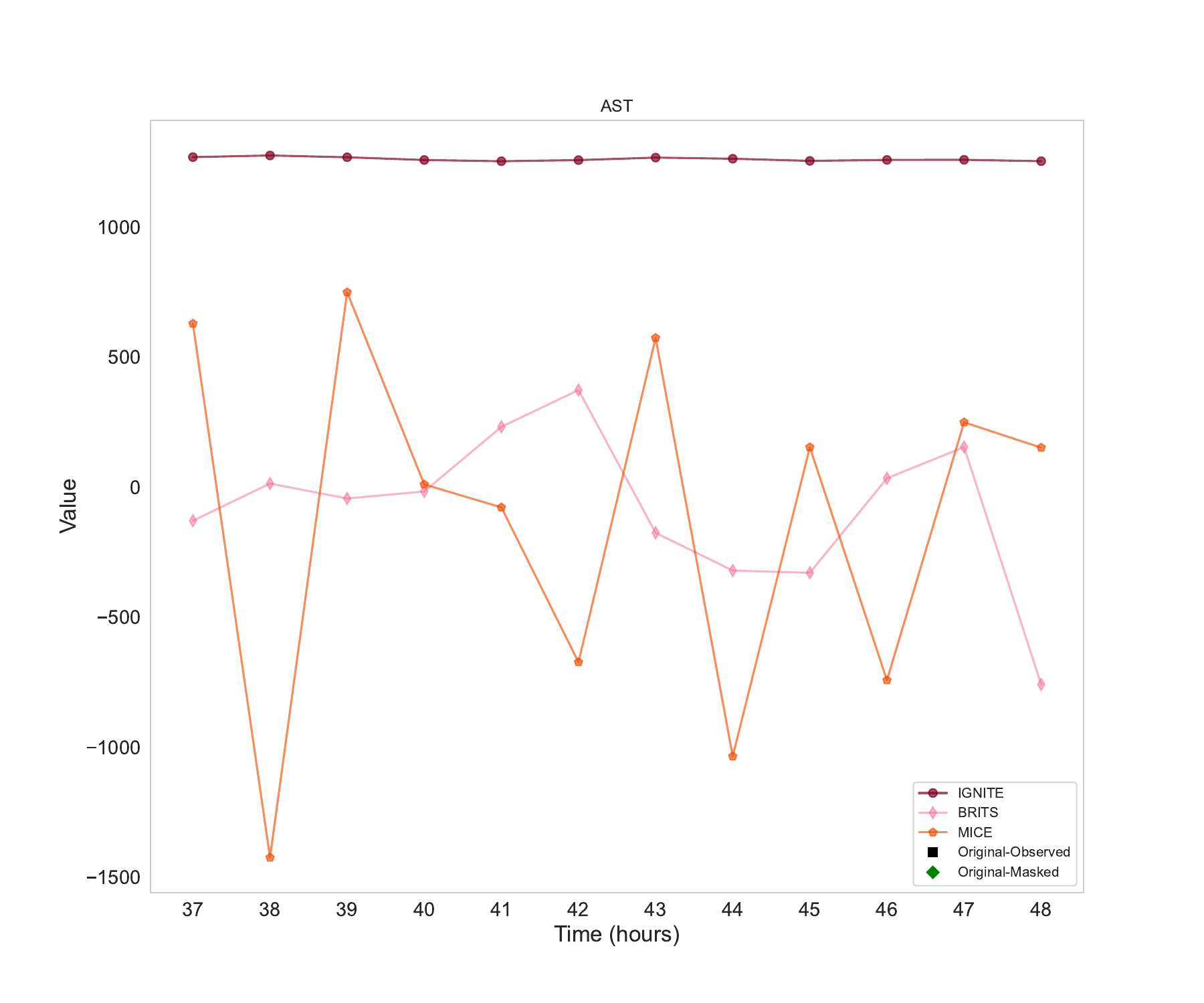}}	\subfigure[{ALT}]{\includegraphics[width=0.3\textwidth]{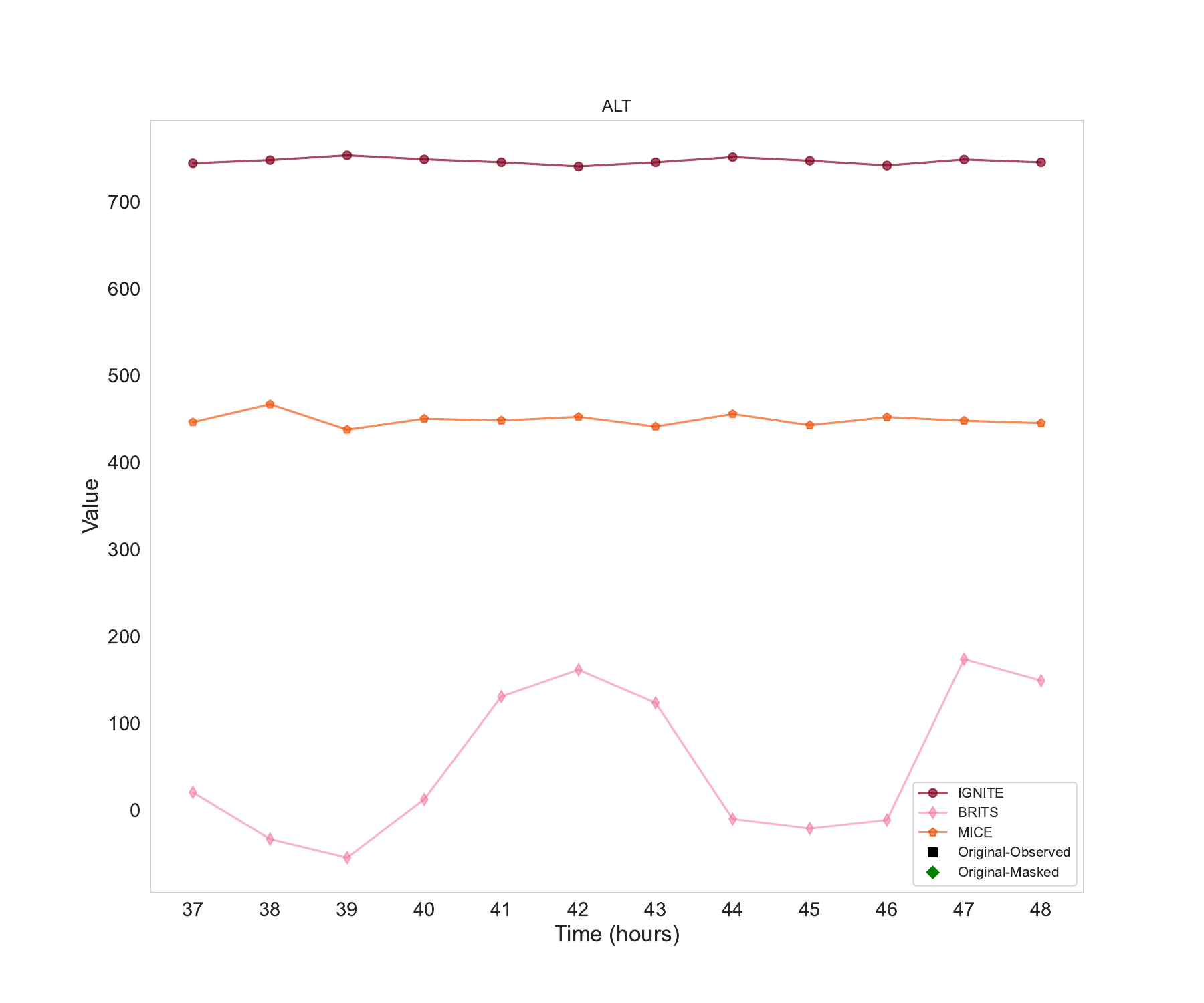}}	\subfigure[{Glucose}]{\includegraphics[width=0.3\textwidth]{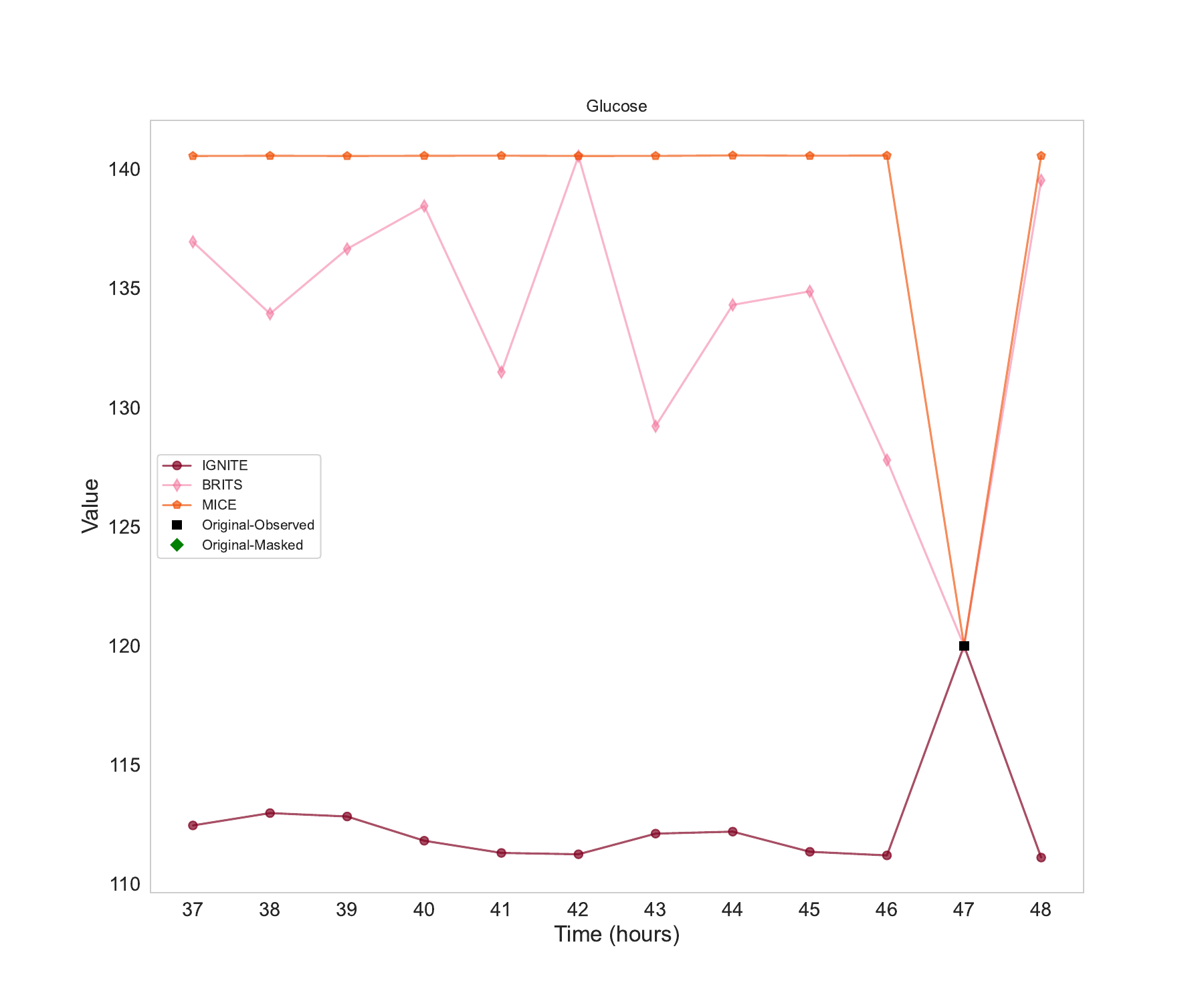}}
   \subfigure[{MAP}]
   {\includegraphics[width=0.3\textwidth]{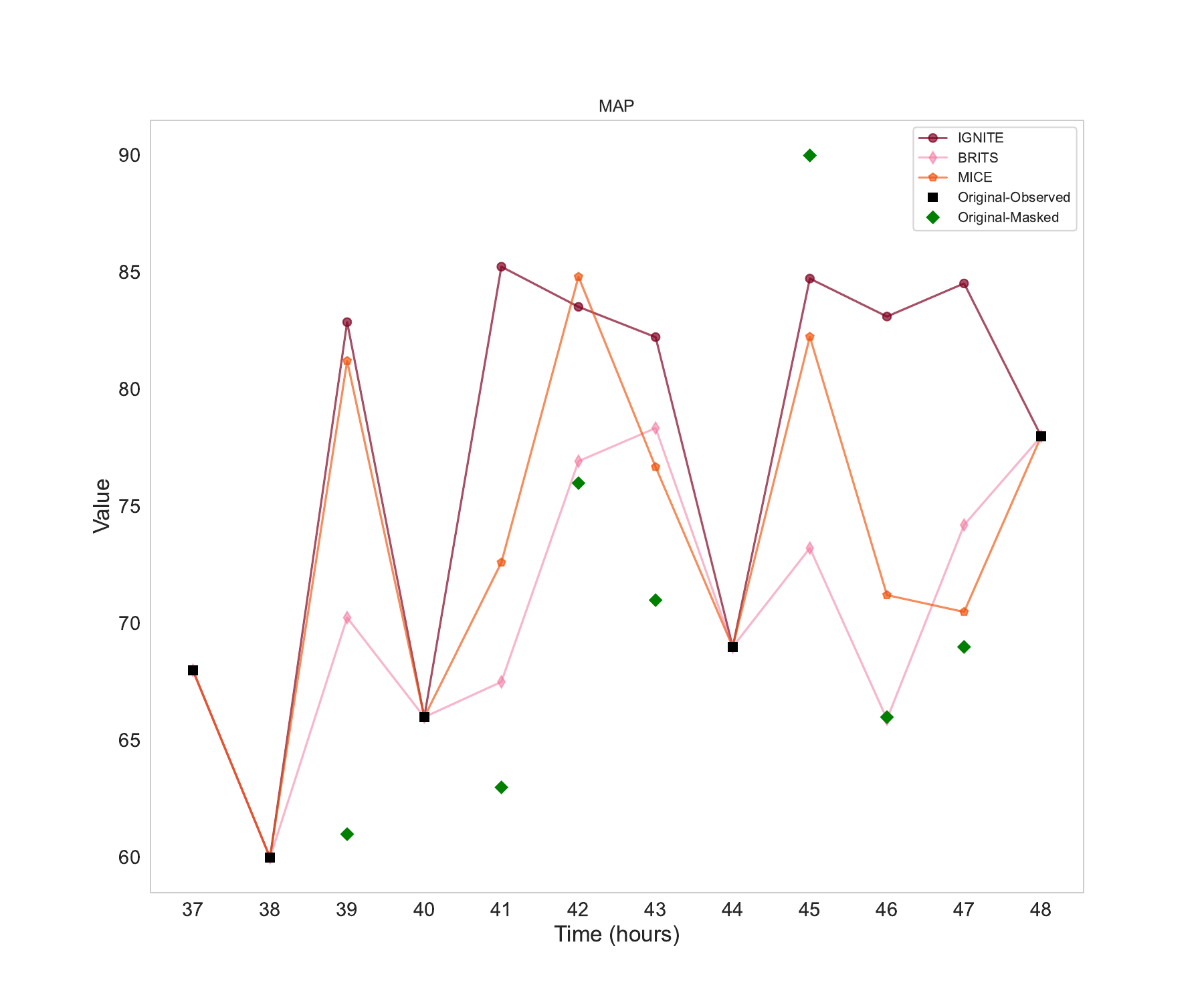}}	\subfigure[{K}]{\includegraphics[width=0.3\textwidth]{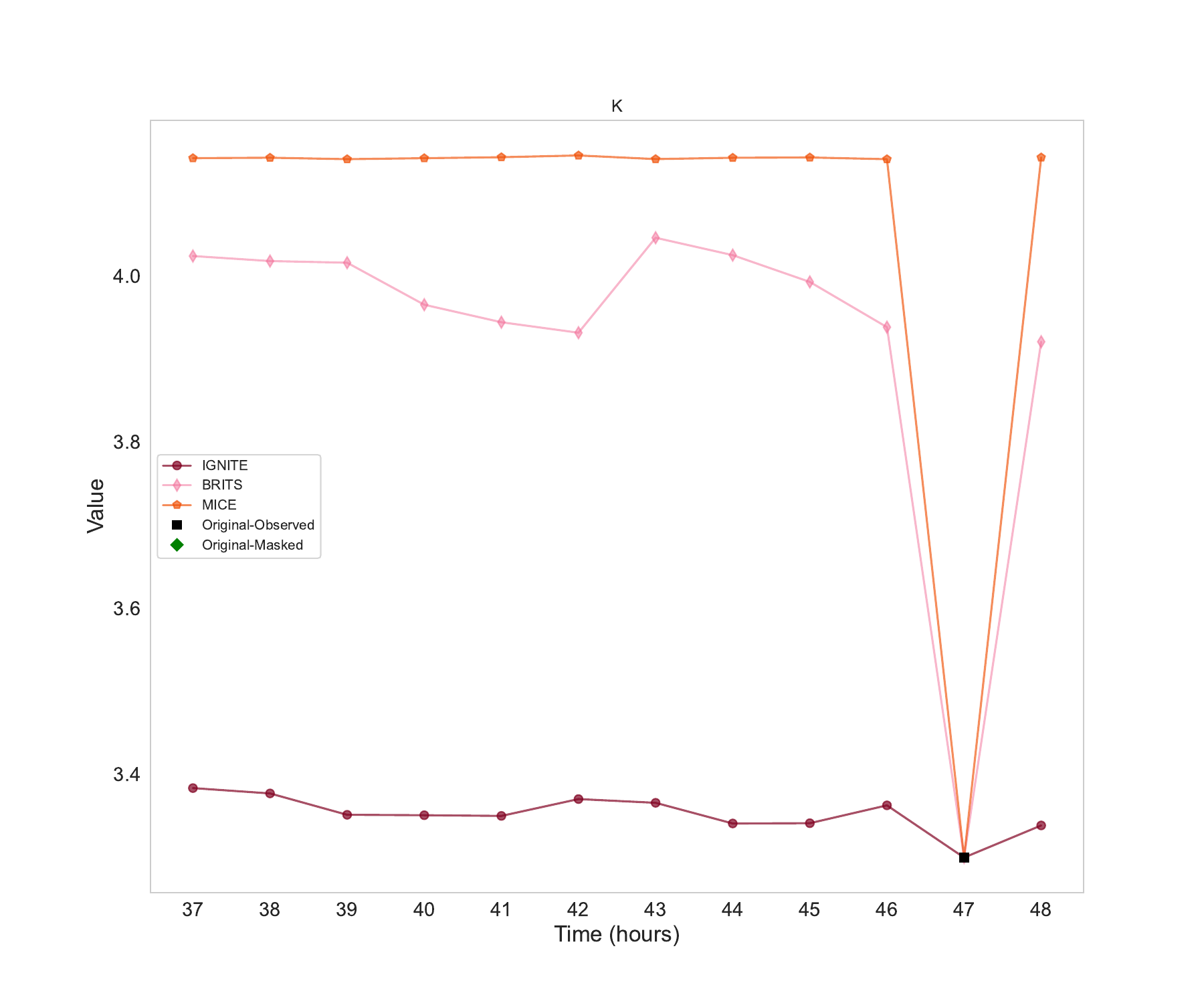}}	\subfigure[{NISysABP}]{\includegraphics[width=0.3\textwidth]{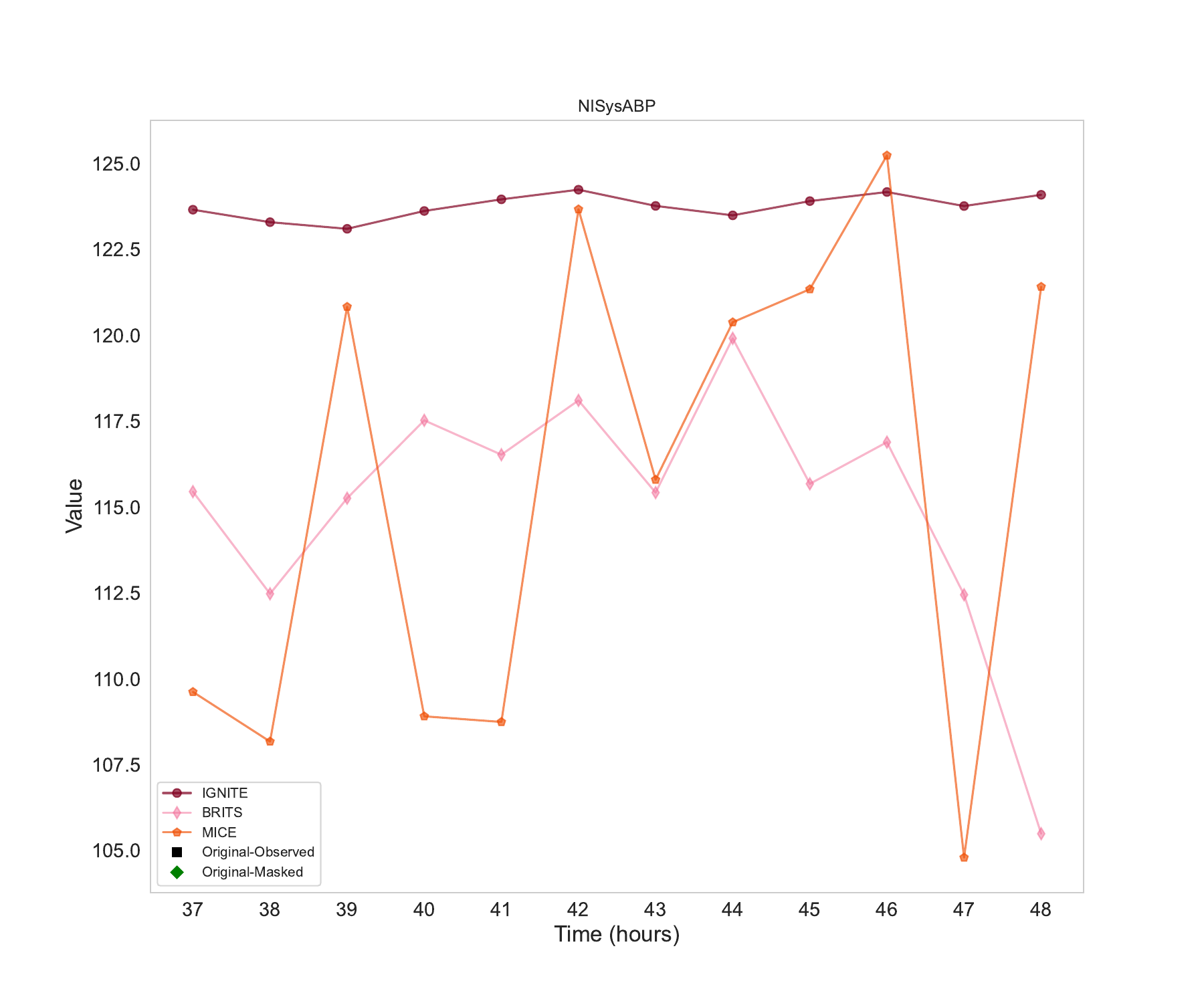}}	

 \end{figure}

 \begin{figure}[h!]
\centering
	\subfigure[{Creatinine}]{\includegraphics[width=0.3\textwidth]{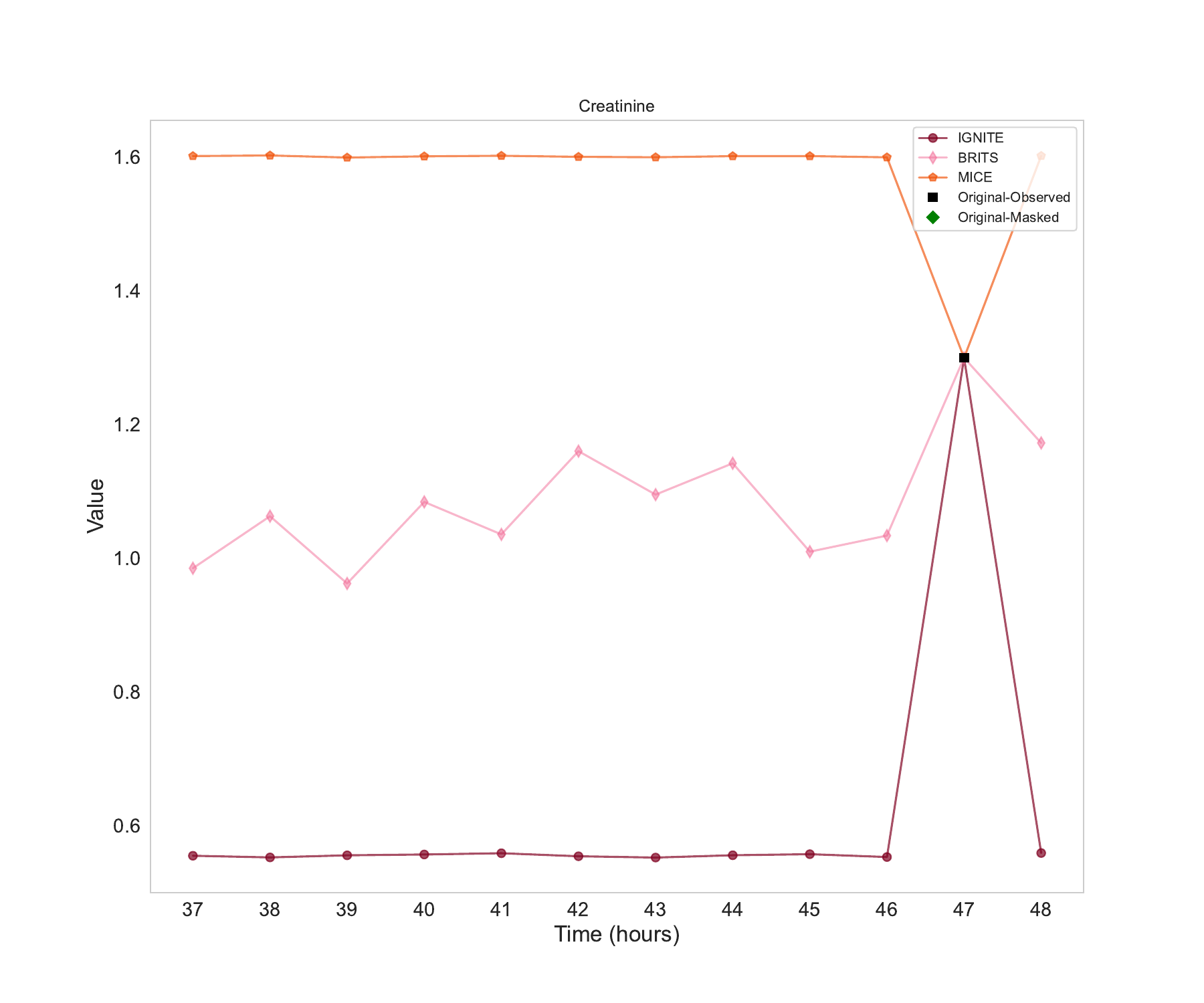}}
		\label{Colab}
	\subfigure[{Cholesterol}]{\includegraphics[width=0.3\textwidth]{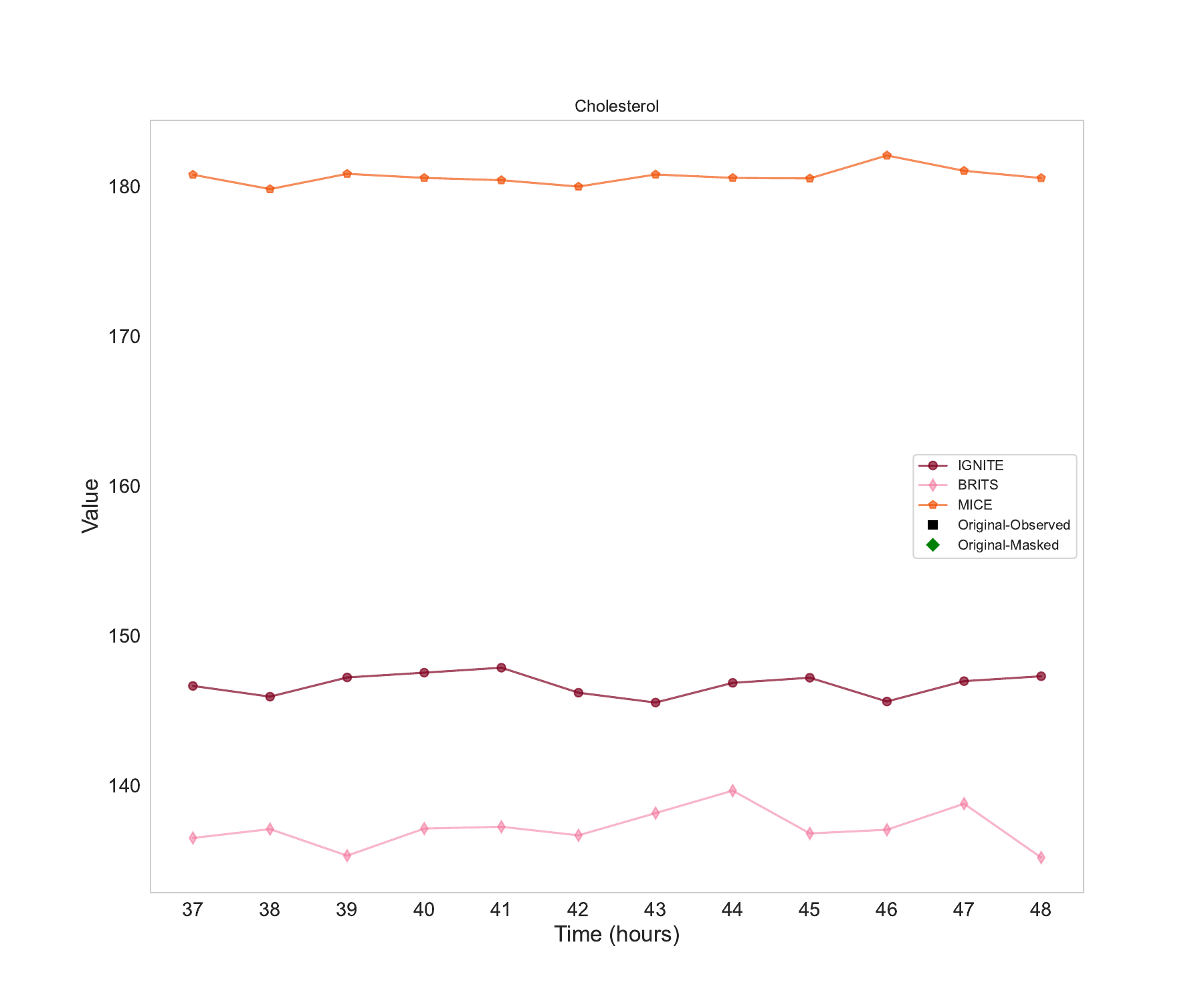}}
	\subfigure[{GCS}]{\includegraphics[width=0.3\textwidth]{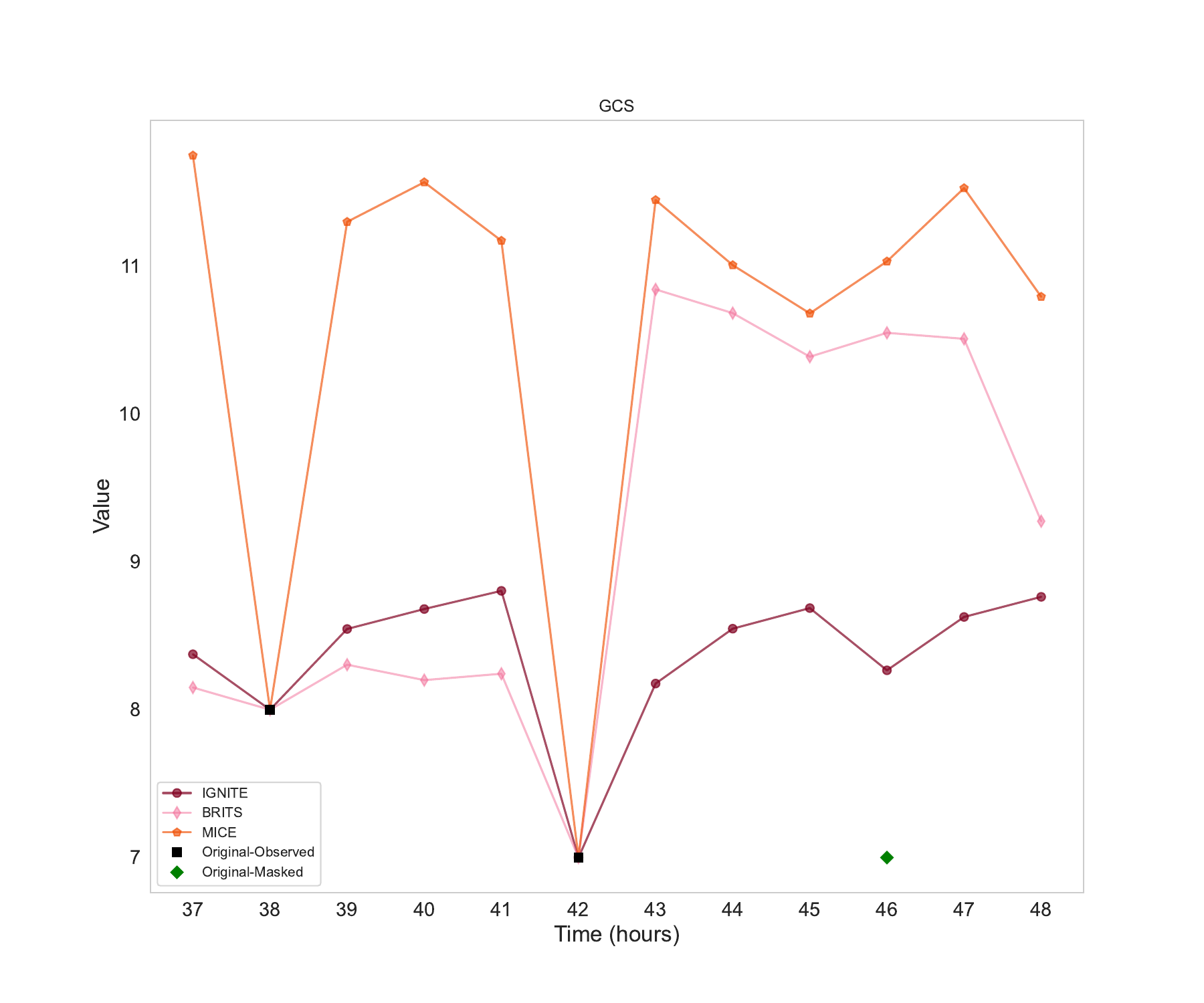}}
 	\subfigure[{NIMAP}]{\includegraphics[width=0.3\textwidth]{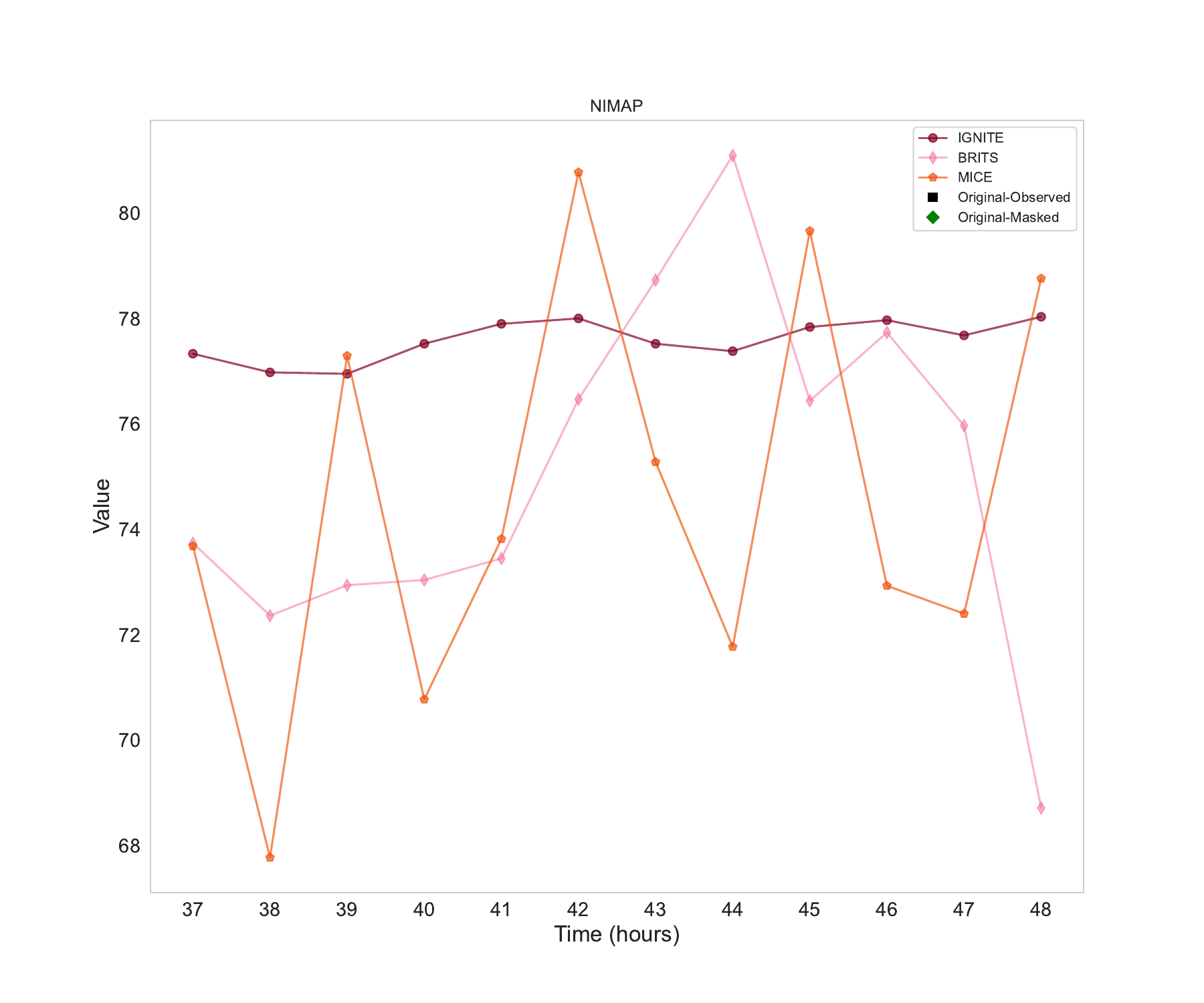}}
  	\subfigure[{PaCo2}]{\includegraphics[width=0.3\textwidth]{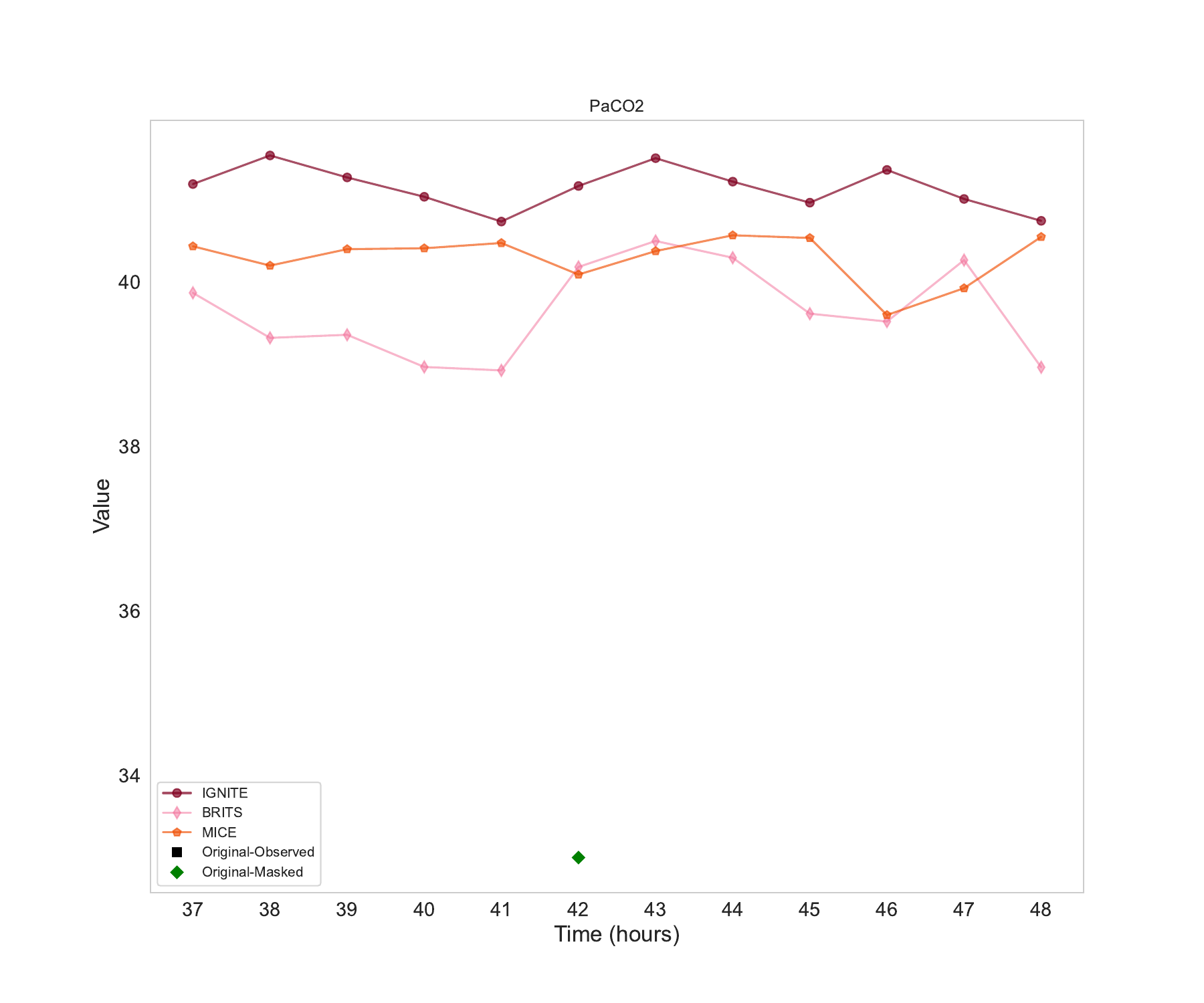}}	\subfigure[{Platlets}]{\includegraphics[width=0.3\textwidth]{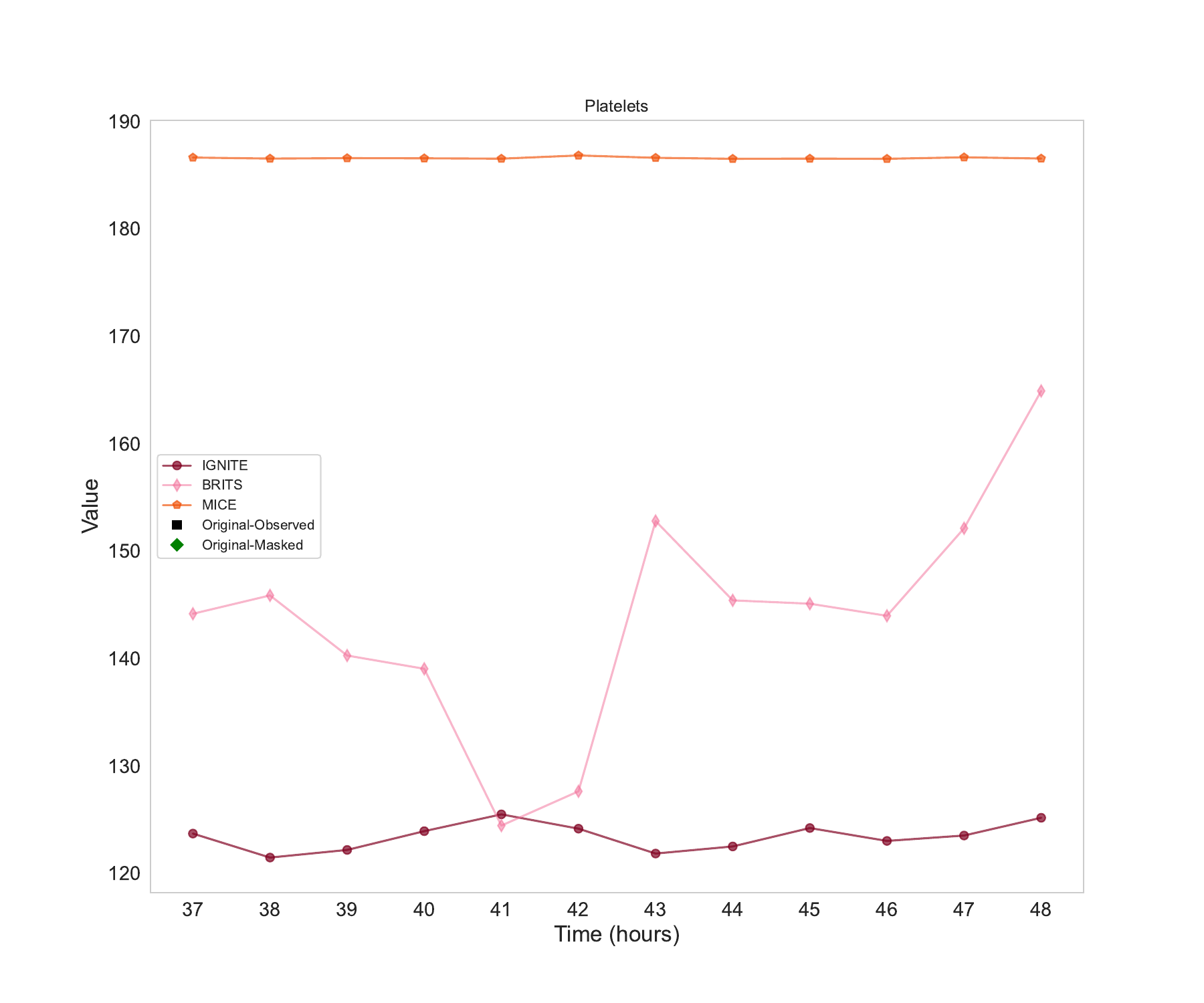}}
\subfigure[{SaO2}]{\includegraphics[width=0.3\textwidth]{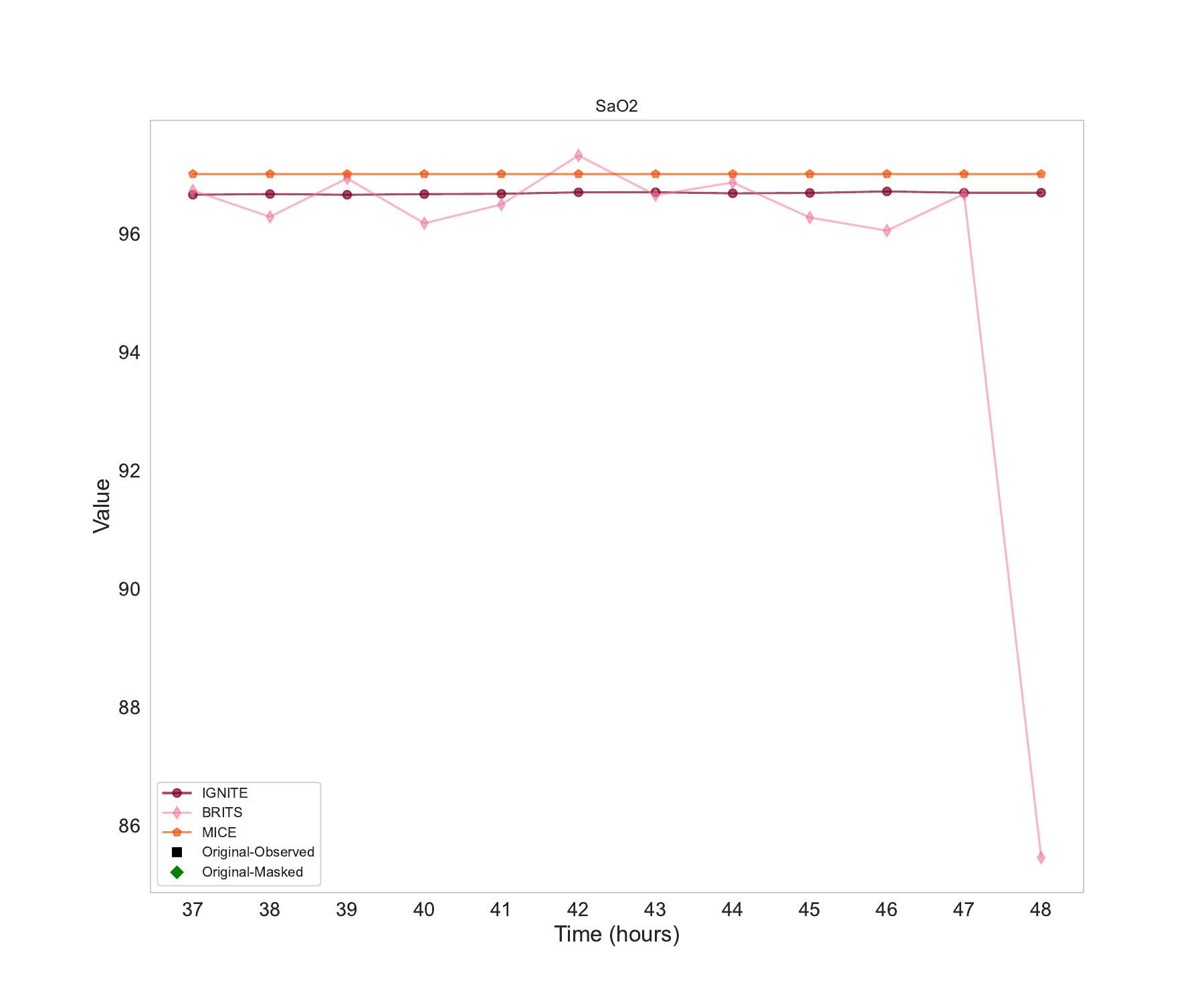}}	\subfigure[{Respiratory Rate}]{\includegraphics[width=0.3\textwidth]{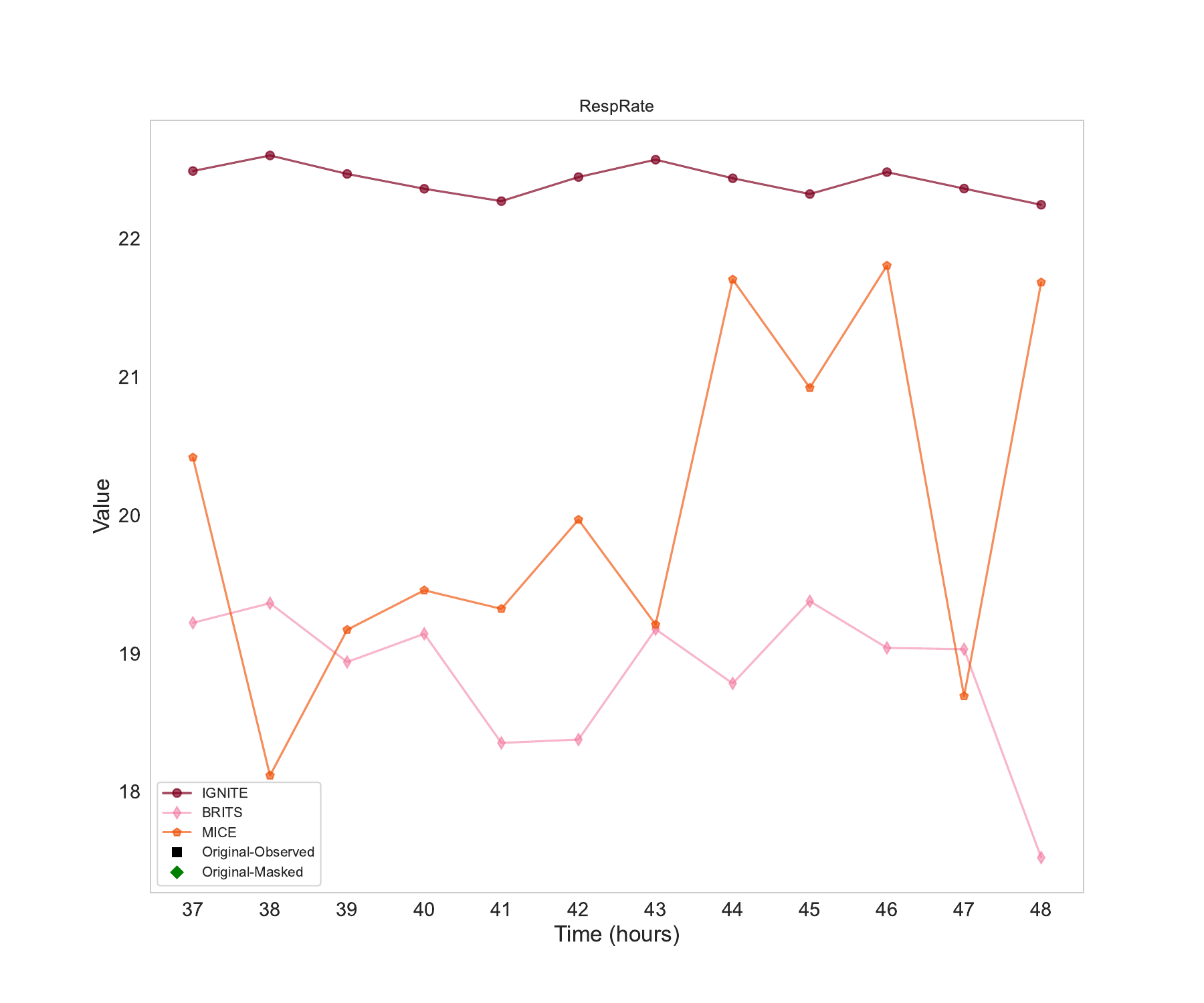}}
      
 \end{figure}
\end{document}